\documentclass[Afour,sageh,times]{sagej}
\usepackage{algorithmic}
\usepackage{algorithm}
\usepackage{moreverb,url}
\usepackage{subfigure}
\usepackage{parskip}
\usepackage{overcite}
\usepackage{cite}
\usepackage{soul}
\usepackage{color, xcolor}
\usepackage[colorlinks,bookmarksopen,bookmarksnumbered,citecolor=red,urlcolor=red]{hyperref}

\usepackage{natbib}
\setcitestyle{numbers,square}

\usepackage[colorlinks,bookmarksopen,bookmarksnumbered,citecolor=red,urlcolor=red]{hyperref}

\newcommand\BibTeX{{\rmfamily B\kern-.05em \textsc{i\kern-.025em b}\kern-.08em
T\kern-.1667em\lower.7ex\hbox{E}\kern-.125emX}}

\begin{document}

\title{Coordinated Control of Path Tracking and Yaw Stability for Distributed Drive Electric Vehicle Based on AMPC and DYC}

\author{Dongmei Wu \affilnum{1,2,3}, Yuying Guan \affilnum{1,2,3}, Xin Xia \affilnum{4}, Changqing Du \affilnum{1,2,3}, Fuwu Yan \affilnum{1,2,3}, Yang Li \affilnum{5}, Min Hua \affilnum{6}, Wei Liu \affilnum{7}}

\affiliation{\affilnum{1}Hubei Key Laboratory of Advanced Technology for Automotive Components, Wuhan University of Technology, Wuhan 430070, China\\
\affilnum{2}Foshan Xianhu Laboratory of the Advanced Energy Science and Technology Guangdong Laboratory, Foshan 528200, China\\
\affilnum{3}Hubei Research Center for New Energy \& Intelligent Connected Vehicle, Wuhan University of Technology, Wuhan 430070, China\\
\affilnum{4}Department of Civil and Environmental Engineering, University of California, Los Angeles, CA, 90024, USA\\
\affilnum{5}
Technical Center of Dongfeng Commercial Vehicle, Wuhan, China\\
\affilnum{6}School of Engineering, University of Birmingham, Birmingham, B15 2TT, UK\\
\affilnum{7}School of Electrical and Computer Engineering, Purdue University, West Lafayette, IN 47907, USA}

\corrauth{Wei Liu, School of Electrical and Computer Engineering, Purdue University, West Lafayette, IN 47907, USA}

\email{liuw1619@gmail.com}

\begin{abstract}
Maintaining both path-tracking accuracy and yaw stability of distributed drive electric vehicles (DDEVs) under various driving conditions presents a significant challenge in the field of vehicle control. To address this limitation, a coordinated control strategy that integrates adaptive model predictive control (AMPC) path-tracking control and direct yaw moment control (DYC) is proposed for DDEVs. The proposed strategy, inspired by a hierarchical framework, is coordinated by the upper layer of path-tracking control and the lower layer of direct yaw moment control. Based on the linear time-varying model predictive control (LTV MPC) algorithm, the effects of prediction horizon and weight coefficients on the path-tracking accuracy and yaw stability of the vehicle are compared and analyzed first. According to the aforementioned analysis, an AMPC path-tracking controller with variable prediction horizon and weight coefficients is designed considering the vehicle speed's variation in the upper layer. The lower layer involves DYC based on the linear quadratic regulator (LQR) technique. Specifically, the intervention rule of DYC is determined by the threshold of the yaw rate error and the phase diagram of the sideslip angle. Extensive simulation experiments are conducted to evaluate the proposed coordinated control strategy under different driving conditions. The results show that, under variable speed and low adhesion conditions, the vehicle's yaw stability and path-tracking accuracy have been improved by 21.58\% and 14.43\%, respectively, compared to AMPC. Similarly, under high speed and low adhesion conditions, the vehicle's yaw stability and path-tracking accuracy have been improved by 44.30\% and 14.25\%, respectively, compared to the coordination of LTV MPC and DYC. The results indicate that the proposed adaptive path-tracking controller is effective across different speeds. Furthermore, the proposed coordinated control strategy successfully enhances the vehicle's stability while maintaining good path-tracking accuracy even under extreme conditions.

\end{abstract}

\keywords{Distributed drive electric vehicle, coordinated control, path tracking, adaptive model predictive control, direct yaw moment control}

\maketitle

\section{Introduction}
Powered by the Internet of things (IoT) and artificial intelligence\cite{xia2023automated, zhang2023mm, liu2022yolov5, zhou2023high}, intelligent vehicles have seen rapid development and deployment in recent years as a part of the broader intelligent transportation system (ITS) initiative \cite{luo2022multisource, liu2023systematic, ma2020statistical}. Especially, it mainly includes four parts: environment perception, motion planning, decision-making, and motion control. Path-tracking control is a crucial technology within the field of intelligent driving, aimed at controlling the chassis actuator's movements based on the reference path and the vehicle's real-time state. This technology enables autonomous decision-making for driving the vehicle along the reference path. Under safety conditions such as low speed, high adhesion, small curvature, the vehicle's stability is better, and the path-tracking control algorithm can focus solely on the vehicle's path-tracking accuracy. However, under extreme conditions such as high speed, low adhesion, and large curvature, the vehicle's stability is significantly reduced, increasing the risk of losing path-tracking ability due to instability \cite{ xia2022estimation,zhou2022yaw, xiong2020imu}. Thus, the path-tracking control algorithm must consider both the vehicle's path-tracking accuracy and stability to ensure that the vehicle tracks the reference path quickly, accurately and with stability.

Existing research typically employs various algorithms, including PID control, full state feedback control, sliding mode control, robust control, active disturbance rejection control, fuzzy control, and model predictive control (MPC), to design path tracking controller \cite{5, zhou2022robust, 12}. MPC has gained significant attention in the field of path-tracking control due to its capability to handle multi-variable states and constraints. In recent years, many researchers have focused on improving the performance of MPC through various means, including optimizing the prediction model, setting multiple constraints, adjusting the prediction horizon, and tuning the weight coefficients of the objective function. Pang et al. \cite{pang2022practical} proposed a comprehensive linear time-varying model predictive controller for vehicle trajectory tracking control. Notably, the above method can be solved with relatively higher computational efficiency and lower computational cost. Geng et al. \cite{geng2023design} designed a neural network predictive controller for vehicle trajectory tracking control. Especially, it relies on a large number of reliable training data. Ji et al. \cite{17} proposed a comprehensive control method that combines path planning and path tracking to optimize the high-speed driving stability of the vehicle. The upper layer generates a collision-free trajectory based on the artificial potential field method, while the lower layer calculates the front wheel angle based on the multi-constrained model predictive control that considers the state constraints of the vehicle’s lateral position, yaw rate, and sideslip angle \cite{gao2022improved,xia2021vehicle}. Li et al. \cite{18} established a comprehensive evaluation index of path-tracking performance to obtain the optimal prediction horizon and control horizon for different vehicle speeds. Based on this, they designed an adaptive dual horizon model predictive controller with constraints on the vehicle's sideslip angle and yaw rate to optimize the vehicle's tracking accuracy and stability.

When the vehicle is driven at low speed, with low adhesion and small curvature, the tires primarily operate in the linear region, making the previously mentioned control methods highly effective for achieving precise tracking. However, under extreme conditions such as high speed and low adhesion, the lateral nonlinear factors of the tires become significantly amplified \cite{xia2016estimation}. This leads to increased uncertainty in the vehicle's state parameters, making it challenging to ensure the vehicle's path-tracking accuracy and stability solely through the control of the front wheel angle. At this point, the vehicle may easily become unstable, experiencing conditions such as side slip and tail flick.

DDEV is an innovative experimental platform for intelligent driving, and its main structural feature is to arrange drive motors in each wheel. This unique configuration enables independent control of torque in all four wheels, thereby enhancing the vehicle's ability to perform yaw moment control \cite{21, zhang2017nonlinear}. To avoid instability of the vehicle under extreme conditions, various coordinated control strategies for path tracking and yaw stability have been proposed, leveraging the independent controllability of the four-wheel torque on DDEV. Liang et al. \cite{22} performed coordinated control of the steering system and the active yaw moment system, where both systems independently calculate the front wheel angle and the active yaw moment. The active yaw moment system adopts a layered architecture, with the upper layer utilizing LTV MPC to calculate the desired yaw rate and the lower layer using sliding mode control to calculate the active yaw moment. Zhang et al. \cite{23} proposed a layered coordinated control strategy for trajectory tracking and direct yaw moment. The upper controller calculates the front wheel angle and additional yaw moment based on MPC, and constrains the tire slip angle and sideslip angle. The lower controller optimizes the distribution of four-wheel drive force with the control target of minimum tire adhesion utilization. Guo et al. \cite{24} proposed a trajectory-tracking method that utilizes a layered architecture. The upper layer employs LTV MPC to calculate the wheel angle and additional yaw moment, while the lower layer distributes the four-wheel torque based on the pseudo-inverse control allocation law. Ren et al. \cite{25} proposed an integrated path-tracking control algorithm. The upper layer calculates the vehicle's wheel angle and required longitudinal total driving force based on nonlinear MPC, and also computes the additional expected yaw moment using the sliding mode control algorithm. The lower layer utilizes the optimal torque vector algorithm to distribute the four-wheel torque.

All of the aforementioned coordinated control strategies have been shown to improve the vehicle's path-tracking capability and stability under extreme conditions. Their fundamental approach is to utilize the additional controllable degrees of freedom provided by DDEV for yaw moment control, thereby improving the vehicle's stability and preventing the vehicle from entering unstable conditions. However, none of these schemes consider optimizing the front wheel angle through adjusting the parameters of MPC while performing yaw moment control, to further achieving the comprehensive optimization of the vehicle's path-tracking accuracy and stability. In fact, both steering control and direct yaw moment control can effectively enhance the vehicle's stability by influencing its lateral response. Compared to a single control method, the coordinated control strategy can further improve the vehicle's stability while ensuring good path-tracking accuracy.

\begin{figure*}[htbp]
\centerline{\includegraphics[width=1.5\columnwidth]{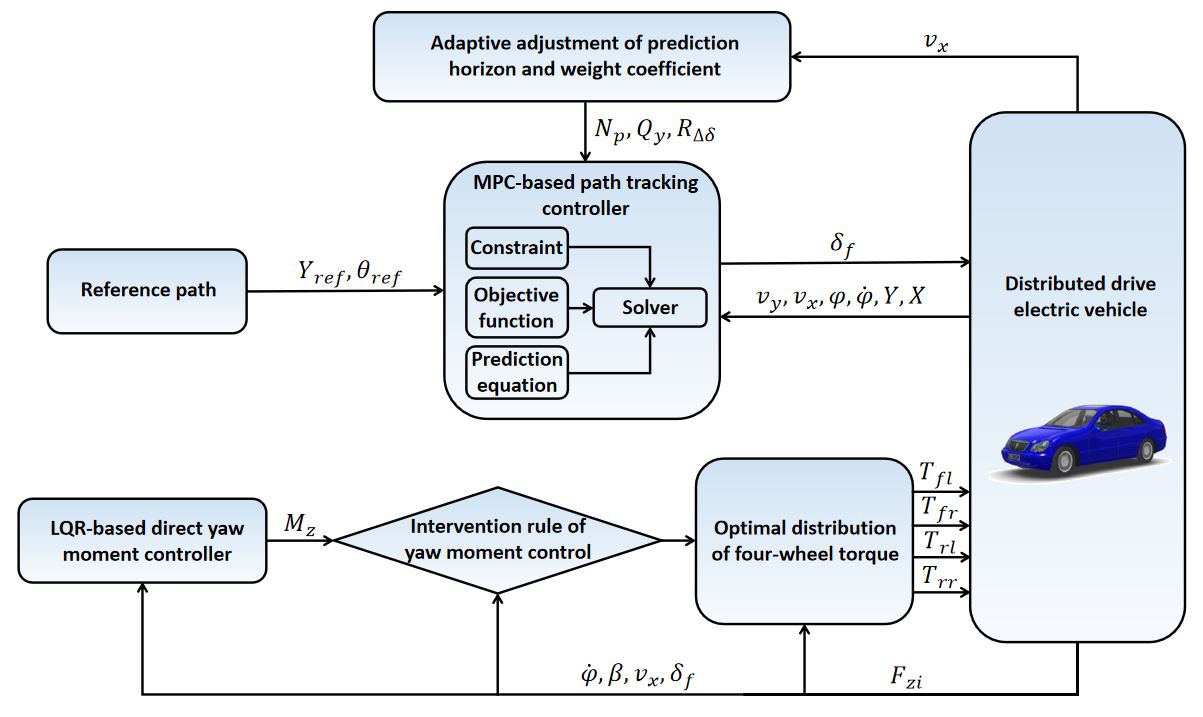}}
\caption{Architecture for coordinated control of path tracking and yaw stability}
\label{fig0}
\end{figure*}

Based on the above analysis, this paper proposes a coordinated control strategy of path tracking and direct yaw moment with a DDEV as the research object. To the best of our knowledge, the main contributions of this work are highlighted as follows (illustrated by Fig. \ref{fig0}).

(1) In order to improve the performance of the path-tracking controller, the effects of prediction horizon and weight coefficients on the path-tracking accuracy and stability of the vehicle are analyzed. An adaptive adjustment of the prediction horizon and weight coefficients based on the variation of vehicle speed is implemented, and a path-tracking controller based on AMPC is proposed.

(2) Based on the optimization of steering control, the stability state of the vehicle is considered and a coordinated control strategy is proposed based on AMPC and DYC. Specifically, when the vehicle is stable, AMPC coordinates the path-tracking accuracy and stability of the vehicle at different speeds. At low speeds, it prioritizes improving the path-tracking accuracy; while at high speeds, it focuses on enhancing vehicle stability. However, when the vehicle is nearly unstable, AMPC and DYC work together to improve vehicle stability while maintaining good path-tracking accuracy, utilizing the intervention signal from DYC.

\section{Vehicle dynamics model}

\begin{figure}[htbp]
\centerline{\includegraphics[width=\columnwidth]{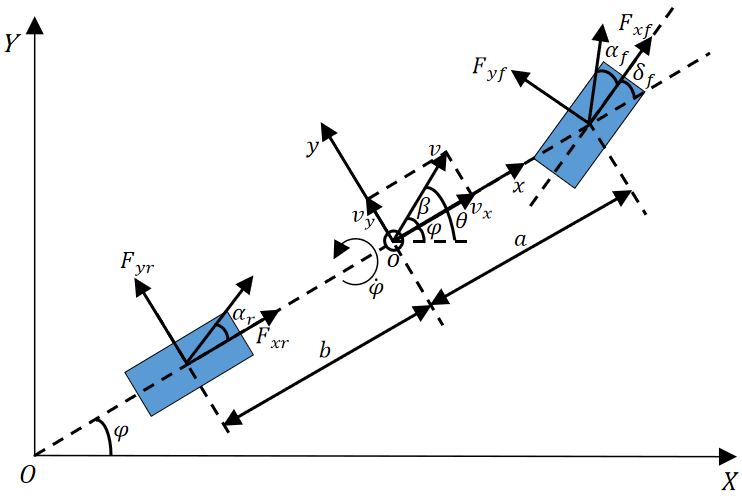}}
\caption{Vehicle 3-DOF monorail dynamics model}
\label{fig1}
\end{figure}

To enable the vehicle to track the reference path quickly and stably, the vehicle 3-DOF monorail dynamics model shown in Fig. \ref{fig1}. is established as the prediction model for the path-tracking controller. The vehicle model involves two coordinate systems, where $XOY$ is the global coordinate system and $xoy$ is the vehicle body-fixed coordinate system. Compared to the kinematics model, the vehicle dynamics model considers the forces from the tires and the mass of vehicle, as well as the vehicle's dynamic characteristics under conditions of high speed and large curvature, which improves the control accuracy of the model predictive controller \cite{chen2019dynamics}.
A force analysis is performed on the vehicle dynamics model. According to its force balance and moment balance \cite{liu2020vision, xia2022autonomous, chen2019comprehensive}, the differential equation of the system can be established as follows:
\begin{equation}
\left\{\begin{array}{c}
m\left(\dot{v}_x-v_y \dot{\varphi}\right)=F_{x f}+F_{x r} \\
m\left(\dot{v}_y+v_x \dot{\varphi}\right)=F_{y f}+F_{y r} \\
I_z \ddot{\varphi}=a F_{y f}-b F_{y r}
\end{array}\right.
\label{eq1}
\end{equation}
where $m$ is the vehicle's curb weight, $v_x$, $v_y$ and $\dot{\varphi}$ are the longitudinal velocity, lateral velocity and yaw rate of the vehicle at the center of mass in the vehicle body-fixed coordinate system, $I_z$ is the moment of inertia of the vehicle, $a$ and $b$ are the distances from the center of mass of the vehicle to the front and rear axles, $F_{y f}$ and $F_{y r}$ are the lateral forces on the front and rear axles, and $F_{x f}$ and $F_{x f}$ are the longitudinal tire forces on the front and rear axles. 

From Fig. \ref{fig1}, it can be seen that the sideslip angle $\beta$ is the angle between the vehicle's longitudinal axis and the velocity direction at the center of mass, and it can be approximately calculated by $v_y / v_x$ \cite{chen2023planning}, so the above formula can be approximately expressed as:
\begin{equation}
m v_x(\dot{\beta}+\dot{\varphi})=F_{y f}+F_{y r}
\label{eq2}
\end{equation}
The tire's lateral force is closely related to the motion state of the vehicle. Considering the nonlinear dynamics of the tire can improve the simulation accuracy of the vehicle's state response under high-speed and large-curvature conditions. The brush tire model utilizes physical parameters to describe the characteristics of the tire, which offers the advantages of simplicity and accuracy. Therefore, the brush tire model is employed to calculate the tire's lateral force. The formula for calculating the tire's lateral force is as follows:
\begin{equation}
F_y=\left\{\begin{array}{c}
-C_\alpha \tan (\alpha)+\frac{c_\alpha{ }^2}{3 \mu F_z}|\tan (\alpha)| \tan (\alpha)-\frac{c_\alpha{ }^3}{27 \mu^2 F_z{ }^2} \tan ^3(\alpha), \\
|\alpha|<\arctan \left(\frac{3 \mu F_z}{c_\alpha}\right) \\
-\mu F_z \operatorname{sgn}(\alpha),|\alpha| \geq \arctan \left(\frac{3 \mu F_z}{c_\alpha}\right)
\end{array}\right.
\label{eq3}
\end{equation}
where $\alpha$ is the slip angle of the tire, $C_\alpha$ is the linear cornering stiffness of the tire, $\mu$ is the road adhesion coefficient of the tire and $F_z$ is the vertical force of the tire. 

Since Eq. (\ref{eq3}) for the tire's lateral force is too complicated to be directly applied to calculate the tire's lateral force in MPC, Eq. (\ref{eq3}) is simplified. The simplified equation for calculating the tire's lateral force is as follows:
\begin{equation}
\left\{\begin{array}{l}
F_{y f}=C_{\alpha f} \alpha_f \\
F_{y r}=C_{\alpha r} \alpha_r
\end{array}\right.
\label{eq4}
\end{equation}
where $C_{\alpha f}$ and $C_{\alpha r}$ are the cornering stiffness of the front and rear tires, $\alpha_f$ and $\alpha_r$ are the cornering angles of the front and rear tires. During the path-tracking process, the tire's cornering stiffness is corrected based on the brush tire model mentioned above. The cornering angle of the front and rear tires can be approximately calculated by the following formula:
\begin{equation}
\left\{\begin{array}{c}
\alpha_f=\frac{v_y+a \dot{\varphi}}{v_x}-\delta_f \\
\alpha_r=\frac{v_y-b \dot{\varphi}}{v_x}
\end{array}\right.
\label{eq5}
\end{equation}
Given the small values of the vehicle's steering angle and the tire's slip ratio, the tire's longitudinal force can be approximated as a linear function of the tire's slip ratio. The calculation formula for the tire's longitudinal force is as follows:
\begin{equation}
\left\{\begin{array}{l}
F_{x f}=C_{l f} s_f \\
F_{x r}=C_{l r} s_r
\end{array}\right.
\label{eq6}
\end{equation}
where $C_{l f}$ and $C_{l r}$ are the longitudinal stiffness of the front and rear tires, $s_f$ and $s_r$ are the slip ratios of the front and rear tires. Combining Eq. (\ref{eq1}), Eq. (\ref{eq4}), Eq. (\ref{eq5}) and Eq. (\ref{eq6}), a nonlinear model of vehicle dynamics is developed under the assumption of small angle. The expressions are as follows:
\begin{equation}
\left\{\begin{array}{c}
\dot{v}_y=-v_x \dot{\varphi}+\frac{2}{m}\left[C_{\alpha f}\left(\frac{v_y+a \dot{\varphi}}{v_x}-\delta_f\right)+C_{\alpha r} \frac{v_y-b \dot{\varphi}}{v_x}\right] \\
\dot{v}_x=v_y \dot{\varphi}+\frac{2}{m}\left[C_{l f} s_f+C_{\alpha f}\left(\frac{v_y+a \dot{\varphi}}{v_x}-\delta_f\right) \delta_f+C_{l r} s_r\right] \\
\ddot{\varphi}=\frac{2}{I_z}\left[a C_{\alpha f}\left(\frac{v_y+a \dot{\varphi}}{v_x}-\delta_f\right)-b C_{\alpha r} \frac{v_y-b \dot{\varphi}}{v_x}\right] \\
\dot{Y}=v_x \sin \varphi+v_y \cos \varphi \\
\dot{X}=v_x \cos \varphi-v_y \sin \varphi \\
\dot{\beta}=\frac{2}{m v_x}\left[C_{\alpha f}\left(\frac{v_y+a \dot{\varphi}}{v_x}-\delta_f\right)+C_{\alpha r} \frac{v_y-b \dot{\varphi}}{v_x}\right]-\dot{\varphi}
\end{array}\right.
\label{eq7}
\end{equation}

\section{Path-tracking controller with linear time-varying MPC}
\subsection{Linearization and discretization of nonlinear system}
Eq. (\ref{eq7}) can be described as the following state space equation:
\begin{equation}
\dot{X}(t)=f(X(t), u(t))
\label{eq8}
\end{equation}
where the system's state quantity is chosen as $X=\left[v_y, v_x, \varphi, \dot{\varphi}, Y, X, \beta\right]^T$, the system's control quantity is chosen as $u=\delta_f$. It can be seen from Fig. \ref{fig0} that the vehicle's heading angle $\theta=\varphi+\beta$ \cite{liu2021automated, chen2023dynamic}, so the system's output quantity is chosen as $\eta=[\theta, Y]^T$. 

Expanding Eq. (\ref{eq8}) at the system point $(X(t), u(t))$ with Taylor series and ignoring the higher order terms, we can obtain the following equation:
\begin{equation}
\dot{X}=f(X(t), u(t))+A(t)(X-X(t))+B(t)(u-u(t))
\label{eq9}
\end{equation}
Combining Eq. (\ref{eq8}) and Eq. (\ref{eq9}), we can obtain the following equation:
\begin{equation}
\dot{(X-X(t))}=A(t)(X-X(t))+B(t)(u-u(t))
\label{eq10}
\end{equation}
where $A(t)=\left.\frac{\partial f}{\partial X}\right|_{X(t), u(t)}$, $B(t)=\left.\frac{\partial f}{\partial u}\right|_{X(t), u(t)}$.

The first-order difference quotient method is used to discretize Eq. (\ref{eq10}), and the discrete state space expression is obtained as follows:
\begin{equation}
\left\{\begin{array}{c}
X(k+1)=A(k) X(k)+B(k) u(k) \\
\eta(k)=C(k) X(k)
\end{array}\right.
\label{eq11}
\end{equation}
where $A(k)=I+T A(t)$, $B(k)=T B(t)$, $C(k)=\left[\begin{array}{lllllll}0 & 0 & 1 & 0 & 0 & 0 & 1 \\ 0 & 0 & 0 & 0 & 1 & 0 & 0\end{array}\right]$, $I$ is the unit matrix, and $T$ is the sampling time period of the MPC controller.

To improve the continuity of the system's control quantity and avoid sudden changes in the control quantity, the control increment $\Delta u(t)$ is used instead of the control quantity $u(t)$ to limit the control increment within each sampling period. The state matrix is transformed as follows:
\begin{equation}
\widetilde{X}(k \mid t)=\left[\begin{array}{c}
X(k \mid t) \\
u(k-1 \mid t)
\end{array}\right]
\label{eq12}
\end{equation}
A new state space expression is obtained as follows:
\begin{equation}
\left\{\begin{array}{c}
\widetilde{X}(k+1 \mid t)=\widetilde{A}_{k, t} \widetilde{X}(k \mid t)+\widetilde{B}_{k, t} \Delta u(k \mid t) \\
\eta(k \mid t)=\widetilde{C}_{k, t} \widetilde{X}(k \mid t)
\end{array}\right.
\label{eq13}
\end{equation}
where $\widetilde{A}_{k, t}=\left[\begin{array}{cc}A(k) & B(k) \\ 0_{m \times n} & I_m\end{array}\right]$, $\widetilde{B}_{k, t}=\left[\begin{array}{c}B(k) \\ I_m\end{array}\right]$, $\widetilde{C}_{k, t}=\left[\begin{array}{ll}C(k) & 0_{2 \times m}\end{array}\right]$, $n$ is the dimension of state quantity and $m$ is the dimension of control quantity.

Eq. (\ref{eq13}) is the linearized and discretized control system at point $(X(t), u(t))$, which can be used to design a linear time-varying model predictive path-tracking controller.

\subsection{State prediction equation and construction of output model}
The control horizon and prediction horizon of the model predictive controller are set as $N_c$ and $N_p$. The output sequence of the control system within the prediction horizon can be expressed as:
 \begin{equation}
\widetilde{Y}(t)=\Psi_t \widetilde{X}(t \mid t)+\Theta_t \Delta U(t)
\label{eq14}
\end{equation}
where $\widetilde{Y}(t)=\left[\begin{array}{llll}\eta(t+1 \mid t) & \eta(t+2 \mid t) & \cdots & \eta\left(t+N_p \mid t\right)\end{array}\right]^T$, 
$\Psi_t=\left[\begin{array}{llll}\widetilde{C}_t \widetilde{A}_t & \widetilde{C}_t \widetilde{A}_t^2 & \cdots & \widetilde{C}_t \widetilde{A}_t^{N_p}\end{array}\right]^T$, 
$\Delta U(t)=\left[\begin{array}{llll}\Delta u(t \mid t) & \Delta u(t+1 \mid t) & \cdots & \Delta u\left(t+N_c-1 \mid t\right)\end{array}\right]^T$, 
$\Theta_t=\left[\begin{array}{cccc}\widetilde{C}_t \widetilde{B}_t & 0 & \cdots & 0 \\ \widetilde{C}_t \widetilde{A}_t \widetilde{B}_t & \widetilde{C}_t \widetilde{B}_t & \cdots & 0 \\ \vdots & \vdots & \ddots & \vdots \\ \widetilde{C}_t \widetilde{A}_t^{N_p-1} \widetilde{B}_t & \widetilde{C}_t \widetilde{A}_t^{N_p-2} \widetilde{B}_t & \cdots & \widetilde{C}_t \widetilde{A}_t^{N_p-N_c-1} \widetilde{B}_t\end{array}\right]$.

\subsection{Objective function and constraints}
The objective of the path-tracking control is to ensure that the intelligent driving vehicle tracks the reference path quickly and smoothly. Therefore, the optimization of the deviation of the system's output state quantity and the control increment needs to be considered in the objective function of the controller. The objective function is expressed as follows:
\begin{equation}
\begin{aligned}
J(\widetilde{X}(t), u(t-1), \Delta U(t)) & =\sum_{i=1}^{N_p}\left\|\eta(t+i \mid t)-\eta_{r e f}(t+i \mid t)\right\|_Q^2 \\
& +\sum_{i=0}^{N_c-1}\|\Delta u(t+i \mid t)\|_R^2+\rho \varepsilon^2
\end{aligned}
\label{eq15}
\end{equation}
where the first term of the objective function represents the tracking error of the vehicle's lateral position and heading angle, and the second term represents the variation of the vehicle's front wheel angle. The first term reflects the controller's ability to track the reference path, while the second term reflects the constraint on control increment. Due to the complexity of the vehicle dynamics model, the controller may not obtain the optimal solution, so a relaxation factor is introduced in the system to ensure that the controller has a solution. The weight matrices $Q$ and $R$ are used in the objective function, and their specific forms are shown in Eq. (\ref{eq16}) and Eq. (\ref{eq17}):
\begin{equation}
Q=\left[\begin{array}{cccccc}
Q_1 & 0 & \cdots & 0 & \cdots & 0 \\
0 & Q_2 & \cdots & 0 & \cdots & 0 \\
\vdots & \vdots & \ddots & \vdots & \vdots & \vdots \\
0 & 0 & \cdots & Q_{N_c} & \cdots & 0 \\
\vdots & \vdots & \vdots & \vdots & \ddots & \vdots \\
0 & 0 & \cdots & 0 & \cdots & Q_{N_p}
\end{array}\right]
\label{eq16}
\end{equation}
where $Q_i=\left[\begin{array}{cc}Q_\theta & 0 \\ 0 & Q_y\end{array}\right]$, $i=1,2, \ldots N_p$, $Q_\theta$ is the weight coefficient of heading tracking error, and $Q_y$ is the weight coefficient of lateral tracking error. 
\begin{equation}
R=\left[\begin{array}{cccc}
R_0 & 0 & \cdots & 0 \\
0 & R_1 & \cdots & 0 \\
\vdots & \vdots & \ddots & \vdots \\
0 & 0 & \cdots & R_{N_c-1}
\end{array}\right]
\label{eq17}
\end{equation}
where $R_i=R_{\Delta \delta}$, $\quad i=1,2, \ldots, N_c-1$, and $R_{\Delta \delta}$ is the weight coefficient of control increment.

During the process of path tracking, the front wheel angle or its variation being too large can negatively impact the vehicle's stability and safety. Therefore, it is essential to limit the control quantity and control increment of the model predictive controller. The path-tracking controller based on the dynamic model needs to solve the following optimization problem, given the objective function and constraints:
\begin{equation}
\left\{\begin{array}{c}
\min _{\Delta U, \varepsilon} J(\widetilde{X}(t), u(t-1), \Delta U(t)) \\
\text { s.t. } \Delta U_{\min } \leq \Delta U(t) \leq \Delta U_{\max } \\
\qquad\qquad U_{\min } \leq A \Delta U(t)+U(t) \leq U_{\max }
\end{array}\right.
\label{eq18}
\end{equation}
where $U(t)=\left[\begin{array}{llll}u(t \mid t) & u(t \mid t) & \cdots & u(t \mid t)\end{array}\right]_{N_c \times 1}^T$.

\section{MPC with variable prediction horizon and weight coefficients}
\subsection{Influence of different prediction horizons and weight coefficients on tracking performance}
\begin{figure*}[h]
   \centering
\subfigure[Tracking path under different $N_p$]{
   \includegraphics[width=0.32\linewidth]{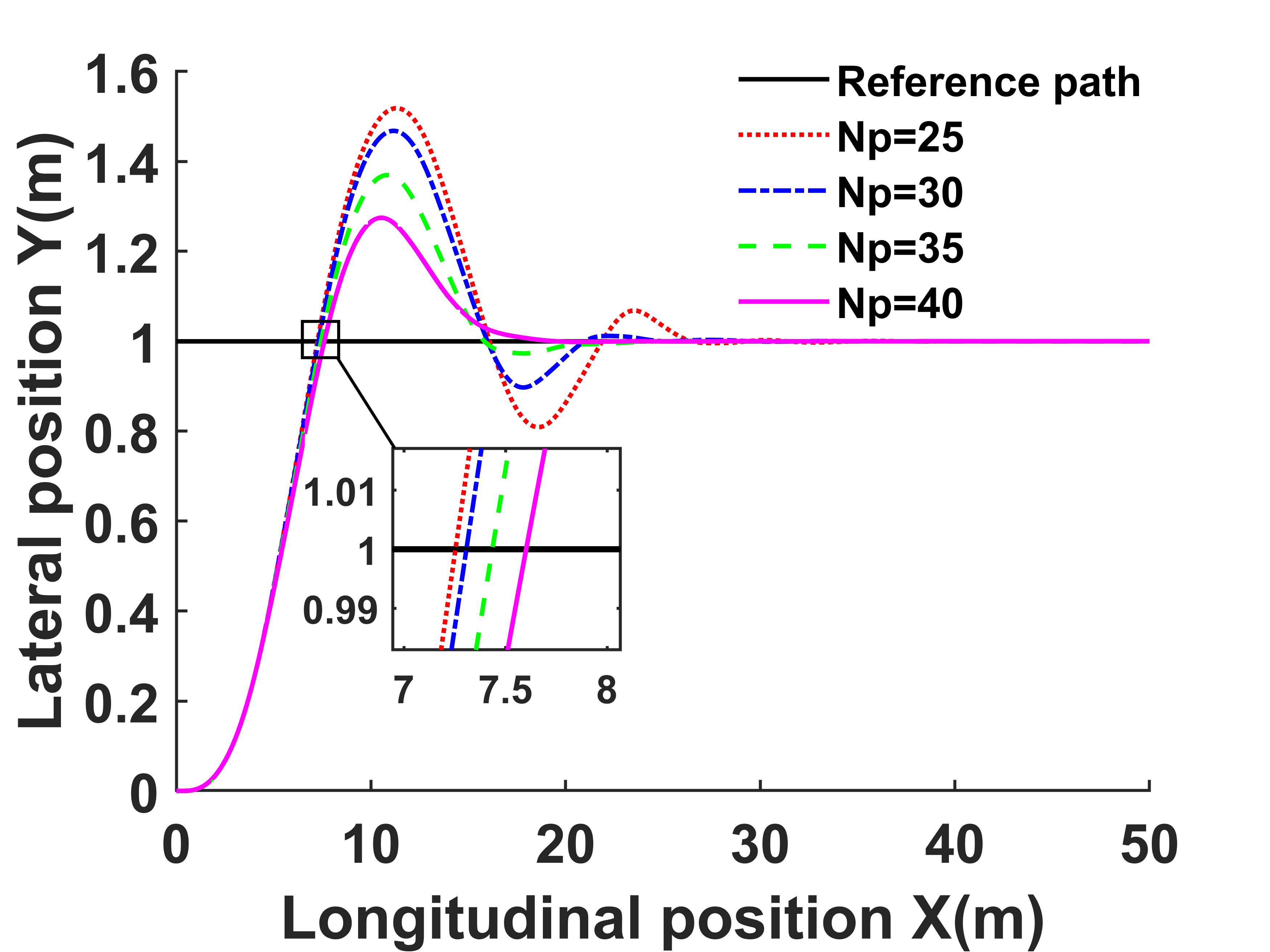}
   }
   \subfigure[Tracking path under different $Q_y$]{
   \includegraphics[width=0.32\linewidth]{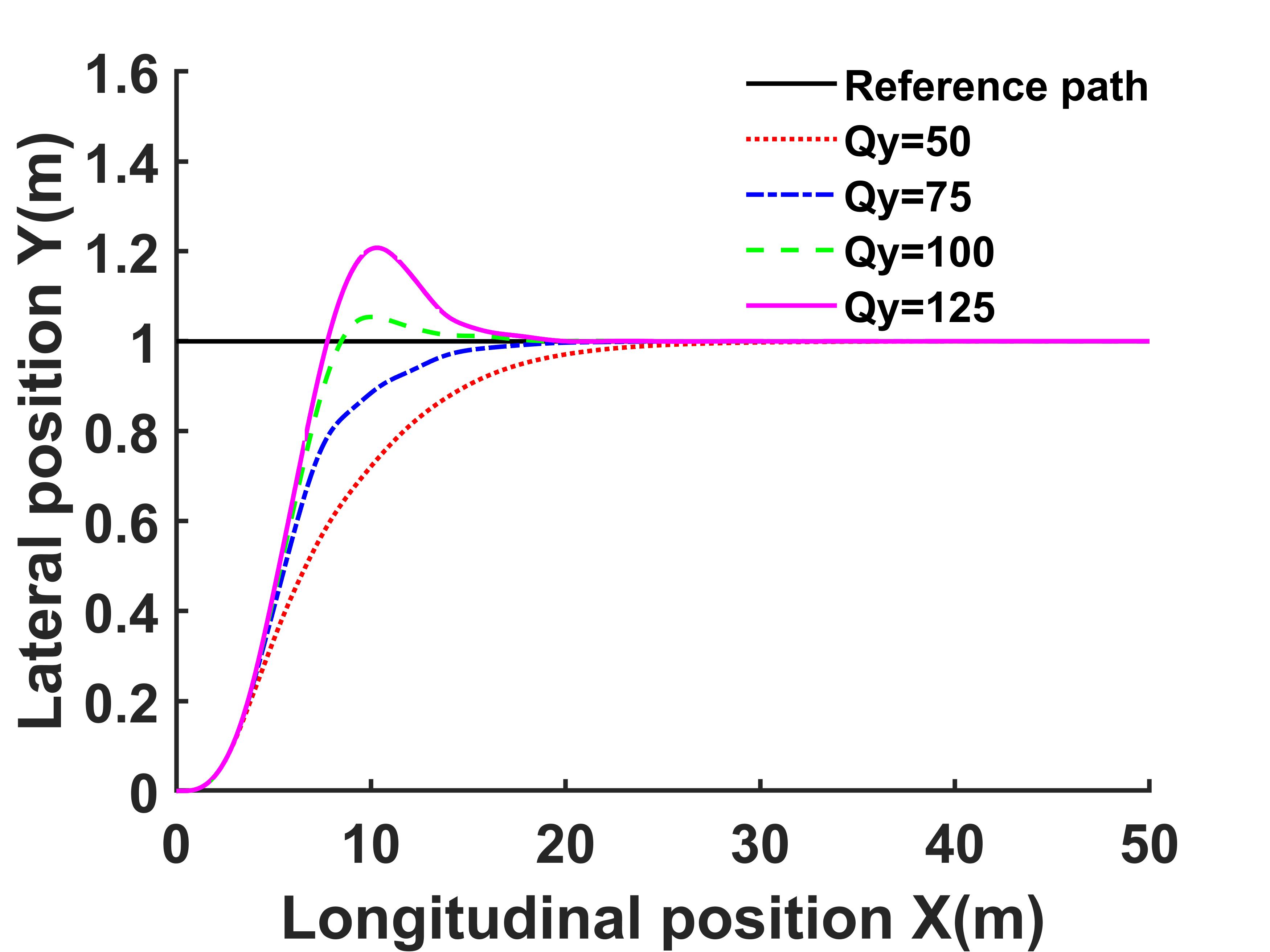}
   }
   \subfigure[Lateral error under different $R_{\Delta \delta}$]{
   \includegraphics[width=0.32\linewidth]{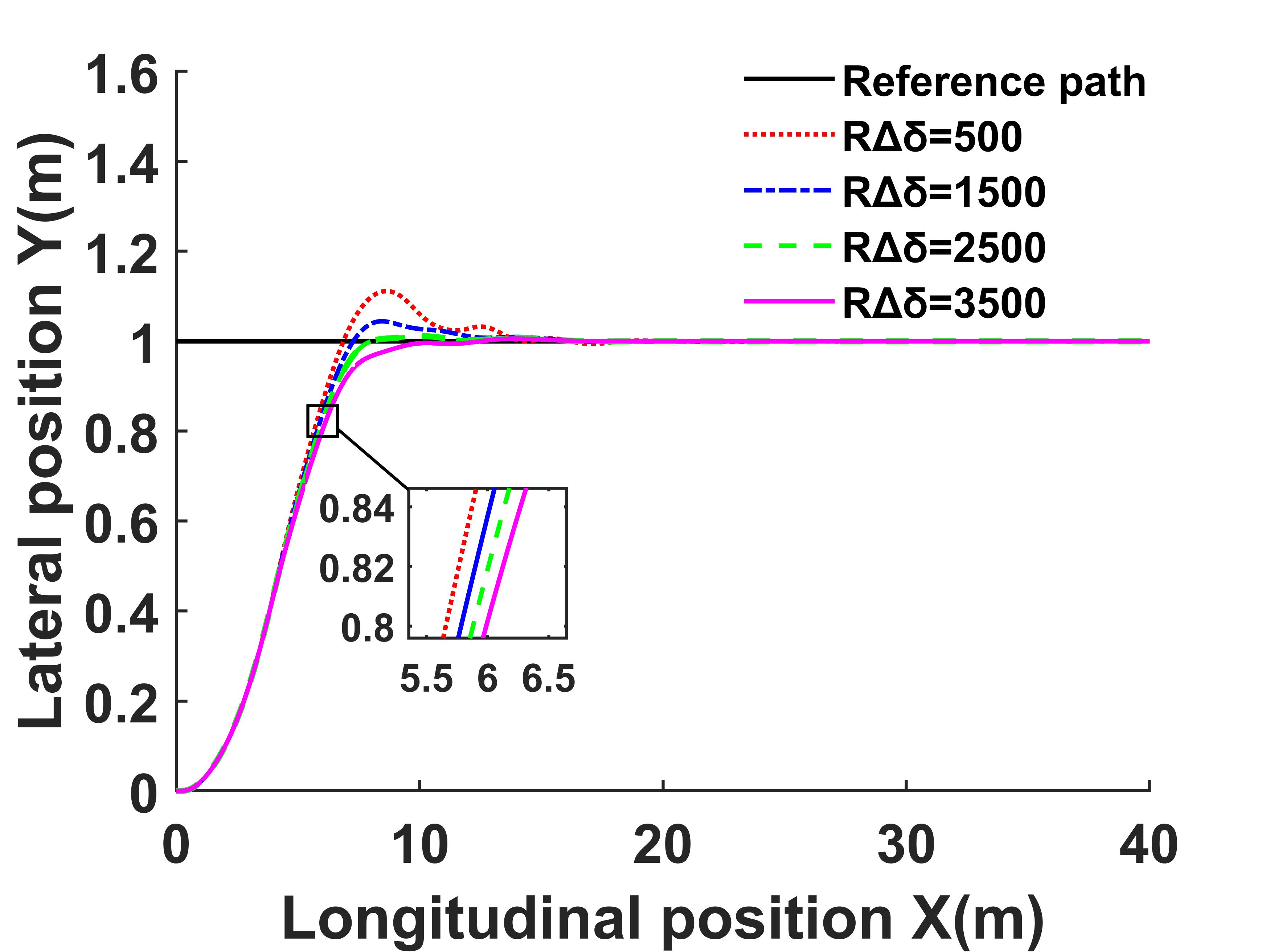}
   }
   \subfigure[Lateral error under different $N_p$]{
   \includegraphics[width=0.32\linewidth]{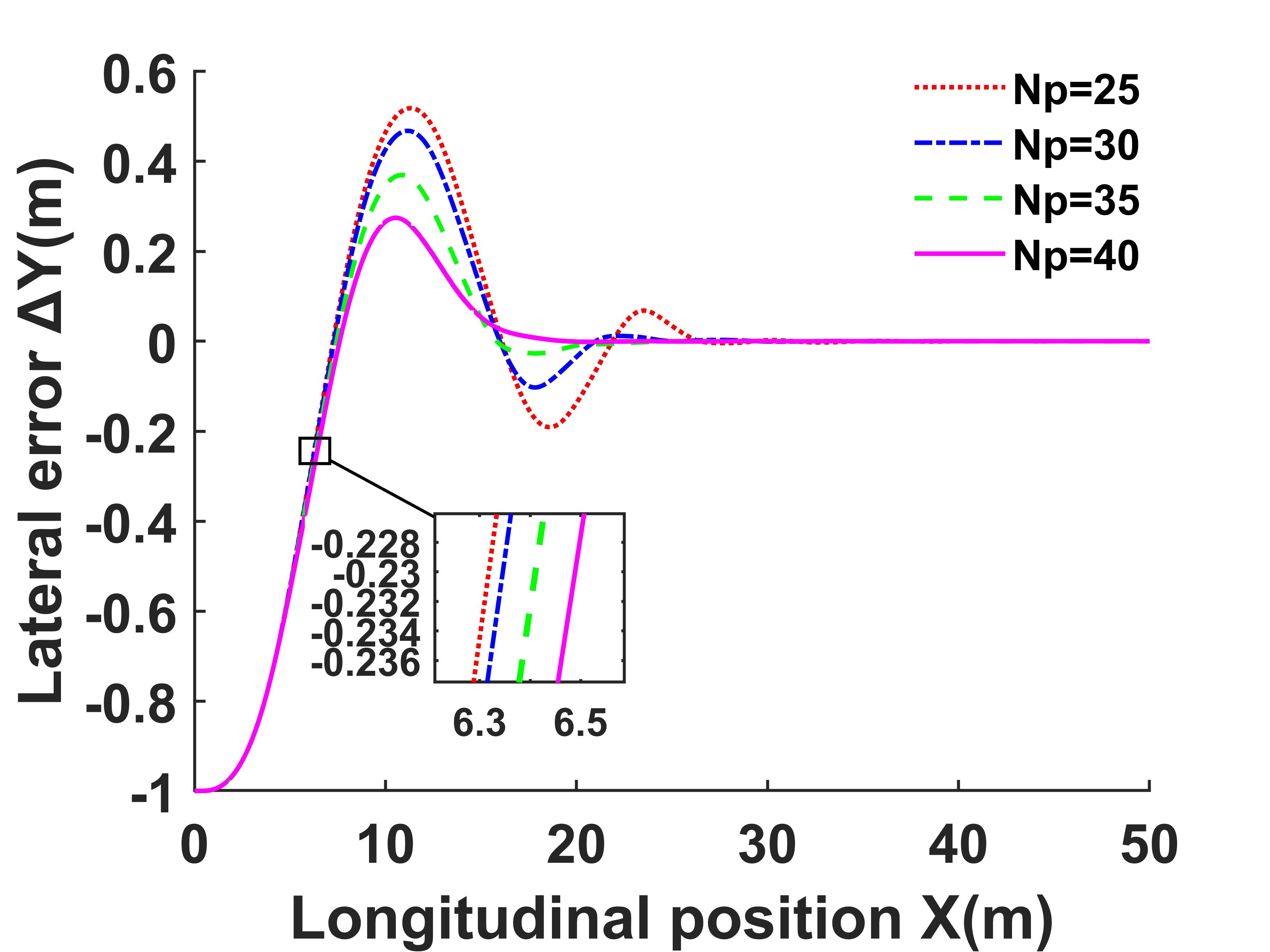}
   }
   \subfigure[Lateral error under different $Q_y$]{
   \includegraphics[width=0.32\linewidth]{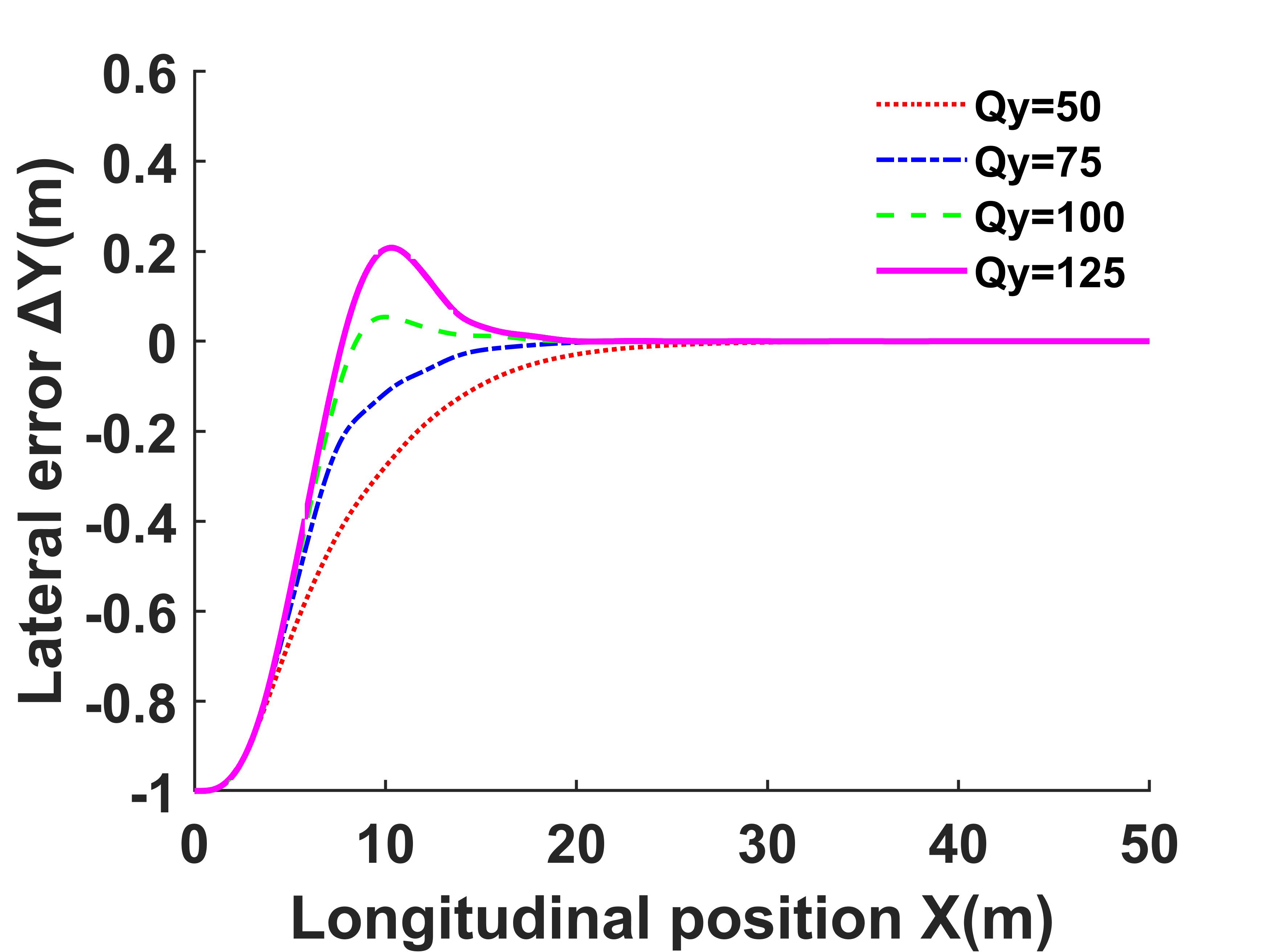}
   }
   \subfigure[Lateral error under different $R_{\Delta \delta}$]{
   \includegraphics[width=0.32\linewidth]{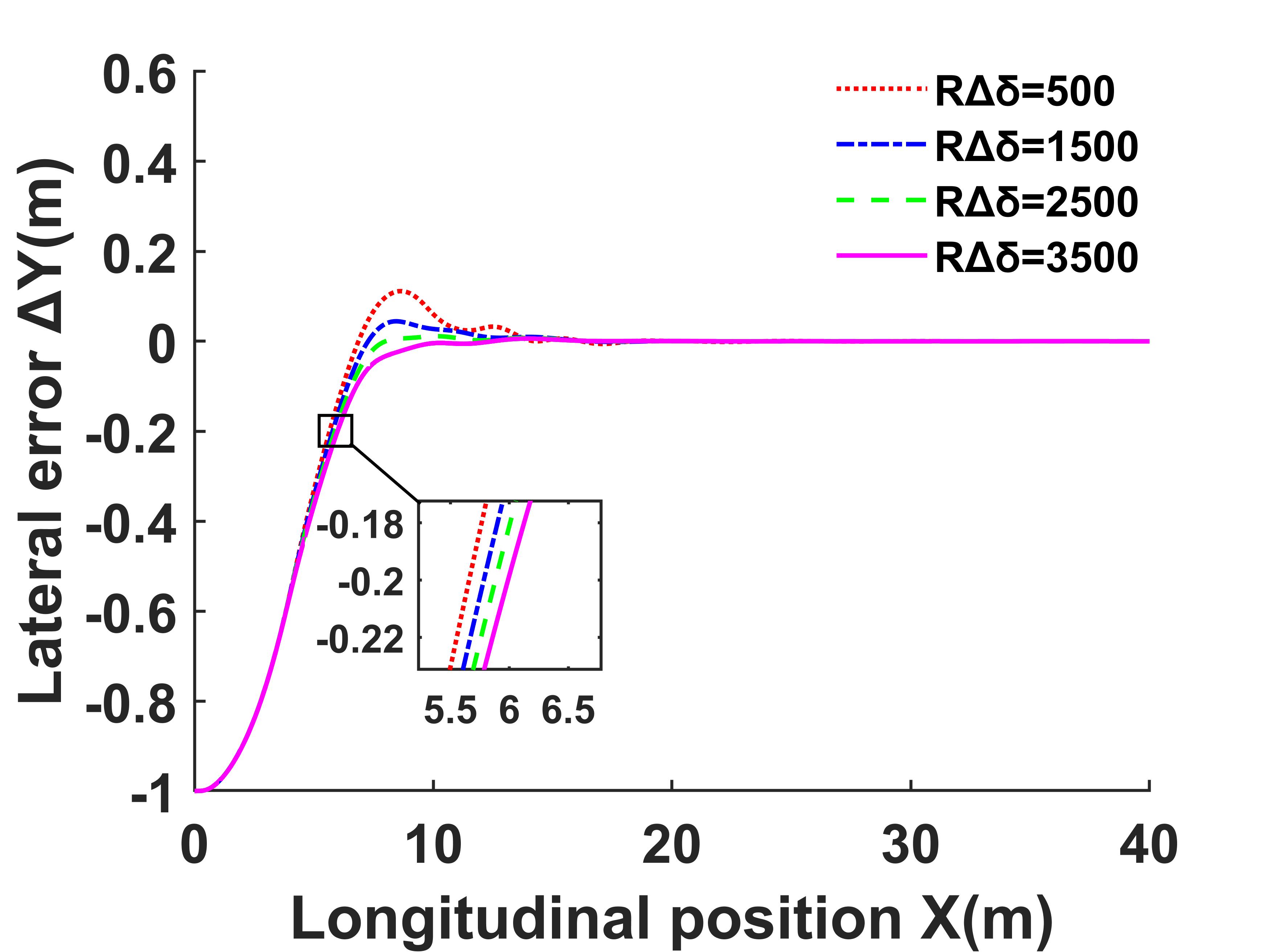}
   }
   \caption{Influence of $N_p$, $Q_y$ and $R_{\Delta \delta}$ on path-tracking accuracy}
   \label{fig2}
\end{figure*}

\begin{figure*}[h]
   \centering
\subfigure[Yaw rate  under different $N_p$]{
   \includegraphics[width=0.32\linewidth]{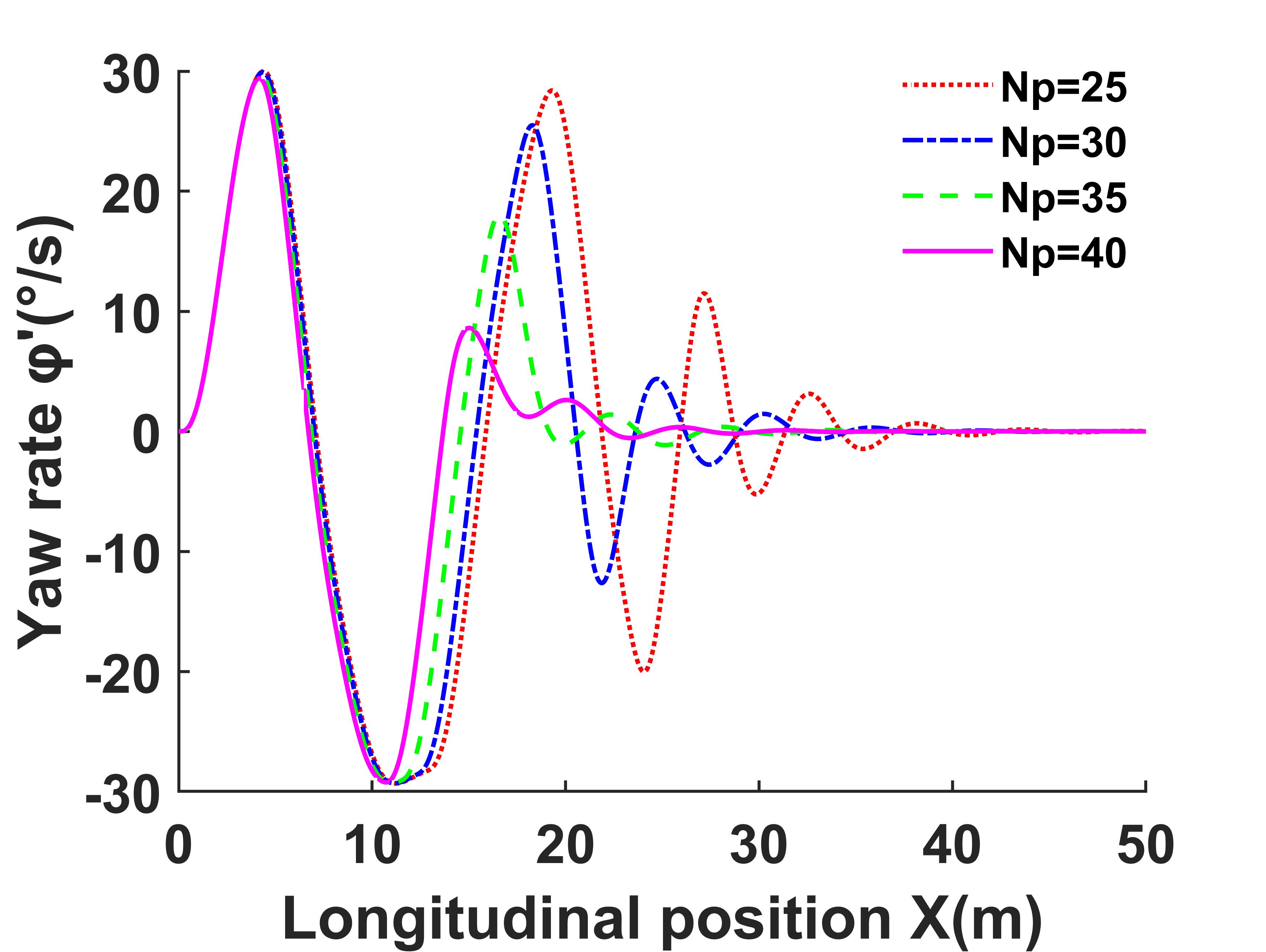}
   }
   \subfigure[Yaw rate under different $Q_y$]{
   \includegraphics[width=0.32\linewidth]{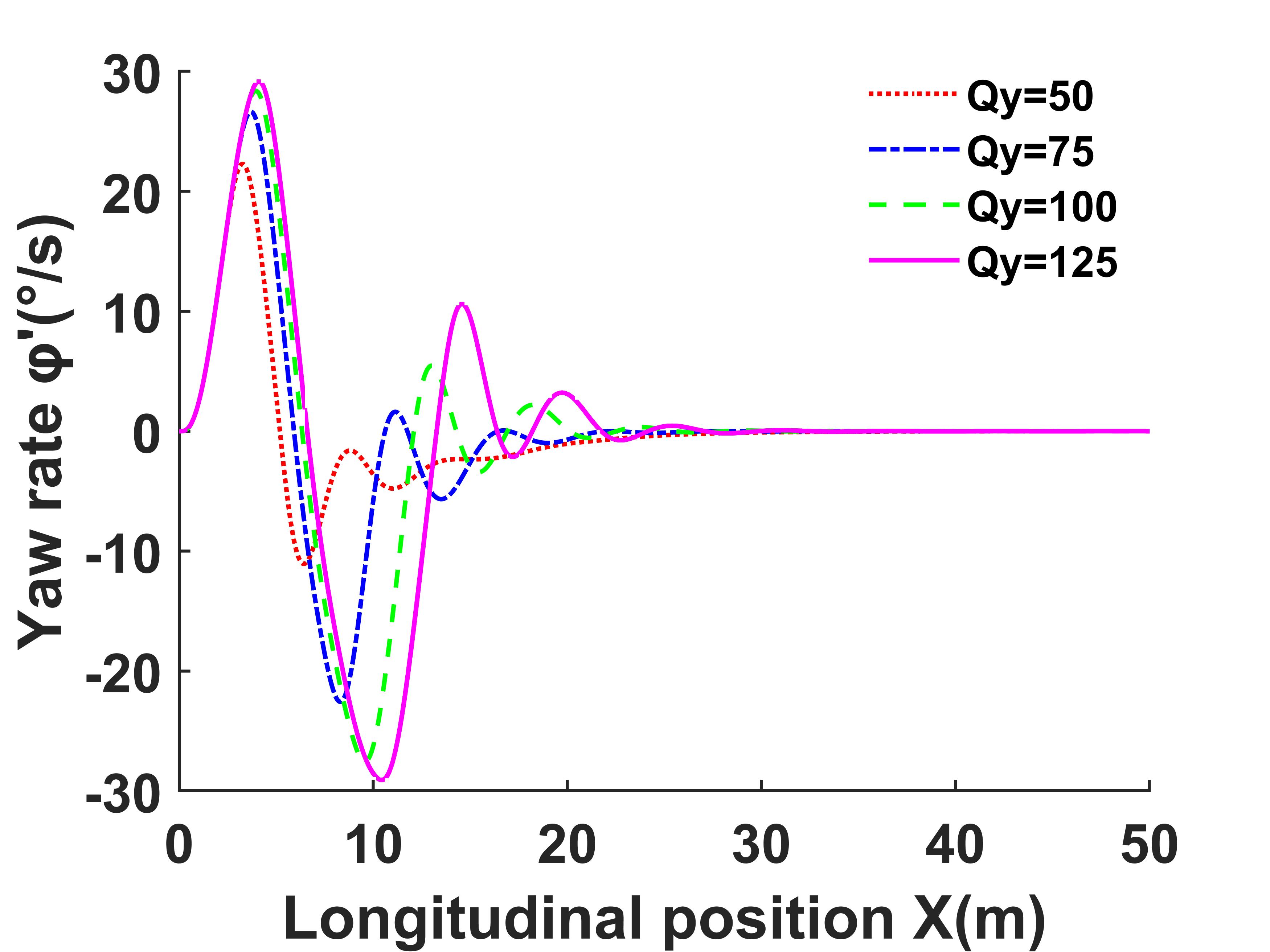}
   }
   \subfigure[Yaw rate under different $R_{\Delta \delta}$]{
   \includegraphics[width=0.32\linewidth]{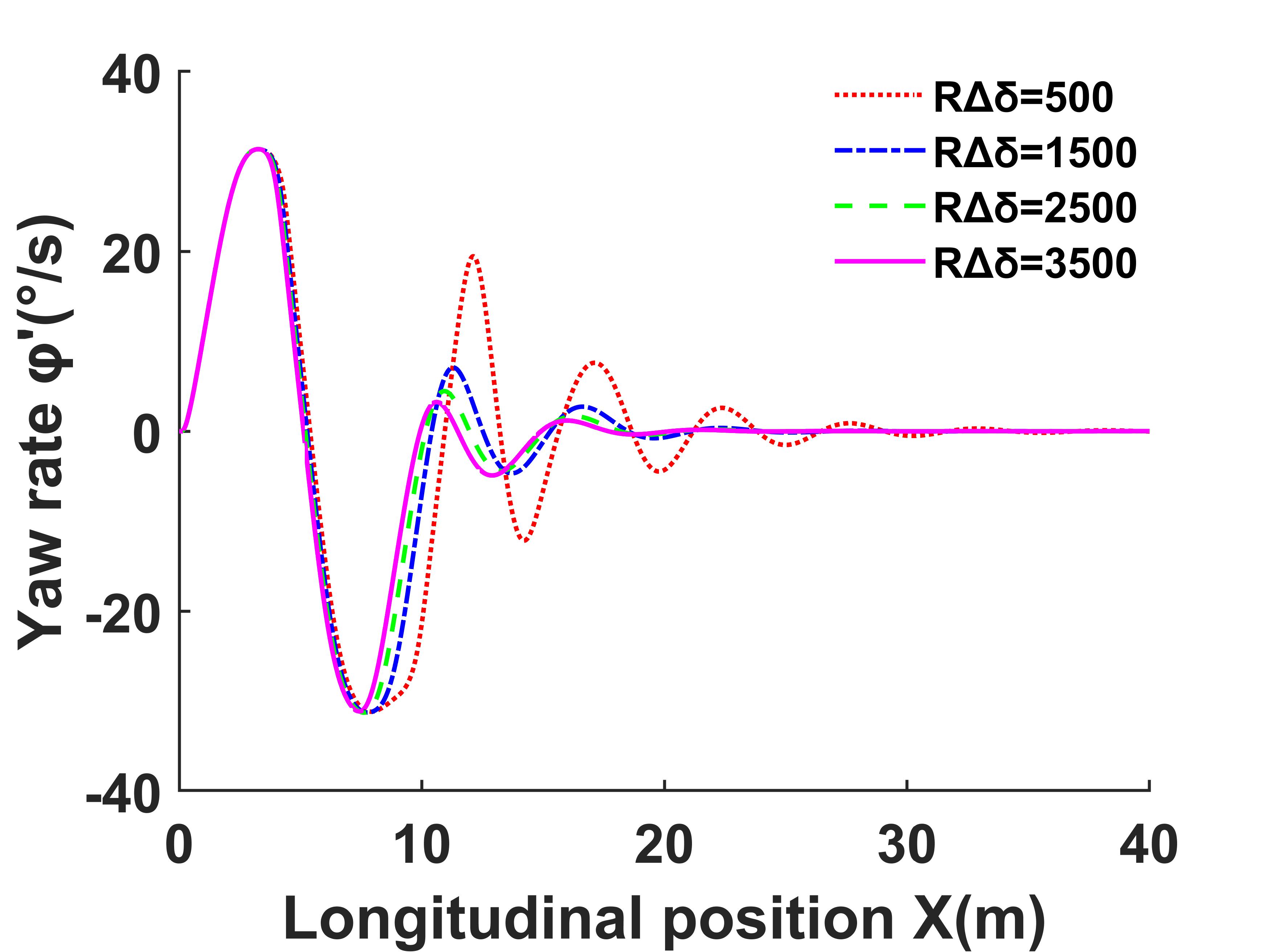}
   }
   \subfigure[Sideslip angle under different $N_p$]{
   \includegraphics[width=0.32\linewidth]{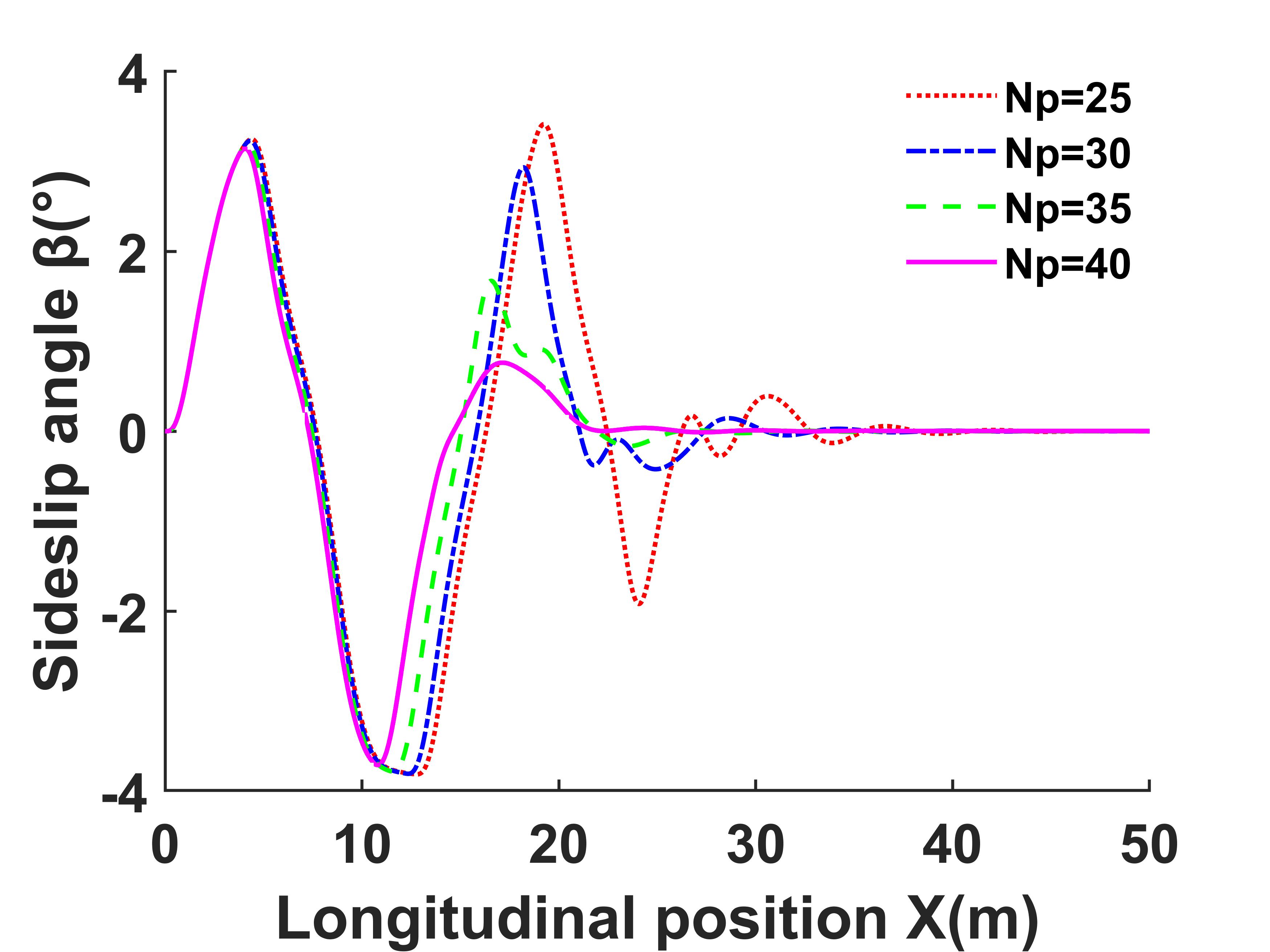}
   }
   \subfigure[Sideslip angle under different $Q_y$]{
   \includegraphics[width=0.32\linewidth]{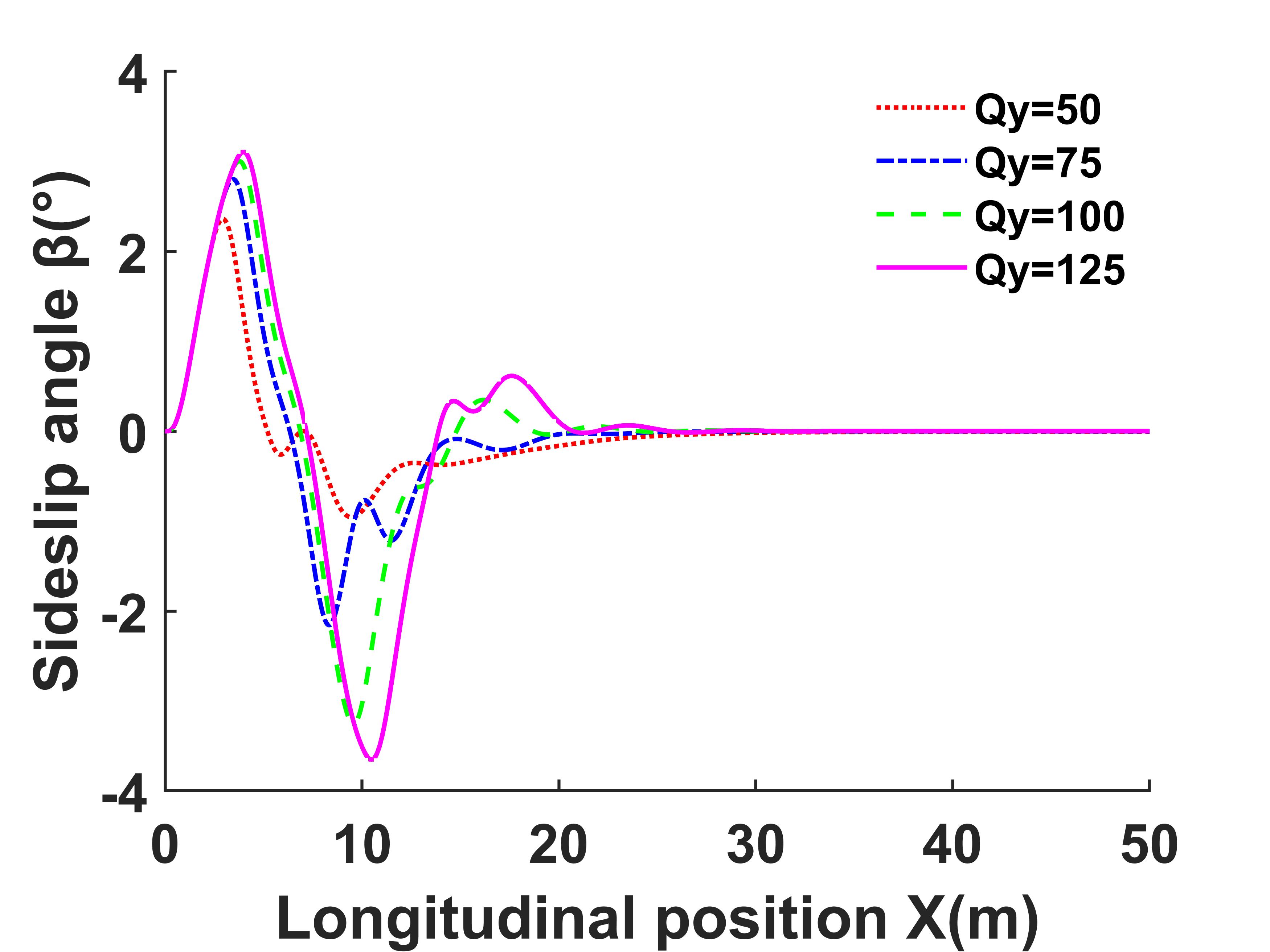}
   }
   \subfigure[Sideslip angle under different $R_{\Delta \delta}$]{
   \includegraphics[width=0.32\linewidth]{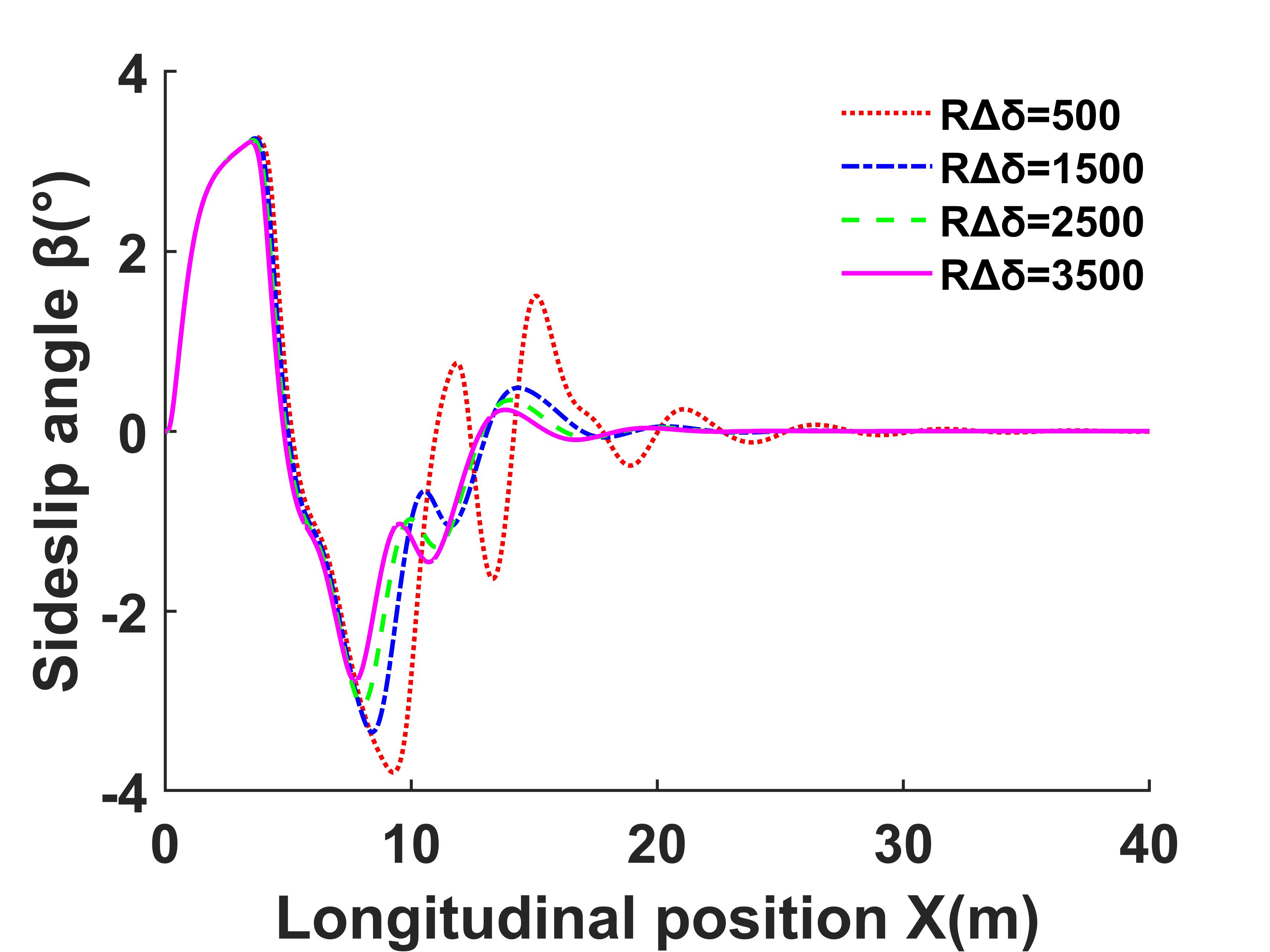}
   }
   \caption{Influence of $N_p$, $Q_y$, $R_{\Delta \delta}$ on vehicle stability}
   \label{fig3}
\end{figure*}

The prediction horizon $N_p$, the weight coefficient of lateral tracking error $Q_y$ and the weight coefficient of control increment $R$ are three crucial parameters of MPC that directly impact the vehicle's path-tracking accuracy and stability. To analyze the influence of these parameters on the vehicle's path-tracking accuracy and stability, a CarSim/Simulink co-simulation platform is built. The simulation is conducted under the condition of a vehicle speed of 30 km/h and a road adhesion coefficient of 0.8, while the reference path is set as a straight line expressed as $Y=1$. The following three sets of simulation data are designed for comparative analysis, and the simulation results are shown in Fig. \ref{fig2} and Fig. \ref{fig3}.

(1) The prediction horizon $N_p$ is set to 25, 30, 35, and 40, respectively, while the weight coefficient of lateral tracking error $Q_y$, weight coefficient of heading tracking error $Q_\theta$ and weight coefficient of control increment $R_{\Delta \delta}$ are kept constant.

(2) The weight coefficient of lateral tracking error $Q_y$ is set to 50, 75,100, and 125, respectively, while the prediction horizon $N_p$, weight coefficient of heading tracking error $Q_\theta$, and weight coefficient of control increment $R_{\Delta \delta}$ are kept constant.

(3) The weight coefficient of control increment $R_{\Delta \delta}$ is set to 500, 1500, 2500 and 3500, respectively, while the prediction horizon $N_p$, weight coefficient of lateral tracking error $Q_y$ and weight coefficient of heading tracking error $Q_\theta$ are kept constant.

The influence of $N_p$, $Q_y$ and $R_{\Delta \delta}$ on the path-tracking accuracy is shown in Fig. \ref{fig2}. It can be seen from Fig. \ref{fig2} that the smaller the prediction horizon $N_p$, the larger the weight coefficient $Q_y$, and the smaller the weight coefficient $R_{\Delta \delta}$, the faster the response speed of path-tracking control. However, the steering controller is more prone to overshoot, resulting in increased lateral tracking error due to the overshoot phenomenon. This is because the smaller the $N_p$, the shorter the driving distance predicted by the MPC, causing the controller to make adjustments later for the possible overshoot phenomenon. Furthermore, the larger the $Q_y$, the more the controller tends to further reduce the lateral tracking error, causing the vehicle to approach the reference path based on the increased heading angle error, resulting in a larger overshoot. Additionally, the smaller the $R_{\Delta \delta}$, the greater the change of front wheel angle, indicating poorer stability of the vehicle, and resulting in the more likely overshoot of the path-tracking controller. On the contrary, the larger the prediction horizon $N_p$, the smaller the weight coefficient $Q_y$, and the larger the weight coefficient $R_{\Delta \delta}$, the slower the response speed of path-tracking control. However, the overshoot phenomenon of path-tracking controller can be effectively suppressed, and the lateral tracking error caused by overshoot phenomenon can be reduced.
The influence of $N_p$, $Q_y$, $R_{\Delta \delta}$ on the vehicle stability is shown in Fig. \ref{fig3}, it can be seen that during the path-tracking process, the larger the prediction horizon $N_p$, the smaller the weight coefficient $Q_y$, the larger the weight coefficient $R_{\Delta \delta}$, the smaller the sideslip angle and yaw rate of the vehicle and their variation range, and the faster they tend to be stable, indicating the better stability of the vehicle and the faster the vehicle returns to the steady state.

Based on the analysis above, it can be concluded that increasing the prediction horizon $N_p$ and weight coefficient $R_{\Delta \delta}$ and reducing the weight coefficient $Q_y$, will lead to the decrease in path-tracking accuracy. However, this approach is beneficial for improving the vehicle's stability, suppressing the overshoot phenomenon of the steering controller, and reducing the tracking error caused by the overshoot phenomenon. In general, as the vehicle speed increases, the vehicle's stability decreases and the steering controller is more likely to overshoot. Hence, as the vehicle speed increases, it is advisable to increase the prediction horizon $N_p$ and weight coefficient $R_{\Delta \delta}$, while decreasing the weight coefficient $Q_y$.
\begin{figure*}[h]
   \centering
\subfigure[Vehicle speed $v_x$- prediction horizon $N_p$ curve]{
   \includegraphics[width=0.32\linewidth]{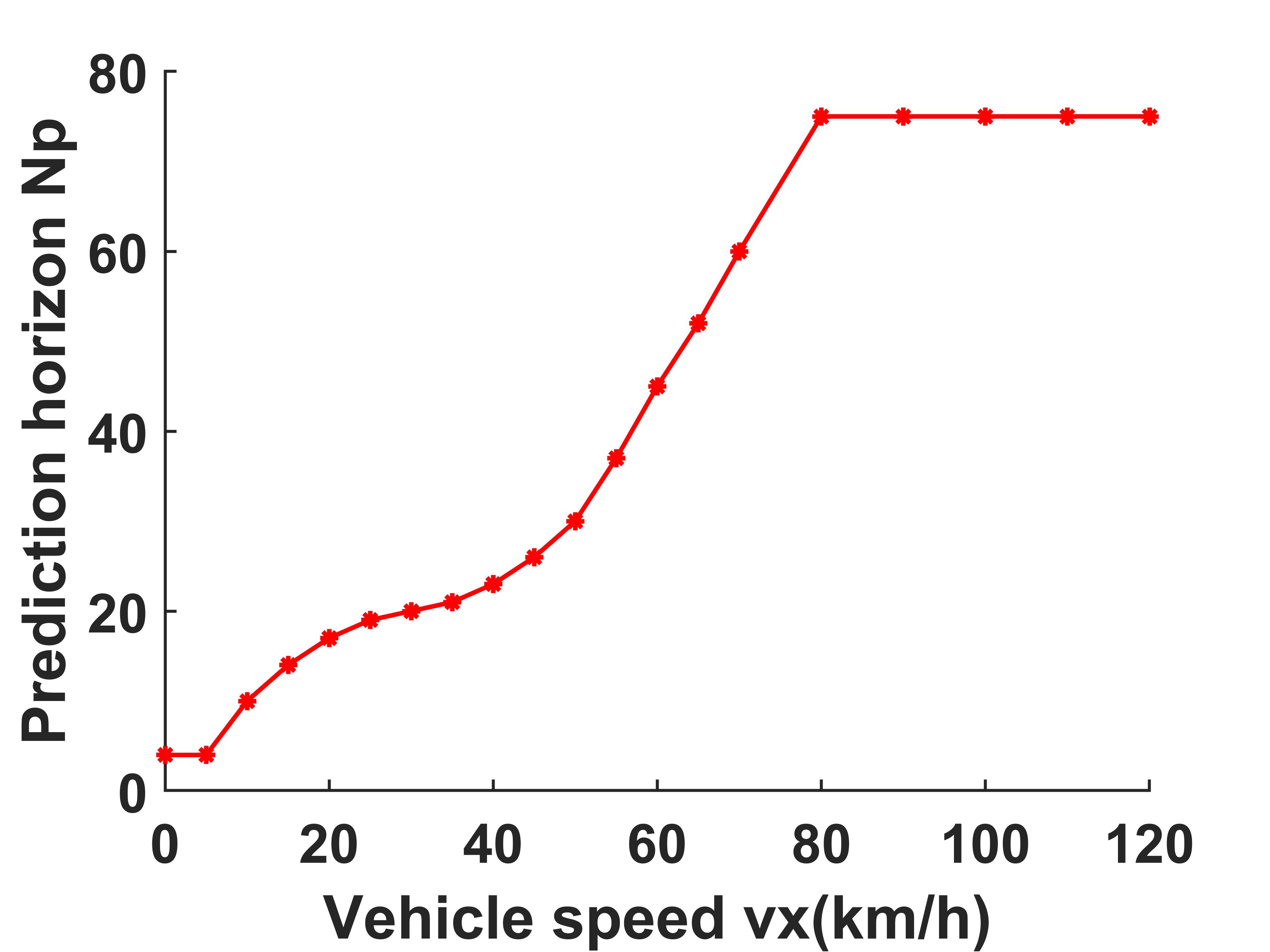}
   }
   \subfigure[Vehicle speed $v_x$- weight coefficient $Q_y$ curve]{
   \includegraphics[width=0.32\linewidth]{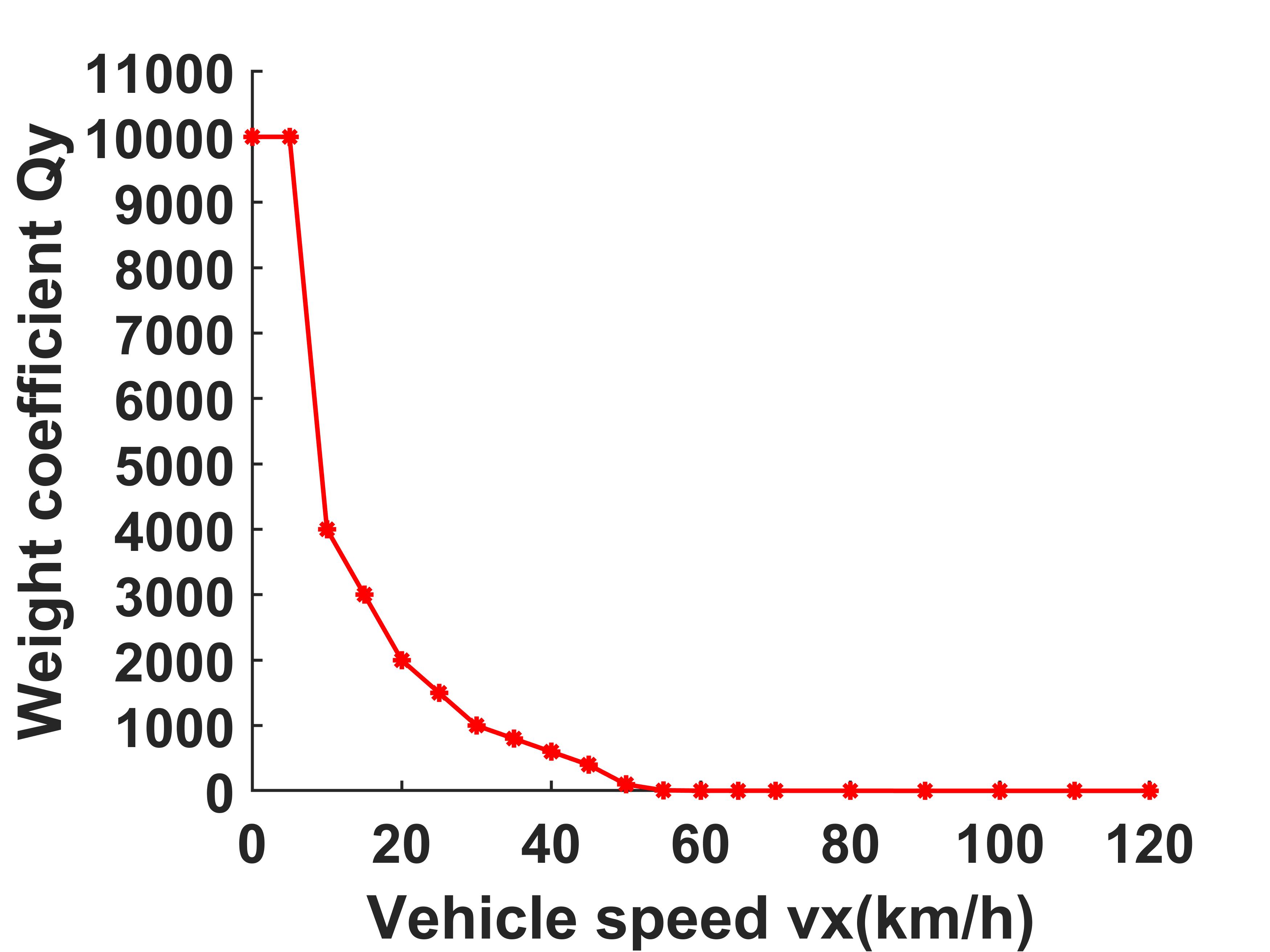}
   }
   \subfigure[Vehicle speed $v_x$- weight coefficient $R_{\Delta \delta}$ curve]{
   \includegraphics[width=0.32\linewidth]{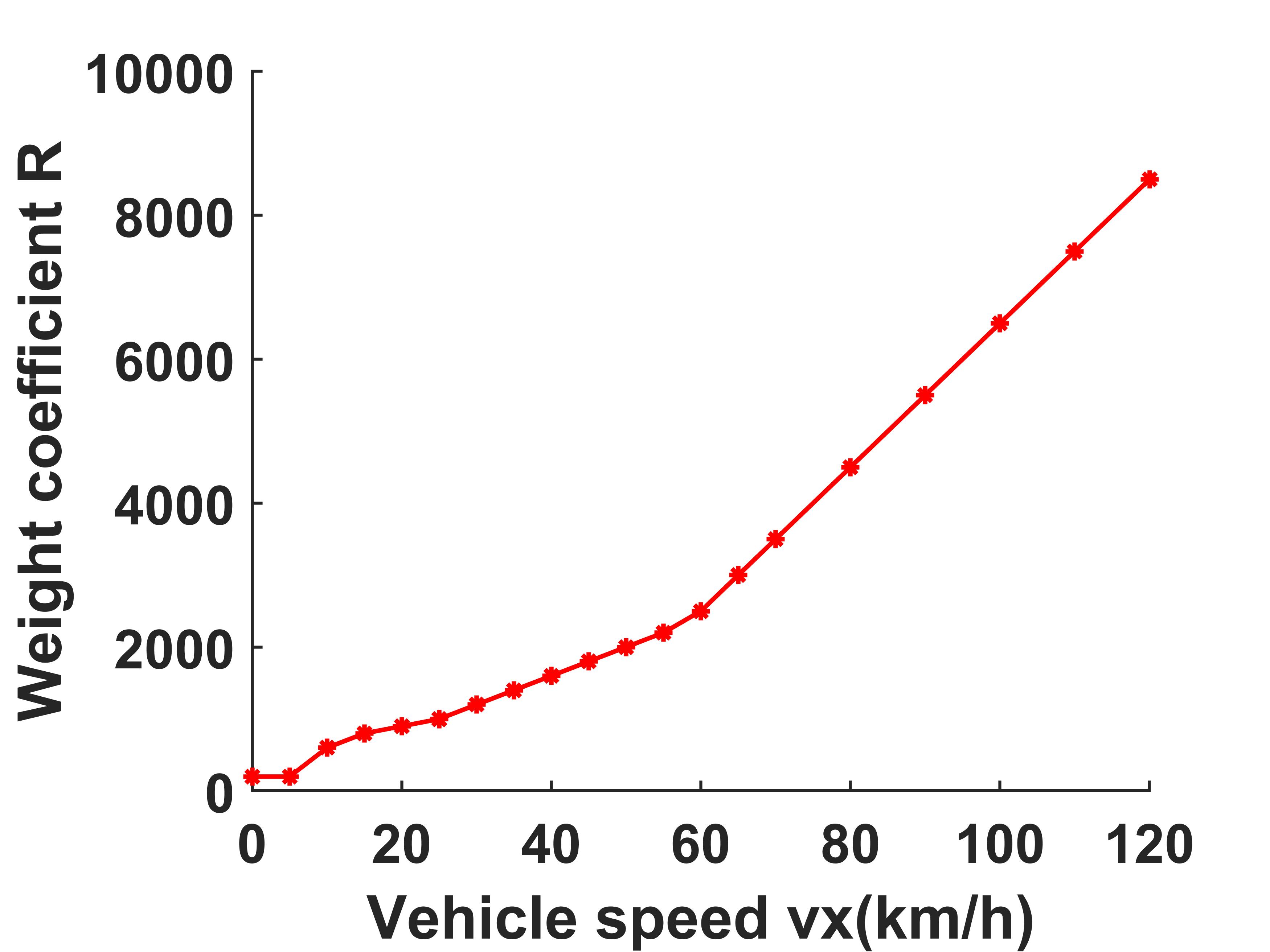}
   }
   \caption{Vehicle speed $v_x$- prediction horizon $N_p$, weight coefficient $Q_y$, and weight coefficient $R_{\Delta \delta}$ curve}
   \label{fig4}
\end{figure*}
\subsection{Adaptive adjustment of prediction horizon and weight coefficient}
When the vehicle is in extreme conditions, such as high speed and low adhesion, it is more prone to dangerous behaviors such as sideslip and tail flick, and the steering controller is more likely to overshoot. At such times, it is advisable to appropriately increase the prediction horizon $N_p$ and weight coefficient $R_{\Delta \delta}$, while reducing the weight coefficient $Q_y$. This can help improve the tracking stability of the vehicle and effectively suppress the overshoot phenomenon of controller. According to the analysis results in Section 3.1, as the vehicle speed increases, the prediction horizon $N_p$ and weight coefficient $R_{\Delta \delta}$ should be increased, while the weight coefficient $Q_y$ should be decreased.

To achieve adaptive adjustment of the prediction horizon and weight coefficients based on the above conclusions, the double lane change line is taken as the reference path, and the corresponding controller parameters are set for different vehicle speeds. The prediction horizon and weight coefficients that achieve a comprehensive improvement in the vehicle's path-tracking accuracy and stability are then selected for different vehicle speeds. To prevent the prediction horizon $N_p$ and weight coefficient $R_{\Delta \delta}$ from being too small and the weight coefficient $Q_y$ from being too large when the vehicle is running at low speed, the controller parameters are kept unchanged within the vehicle speed's range of 0-5 km/h. When the vehicle is running at high speed, if the prediction horizon is too large, the real-time performance of the path-tracking controller will be deteriorated due to excessive calculation. This can result in the poor stability and path-tracking accuracy of the vehicle. To avoid this issue, the maximum limit of $N_p$ is set to 75. Furthermore, when the vehicle is running at high speed, if the weight coefficient $Q_y$ is too small, the lateral tracking error of the vehicle will not be effectively eliminated, therefore the minimum limit of $Q_y$ is set to 2. The selected controller parameters and their corresponding vehicle speeds are plotted as the vehicle speed $v_x$-prediction horizon $N_p$, weight coefficient $Q_y$, and weight coefficient $R_{\Delta \delta}$ curve shown in Fig. \ref{fig4}. Based on the plotted curve, 
three one-dimensional lookup tables have been created to achieve the adaptive adjustment of prediction horizon and weight coefficients for different vehicle speeds, with the vehicle speed as the input variable.

\section{LQR-based direct yaw moment control}

The sideslip angle and yaw rate of the vehicle are two important parameters to judge the vehicle's stability \cite{liu2018vehicle}. In this paper, the intervention rule of yaw moment control are designed based on the threshold of yaw rate error and the $\beta-\dot{\beta}$ phase diagram.

When the yaw rate, sideslip angle and sideslip angle rate of the vehicle satisfy the following inequalities, the vehicle is judged to be stable, otherwise the yaw moment control is intervened \cite{xia2021autonomous}.
\begin{equation}
\left\{\begin{array}{c}
\left|\dot{\varphi}-\dot{\varphi}_d\right| \leq 0.035 \mathrm{rad} \\
\left|B_1 \dot{\beta}+B_2 \beta\right| \leq 1
\end{array}\right.
\label{eq19}
\end{equation}
where the values of $B_1$ and $B_2$ are affected by factors such as road adhesion coefficient, vehicle speed, and front wheel angle.

LQR can obtain the optimal control law of state linear feedback, which is easy to form a closed-loop optimal control. Compared with MPC, LQR offers several advantages, such as small computation, simple solution process, high real-time performance and low cost. When the yaw moment control is applied, the vehicle has a tendency to lose stability, making it necessary for the yaw moment controller to have high real-time performance to adjust the vehicle attitude in a timely manner. The mature solution method of the Riccati equation ensures the real-time performance of LQR. Based on the above analysis, the paper utilizes LQR to design a direct yaw moment controller for achieving real-time control of the vehicle's stability.

Direct yaw moment control considers only two degrees of freedom of the vehicle: yaw motion and lateral motion \cite{hua2019hierarchical}. Following the monorail vehicle dynamics model shown in Fig. \ref{fig0}, the differential equation of the two-degree-of-freedom vehicle model can be obtained by combining Eq. (\ref{eq7}) as follows:
\begin{equation}
\left\{\begin{array}{l}
(\dot{\beta}+\dot{\varphi}) m v_x=2\left[C_{\alpha f}\left(\frac{v_y+a \dot{\varphi}}{v_x}-\delta_f\right)+C_{\alpha r} \frac{v_y-b \dot{\varphi}}{v_x}\right] \\
I_z \ddot{\varphi}=2\left[a C_{\alpha f}\left(\frac{v_y+a \dot{\varphi}}{v_x}-\delta_f\right)-b C_{\alpha r} \frac{v_y-b \dot{\varphi}}{v_x}\right]+M_z
\end{array}\right.
\label{eq20}
\end{equation}
where $M_z$ is the direct yaw moment decided by LQR-based controller. Let $x_1=\beta$, $x_2=\ddot{\varphi}$, the state equation of two-degree-of-freedom vehicle dynamics can be obtained as follows:
\begin{equation}
\dot{\xi}=A \xi+B U+C W
\label{eq21}
\end{equation}
where $A=\left[\begin{array}{cc}\frac{2\left(C_{\alpha f}+C_{\alpha r}\right)}{m v_x} & \frac{2\left(a C_{\alpha f}-b C_{\alpha r}\right)}{m v_x^2}-1 \\ \frac{2\left(a C_{\alpha f}-b C_{\alpha r}\right)}{I_z} & \frac{2\left(a^2 C_{\alpha f}+b^2 c_{\alpha r}\right)}{I_z v_x}\end{array}\right]$, $B=\left[\begin{array}{l}0 \\ \frac{1}{I_z}\end{array}\right]$, $C=\left[\begin{array}{c}\frac{-2 C_{\alpha f}}{m v_x} \\ \frac{-2 a C_{\alpha f}}{I_z}\end{array}\right]$, $\dot{\xi}=\left[\begin{array}{l}\dot{\beta} \\ \ddot{\varphi}\end{array}\right]$, $\xi=\left[\begin{array}{l}\beta \\ \dot{\varphi}\end{array}\right]$, $U=M_z$, $W=\delta_f$.

The control purpose of LQR-based controller is to make the vehicle's yaw rate and sideslip angle close to the reference value smoothly, so the objective function of the controller is designed as follows:
\begin{equation}
J=\frac{1}{2} \int_0^{\infty}\left[\left(X-X_d\right)^T Q\left(X-X_d\right)+U^T R U\right] d t
\label{eq22}
\end{equation}
where $Q$ is the weight matrix of state error, $R$ is the weight matrix of the control quantity, $Q$ is a positive semidefinite matrix, $R$ is a positive definite matrix, $X_d=\left[\begin{array}{l}\beta_{r e f} \\ \dot{\varphi}_{r e f}\end{array}\right]$, $\beta_{r e f}$ is the reference value of the vehicle's sideslip angle, $\dot{\varphi}_{r e f}$ is the reference value of the vehicle's yaw rate. The reference value of the vehicle's sideslip angle is set to 0. 

Since the vehicle's lateral acceleration is limited by the tire-road friction coefficient, the upper limit of yaw rate is calculated by the following equation:
\begin{equation}
\dot{\varphi}_{\max }=0.85 \frac{\mu g}{v_x}
\label{eq23}
\end{equation}
Based on the above analysis, the yaw rate's reference value can be determined by the following equation:
\begin{equation}
\dot{\varphi}_{r e f}=\min \left\{\dot{\varphi}_d, \dot{\varphi}_{\max }\right\}
\label{eq24}
\end{equation}
where $\dot{\varphi}_d$ is the reference value of the vehicle's transient yaw rate when $M_z=0$, which is calculated based on Eq. (\ref{eq20}).

Let $X_d=A_d W$, so there is the following equation:
\begin{equation}
A_d=\frac{X_d}{W}=\left[\begin{array}{c}
0 \\
\frac{\dot{\varphi}_d}{\delta_f}
\end{array}\right]
\label{eq25}
\end{equation}
In the above equation, set $\delta_f \neq 0$.

It is assumed that the control law of LQR-based controller is expressed as follows:
\begin{equation}
U=K_{F B} X+K_{F F} W
\label{eq26}
\end{equation}
where $K_{F B}$ is the state feedback gain matrix and $K_{F F}$ is the feedforward gain matrix.

The variational method is used to solve the quadratic optimal control problem, and the optimal control law is finally solved as follows:
\begin{equation}
\begin{aligned}
U^*(t) & =-R^{-1} B^T P X \\
& +R^{-1} B^T\left[P B R^{-1} B^T-A^T\right]^{-1}\left(Q A_d-P C\right) W
\end{aligned}
\label{eq27}
\end{equation}
Combining Eq. (\ref{eq26}) and Eq. (\ref{eq27}), the following equation can be obtained:
\begin{equation}
\left\{\begin{array}{c}
K_{F B}=-R^{-1} B^T P \\
K_{F F}=R^{-1} B^T\left[P B R^{-1} B^T-A^T\right]^{-1}\left(Q A_d-P C\right)
\end{array}\right.
\label{eq28}
\end{equation}
$P$ is the positive definite solution of the following $Riccati$ equation:
\begin{equation}
A^T P+P A+Q-P B R^{-1} B^T P=0
\label{eq29}
\end{equation}
Because the smaller the tire utilization rate is, the larger the lateral force margin of the tire is, and the better the vehicle stability is, so the four-wheel torque distribution is carried out with the minimum tire adhesion utilization as the optimization goal. The objective function is simplified as follows:
\begin{equation}
\min J=\sum_{i j=f l, f r, r l, r r} \frac{F_{x i j}^2}{\left(\mu F_{z i j}\right)^2}
\label{eq30}
\end{equation}
where $F_{x i j}$ is the tire's longitudinal force, $F_{z i j}$ is the tire's vertical force, $miu$ is the tire-road friction coefficient.

The total longitudinal force $F_x$ and yaw moment $M_z$ required by the vehicle are provided by the four-wheel torque $T_{i j}$, which satisfy the following equation:
\begin{equation}
N=A u
\label{eq31}
\end{equation}
where $N=\left[\begin{array}{ll}F_x & M_z\end{array}\right]^T$, 
$A=\frac{1}{r}\left[\begin{array}{cccc}1 & 1 & 1 & 1 \\ -d / 2 & d / 2 & -d / 2 & d / 2\end{array}\right]$,
$u=\left[\begin{array}{llll}T_{f l} & T_{f r} & T_{r l} & T_{r r}\end{array}\right]^T$, $r$ is the wheel radius, and $d$ is the distance between the left and right wheels.

Considering that the four-wheel torque is limited by the maximum drive torque of the motor and the road adhesion conditions, the constraints are set as follows:
\begin{equation}
\left|T_{i j}\right| \leq \min \left|\left(\mu F_{z i j}, T_{\max } / r\right)\right|
\label{eq32}
\end{equation}
where $T_{max}$ is the maximum drive torque of the motor.

Based on the above objective function and constraint conditions, the four-wheel torque optimization allocation problem can be transformed into a quadratic programming problem as shown in Eq. (\ref{eq33}), so as to obtain the optimal torque of each wheel.
\begin{equation}
\left\{\begin{array}{c}
\min J=u^T \operatorname{diag}\left(1 /\left(\mu F_{z i j}\right)^2\right)_{4 \times 4} u \\
\text { s.t. } N=A u \\
\qquad\enspace|u| \leq u_{\max }
\end{array}\right.
\label{eq33}
\end{equation}

\section{Simulation results and Analysis}
\begin{table}[h]
\centering
\caption{Main parameters of vehicle model}
\label{table}
\begin{tabular}{|c|c|}
\hline
Parameter& Value \\
\hline
Curb weight $m$/kg& 1860\\
Wheel tread $d$/m& 1.6\\
Wheel radius $r$/m& 0.3\\
Cornering stiffness of front tire $C_f$/(N/rad)& -77223\\
Cornering stiffness of rear tire $C_r$/(N/rad)& -66782\\
Moment of inertia $I_z$/($\mathrm{kg} \cdot \mathrm{m}^2$)& 4175\\
Distance from centroid to front axle $a$/m& 1.232\\
Distance from centroid to rear axle $b$/m& 1.468\\
\hline
\end{tabular}
\label{tab1}
\end{table}

To verify the effectiveness of the adaptive MPC and coordinated control strategy, a CarSim/Simulink co-simulation platform is built for the simulation test of path tracking. The double lane change path is selected as the reference path, and the road adhesion coefficient is set to 0.6. The main parameters of the vehicle model are shown in Tab. \ref{tab1}.

\begin{figure*}[h]
   \centering
\subfigure[Tracking path at 18km/h]{
   \includegraphics[width=0.32\linewidth]{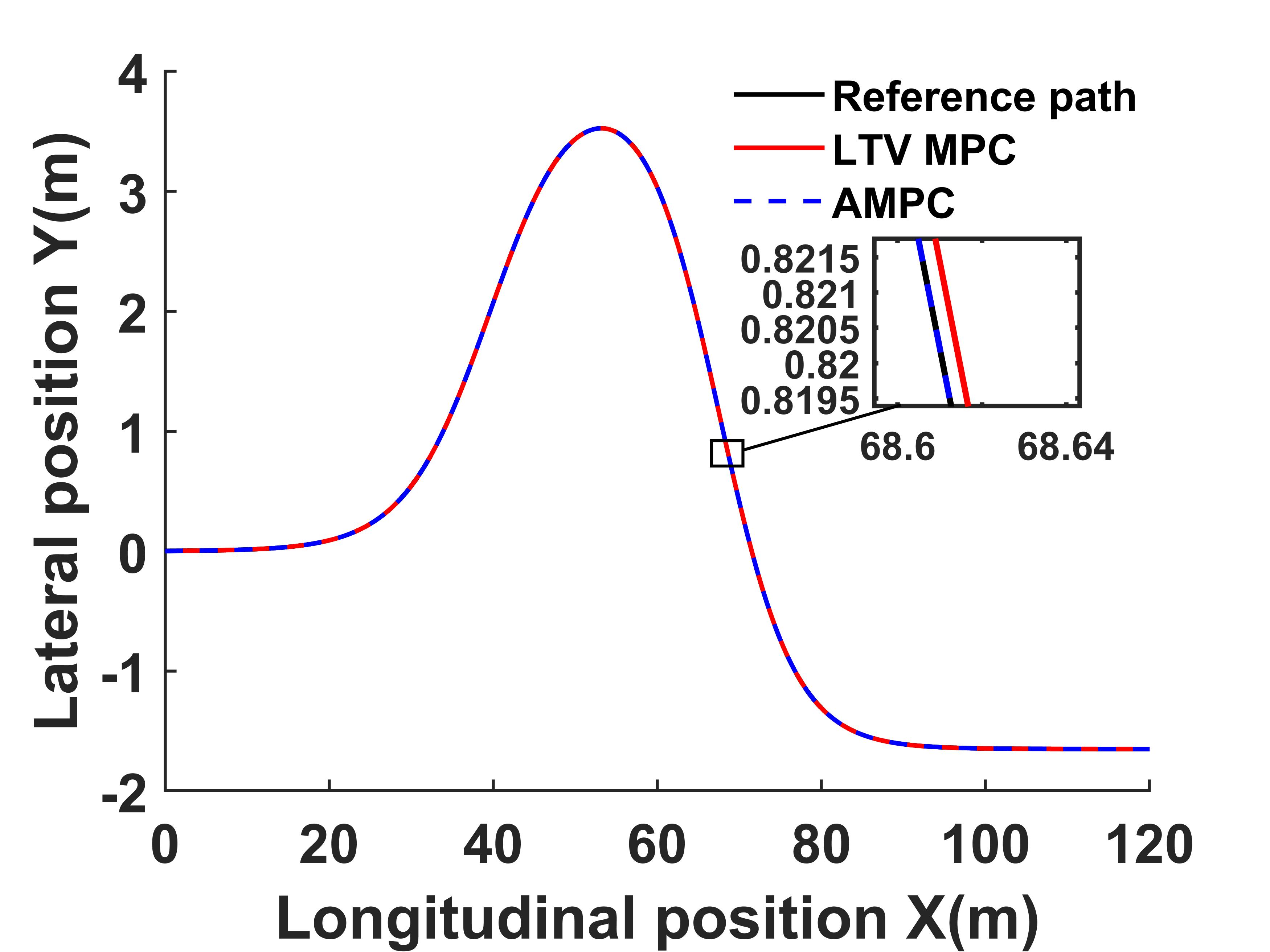}
   }
   \subfigure[Lateral error at 18km/h]{
   \includegraphics[width=0.32\linewidth]{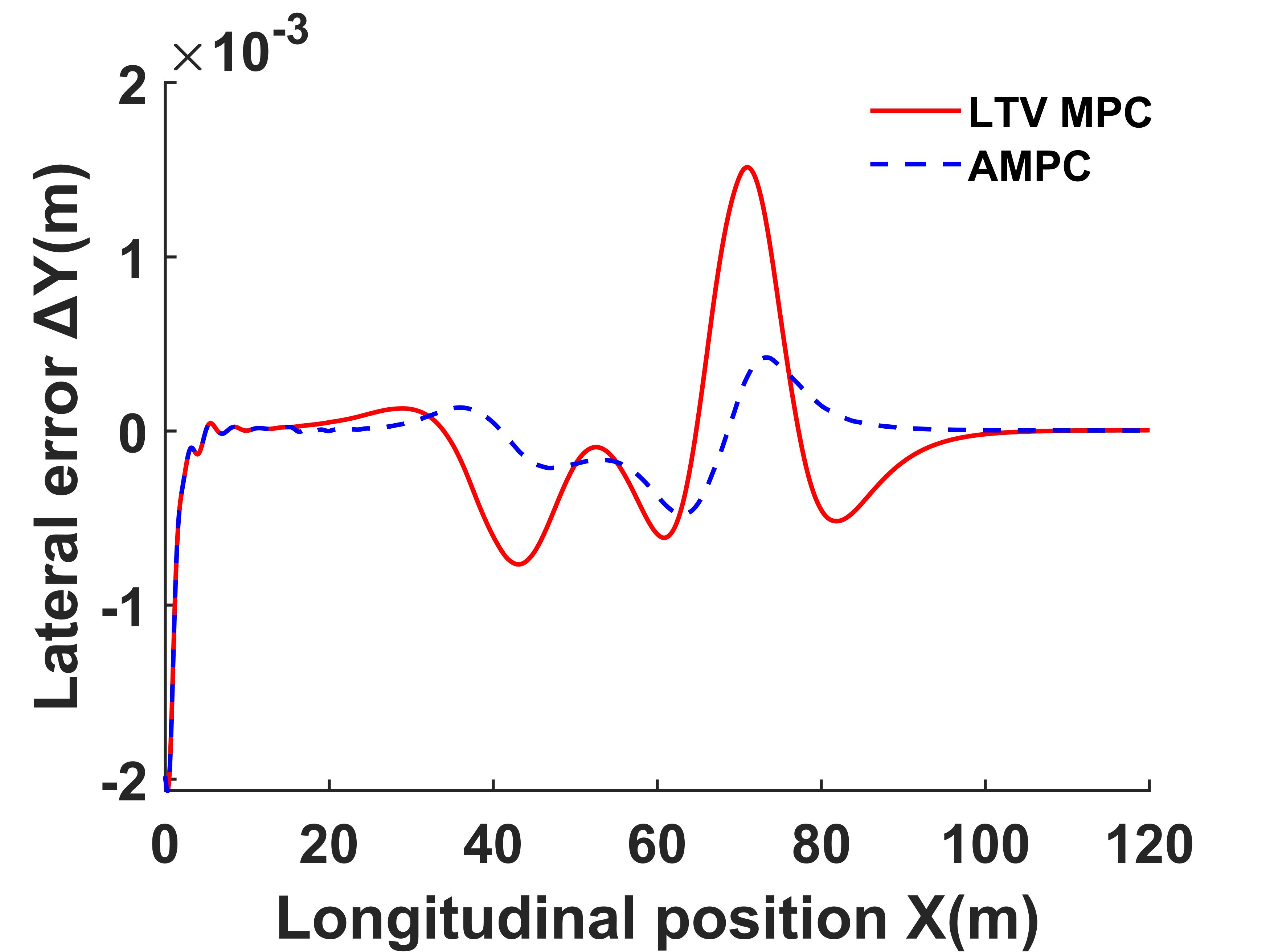}
   }
   \subfigure[Yaw rate at 18km/h]{
   \includegraphics[width=0.32\linewidth]{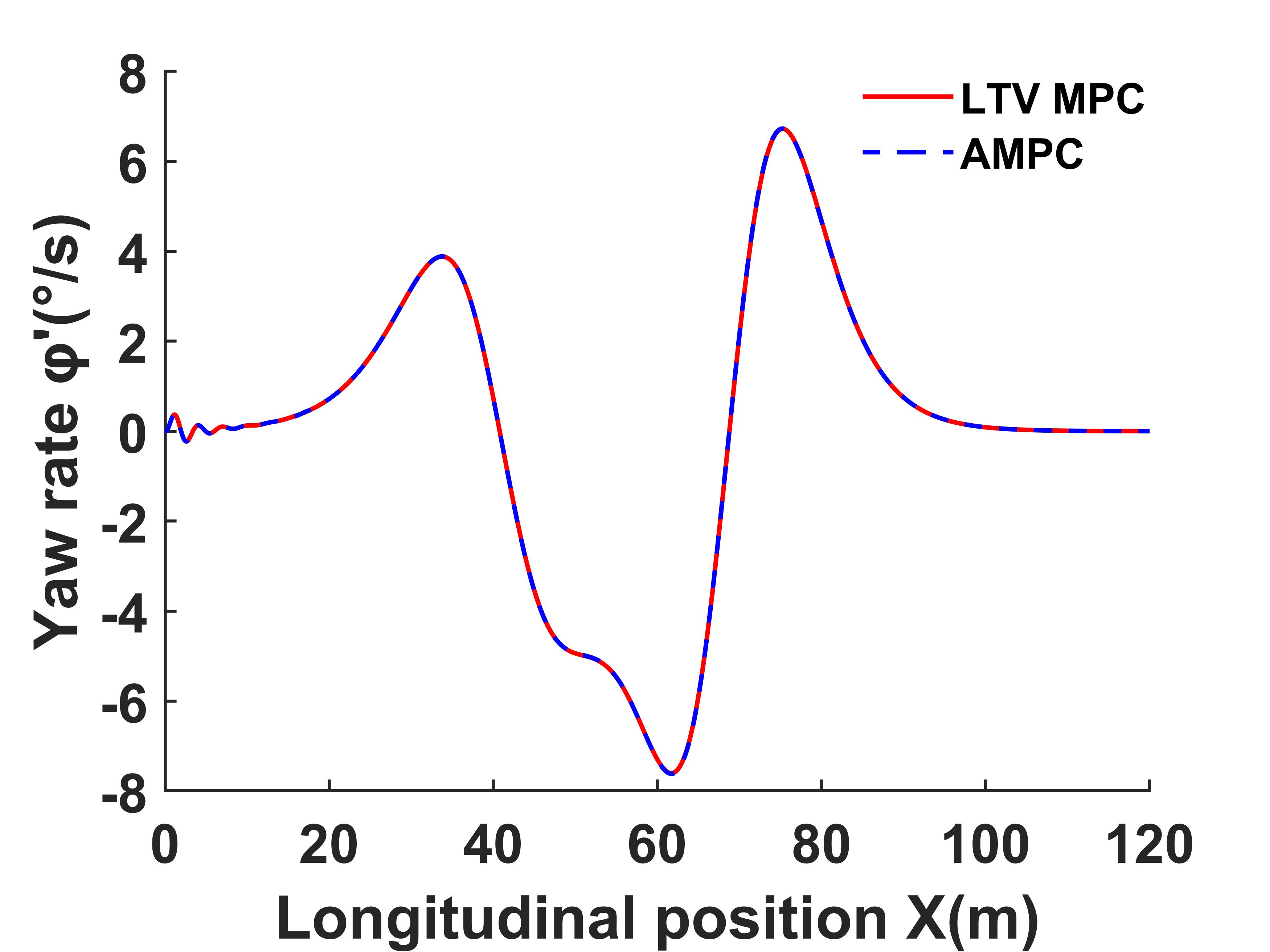}
   }
   \subfigure[Sideslip angle at 18km/h]{
   \includegraphics[width=0.32\linewidth]{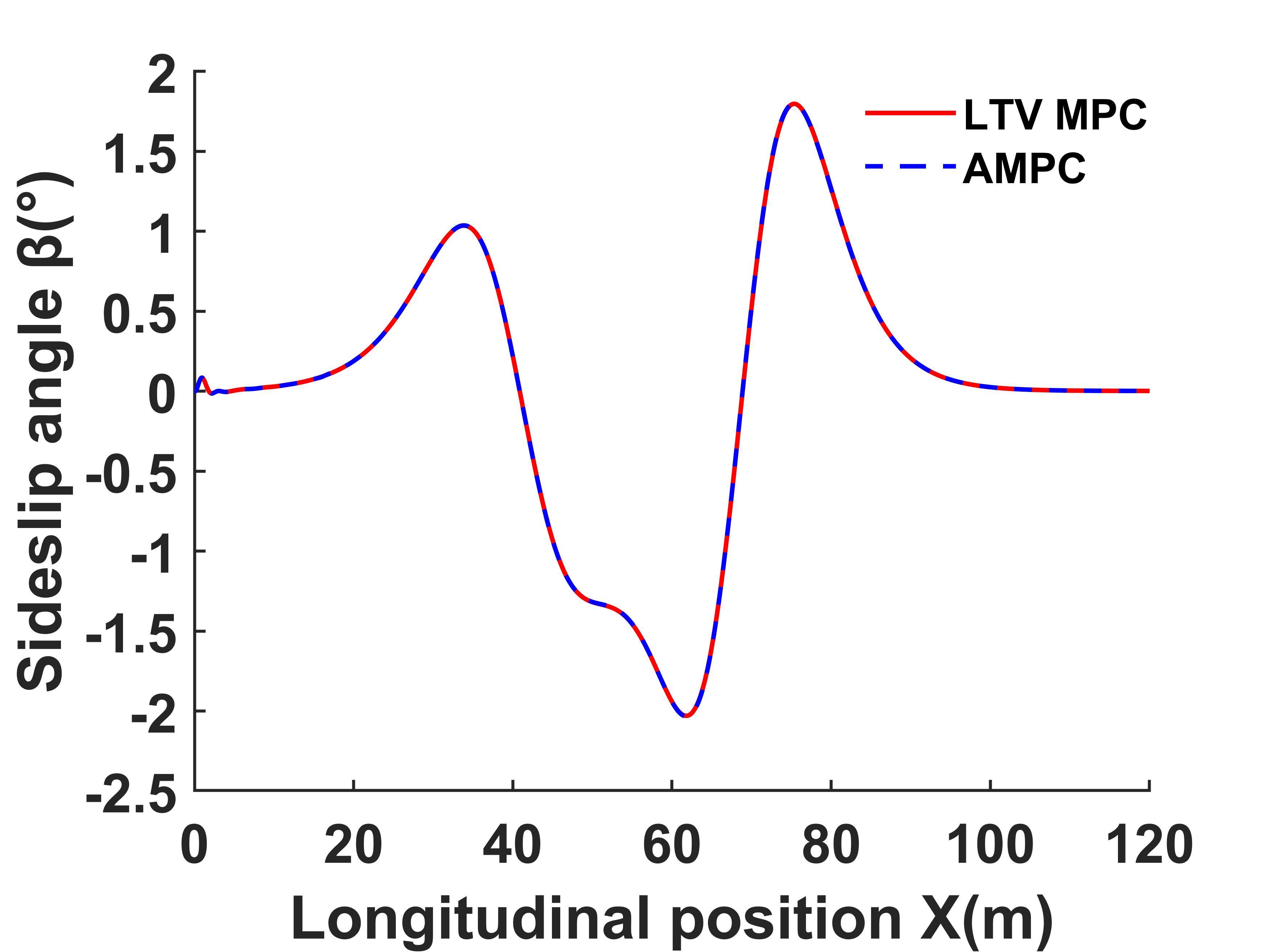}
   }
   \caption{Simulation results of path tracking at 18km/h}
   \label{fig5}
\end{figure*}
\begin{table*}[h]
\centering
\caption{Comparison of vehicle's performance parameters RMS at 18km/h}
\label{table}
\begin{tabular}{|c|c|c|c|}
\hline
Algorithm& Lateral error $\Delta \mathrm{Y}$& Yaw rate $\dot{\varphi}$& Sideslip angle $\Delta \mathrm{Y}$ \\
\hline
LTV MPC& 4.7180e-04& 3.3155& 0.8855\\
AMPC& 2.5031e-04& 3.3170& 0.8859\\
$\triangle \mathrm{RMS}_{\text {AMPC-LTV MPC }}(\%)$& 46.95$\%$& -0.05$\%$& -0.05$\%$\\
\hline
\end{tabular}
\label{tab2}
\end{table*}

\subsection{Verification of control effect of adaptive MPC}
To verify the path-tracking accuracy and stability of the vehicle under the action of adaptive MPC at low and high speeds, a comparison and analysis of the tracking performance of the vehicle under the effect of AMPC and LTV MPC are conducted. The simulation is carried out under uniform driving conditions with the vehicle speed set to 18 km/h and 62 km/h, respectively. In the comparative analysis, LTV MPC should be selected with different control parameters from those of AMPC. Therefore, the control parameters of AMPC are set as the adaptive prediction horizon and weight coefficients corresponding to the vehicle speed of 40 km/h. Under the action of LTV MPC, the prediction horizon $N_p$, weight coefficient $Q_y$ and weight coefficient $R_{\Delta \delta}$ are set to 23, 600, and 1600, respectively. 

(1) The vehicle speed is set to 18 km/h.

 Since AMPC aims to improve the vehicle's path-tracking accuracy when it is running at low speed, its prediction horizon $N_p$ and weight coefficient $R_{\Delta \delta}$ should be reduced compared to LTV MPC, while its weight coefficient $Q_y$ should be increased. Based on the adaptive adjustment of prediction horizon and weight coefficients, the prediction horizon $N_p$, weight coefficient $Q_y$ and weight coefficient $R_{\Delta \delta}$ of AMPC are set to 16, 2400 and 860, respectively. The simulation results are presented in Fig. \ref{fig5}.

Fig. \ref{fig5} shows the simulation results of path tracking at the vehicle speed of 18 km/h. The tracking path and lateral error of the vehicle are shown in Fig. \ref{fig5} (a) and Fig. \ref{fig5} (b), and the yaw rate and sideslip angle of the vehicle are shown in Fig. \ref{fig5} (c) and Fig. \ref{fig5} (d). Tab. \ref{tab2} shows the comparison of root mean square (RMS) of the vehicle's lateral error, yaw rate and sideslip angle of under the action of LTV MPC and AMPC at the vehicle speed of 18 km/h. According to Fig. \ref{fig5} (a), Fig. \ref{fig5} (b) and Tab. \ref{tab2}, the RMS of lateral error under AMPC is reduced by 46.95$\%$ compared to LTV MPC, indicating that the vehicle shows better path-tracking accuracy. According to Fig. \ref{fig5} (c), Fig. \ref{fig5} (d) and Tab. \ref{tab2}, the RMS of yaw rate and sideslip angle under AMPC are increased by 0.05$\%$ compared to LTV MPC, indicating a slight loss in the vehicle's stability. This is because when the vehicle is running at low speed, its stability is better, and AMPC prioritizes improving the vehicle's path-tracking accuracy at the cost of a small decrease in the vehicle's stability. 

Based on the above analysis, it can be concluded that when the vehicle is running at low speed, AMPC significantly improves the path-tracking accuracy compared to LTV MPC with a small loss of the vehicle's stability.
\begin{figure*}[h]
   \centering
\subfigure[Tracking path at 62 km/h]{
   \includegraphics[width=0.32\linewidth]{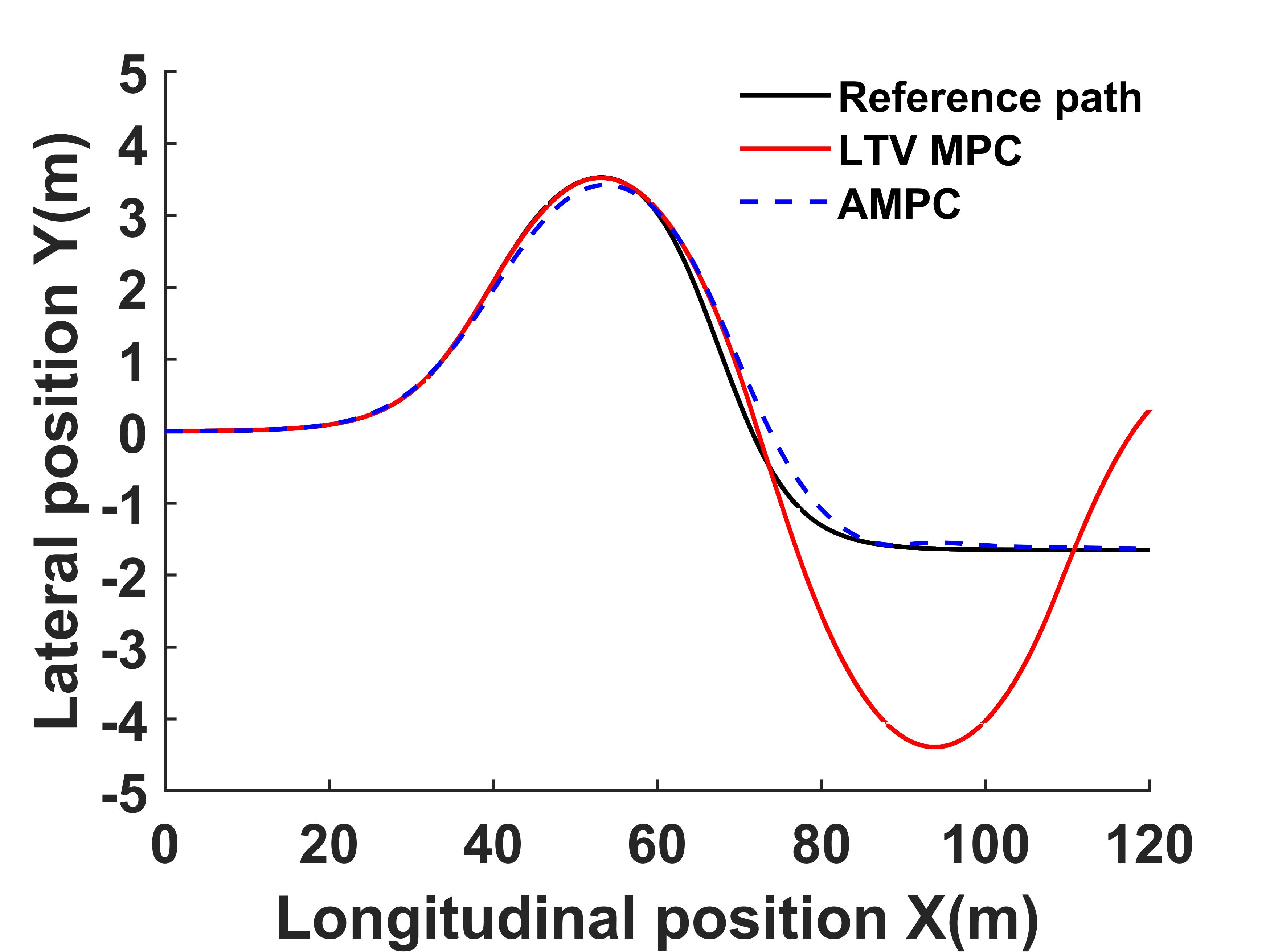}
   }
   \subfigure[Lateral error at 62 km/h]{
   \includegraphics[width=0.32\linewidth]{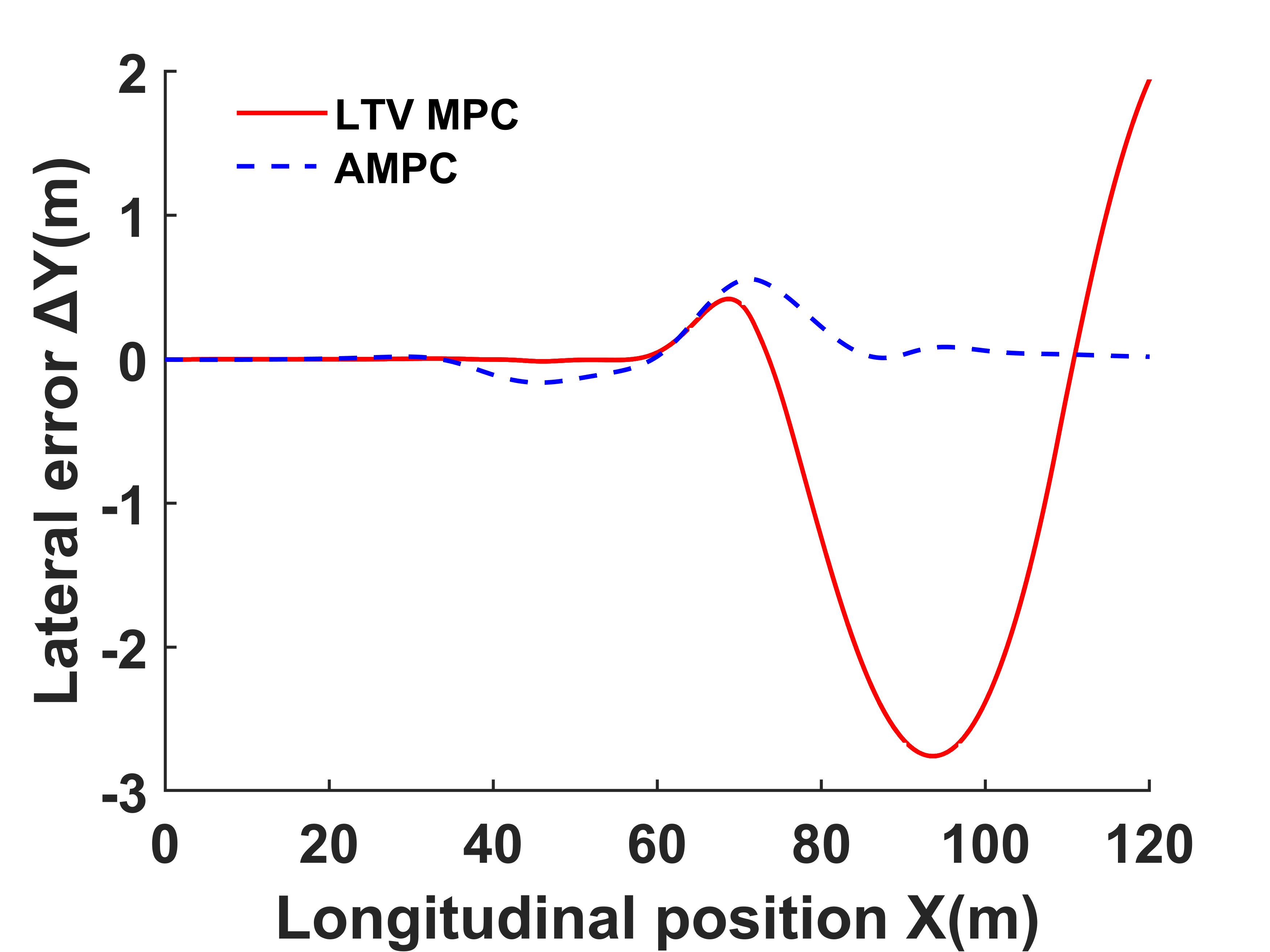}
   }
   \subfigure[Yaw rate at 62 km/h]{
   \includegraphics[width=0.32\linewidth]{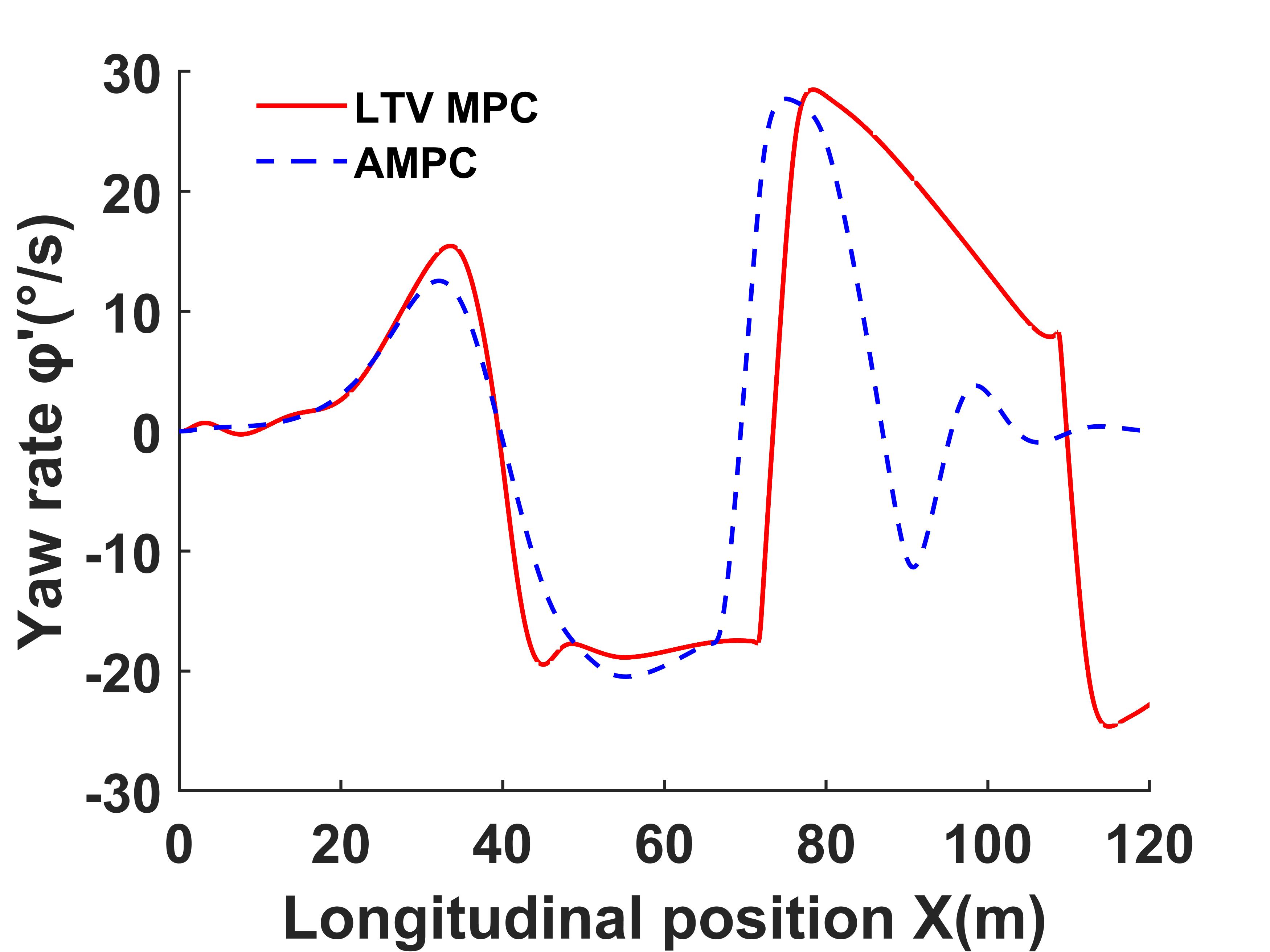}
   }
   \subfigure[Sideslip angle at 62 km/h]{
   \includegraphics[width=0.32\linewidth]{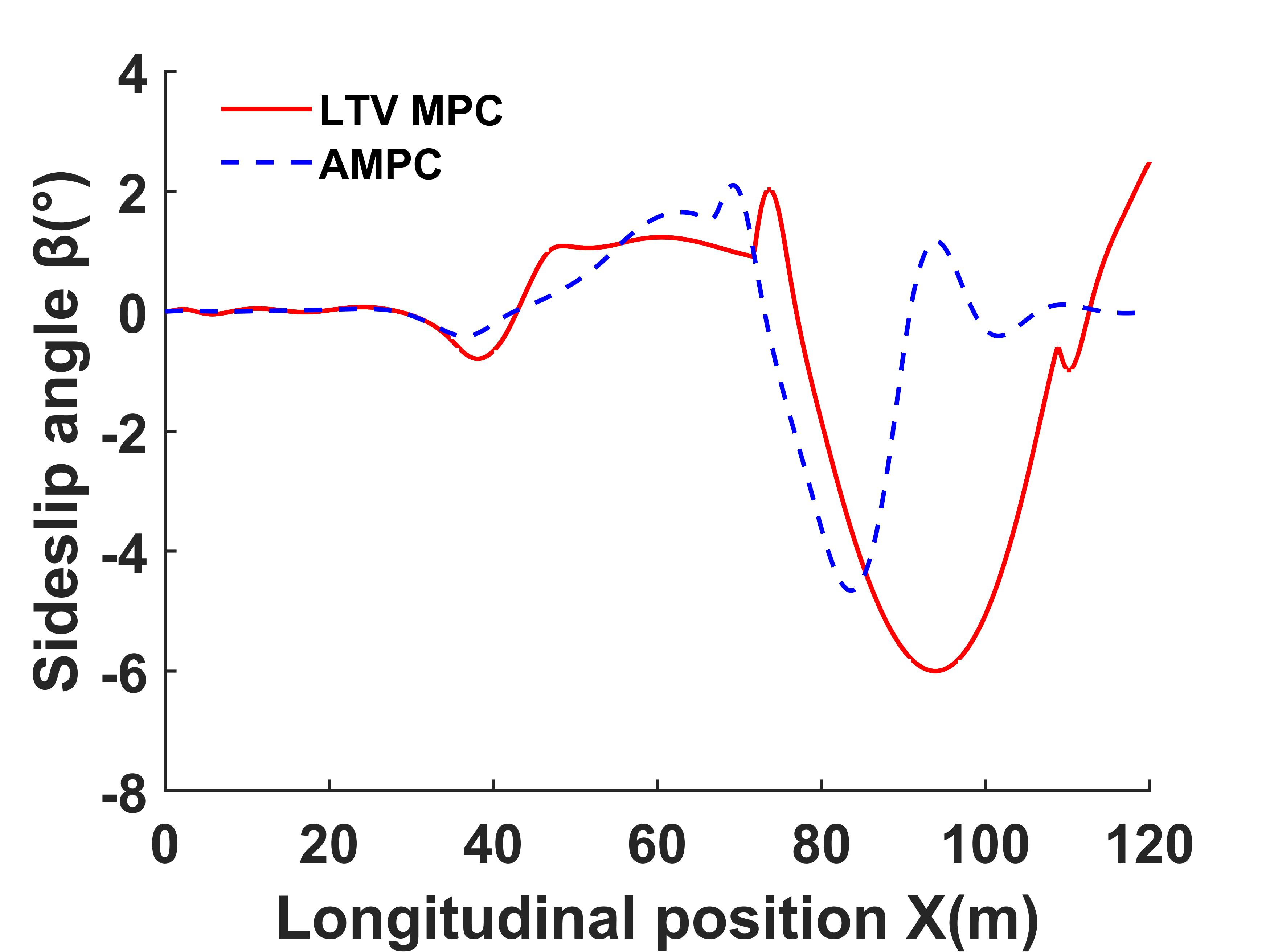}
   }
   \caption{Simulation results of path tracking at 62 km/h}
   \label{fig6}
\end{figure*}
\begin{table*}[h]
\centering
\caption{Comparison of vehicle's performance parameters RMS at 62km/h}
\label{table}
\begin{tabular}{|c|c|c|c|}
\hline
Algorithm& Lateral error $\Delta \mathrm{Y}$& Yaw rate $\dot{\varphi}$& Sideslip angle $\Delta \mathrm{Y}$ \\
\hline
LTV MPC& 1.3045& 16.2317& 2.4144\\
AMPC& 0.1782& 12.3176& 1.3969\\
$\triangle \mathrm{RMS}_{\text {AMPC-LTV MPC }}(\%)$& 86.34$\%$& 24.11$\%$& 42.14$\%$\\
\hline
\end{tabular}
\label{tab3}
\end{table*}

(2) The vehicle speed is set to 62 km/h.

Since AMPC aims to improve the vehicle's stability when it is running at high speed, its prediction horizon $N_p$ and weight coefficient $R_{\Delta \delta}$ should be increased, while its weight coefficient $Q_y$ should be reduced compared to LTV MPC. Based on the adaptive adjustment of prediction horizon and weight coefficients, the prediction horizon $N_p$, weight coefficient $Q_y$ and weight coefficient $R_{\Delta \delta}$ of AMPC are set to 48, 4 and 2700, respectively. The simulation results are shown in Fig. \ref{fig6}. 

Fig. \ref{fig6} shows the simulation results of path-tracking at the vehicle speed of 62 km/h. The tracking path and lateral error of the vehicle are shown in Fig. \ref{fig6} (a) and Fig. \ref{fig6} (b), and the yaw rate and sideslip angle of the vehicle are shown in Fig. \ref{fig6} (c) and Fig. \ref{fig6} (d). Tab. \ref{tab3} shows the comparison of root mean square (RMS) of the vehicle's lateral error, yaw rate and sideslip angle under the action of LTV MPC and AMPC at the vehicle speed of 62 km/h. According to Fig. \ref{fig6} (a), Fig. \ref{fig6} (b) and Tab. \ref{tab3}, the RMS of lateral error under AMPC is reduced by 86.34$\%$ compared to LTV MPC, indicating that the vehicle shows better path-tracking accuracy. According to Fig. \ref{fig6} (c), Fig. \ref{fig6} (d) and Tab. \ref{tab3}, compared to LTV MPC, the RMS of yaw rate and sideslip angle under AMPC are reduced by 24.11$\%$ and 42.14$\%$, respectively, indicating that the vehicle shows better stability. According to Fig. \ref{fig6} (a) and Fig. \ref{fig6} (b), the vehicle's path-tracking accuracy before 74 meters is reduced under the action of AMPC compared to LTV MPC. This is because AMPC focuses on improving the vehicle's stability at high speed and thus compromises a small part of the tracking accuracy. However, throughout the entire path-tracking process, the vehicle's tracking accuracy of under the action of AMPC is significantly better. This is because AMPC improves the vehicle's stability, thereby preventing the steering controller from losing path-tracking ability due to vehicle instability.

Based on the analysis above, it can be concluded that when the vehicle is driving at high speed and has good stability, AMPC significantly improves the vehicle's stability compared to LTV MPC, even if there is some loss of path-tracking accuracy. Moreover, when the vehicle's stability is poor, AMPC helps the steering controller maintain good path-tracking ability and prevents the vehicle from deviating greatly from the reference path.

\subsection{Comparative analysis of control effect of AMPC, LTV MPC+DYC, AMPC+DYC}
In order to verify the effectiveness of direct yaw moment control in improving the vehicle's stability, as well as the adaptability of path-tracking controller to variable speeds, and the control effect of coordinating AMPC and DYC, the following two sets of simulation experiments have been designed. The double lane change path is selected as the reference path, and the road adhesion coefficient is set to 0.6. 

(1) The vehicle speed is set to 5-65 km/h.

\begin{figure}[h]
\centerline{\includegraphics[width=0.7\linewidth]{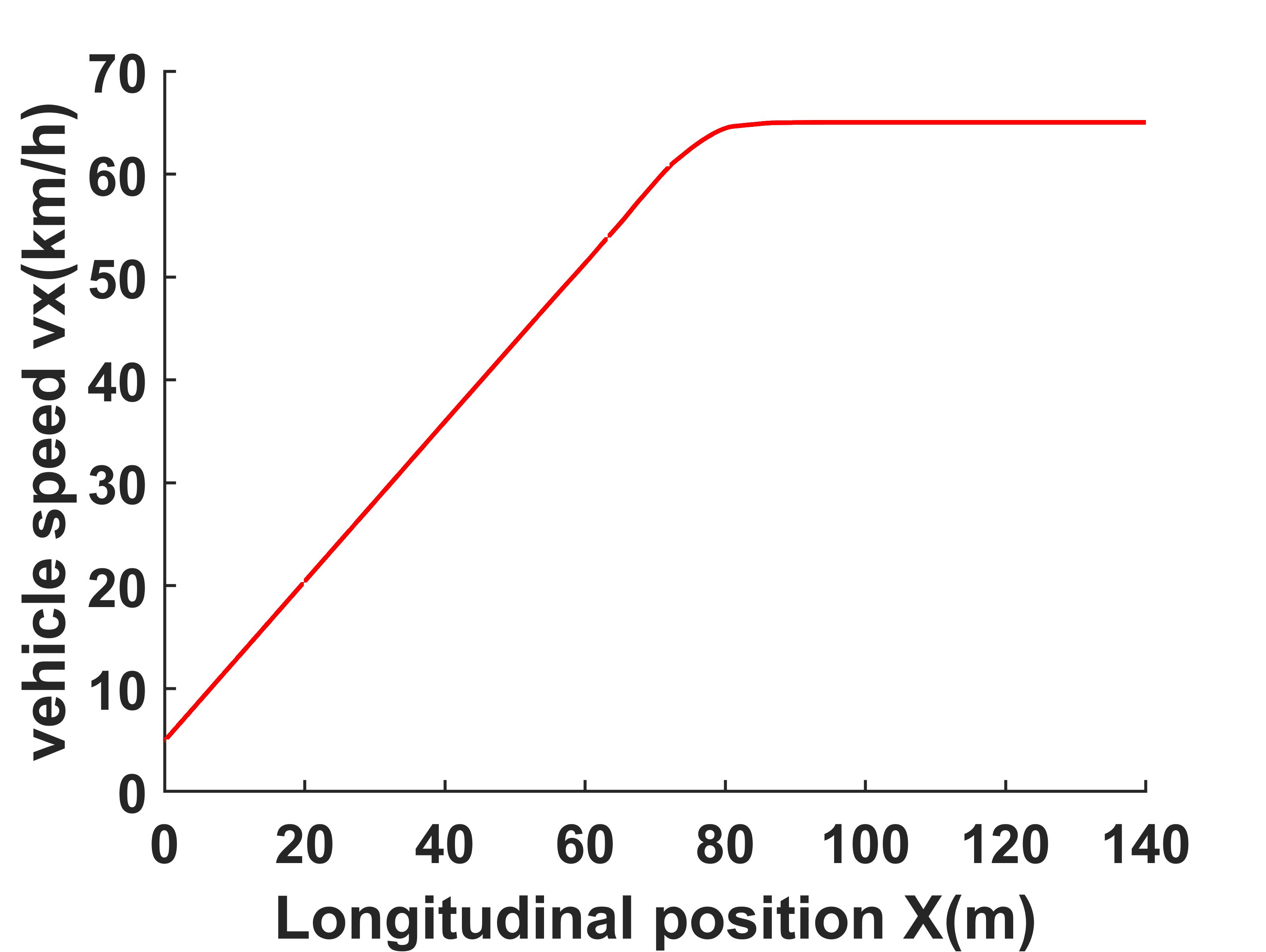}}
\caption{Longitudinal position $X$-vehicle speed $v_x$ curve}
\label{fig7}
\end{figure}

\begin{figure*}[h]
   \centering
   \subfigure[Tracking path for 5-65 km/h]{
   \includegraphics[width=0.32\linewidth]{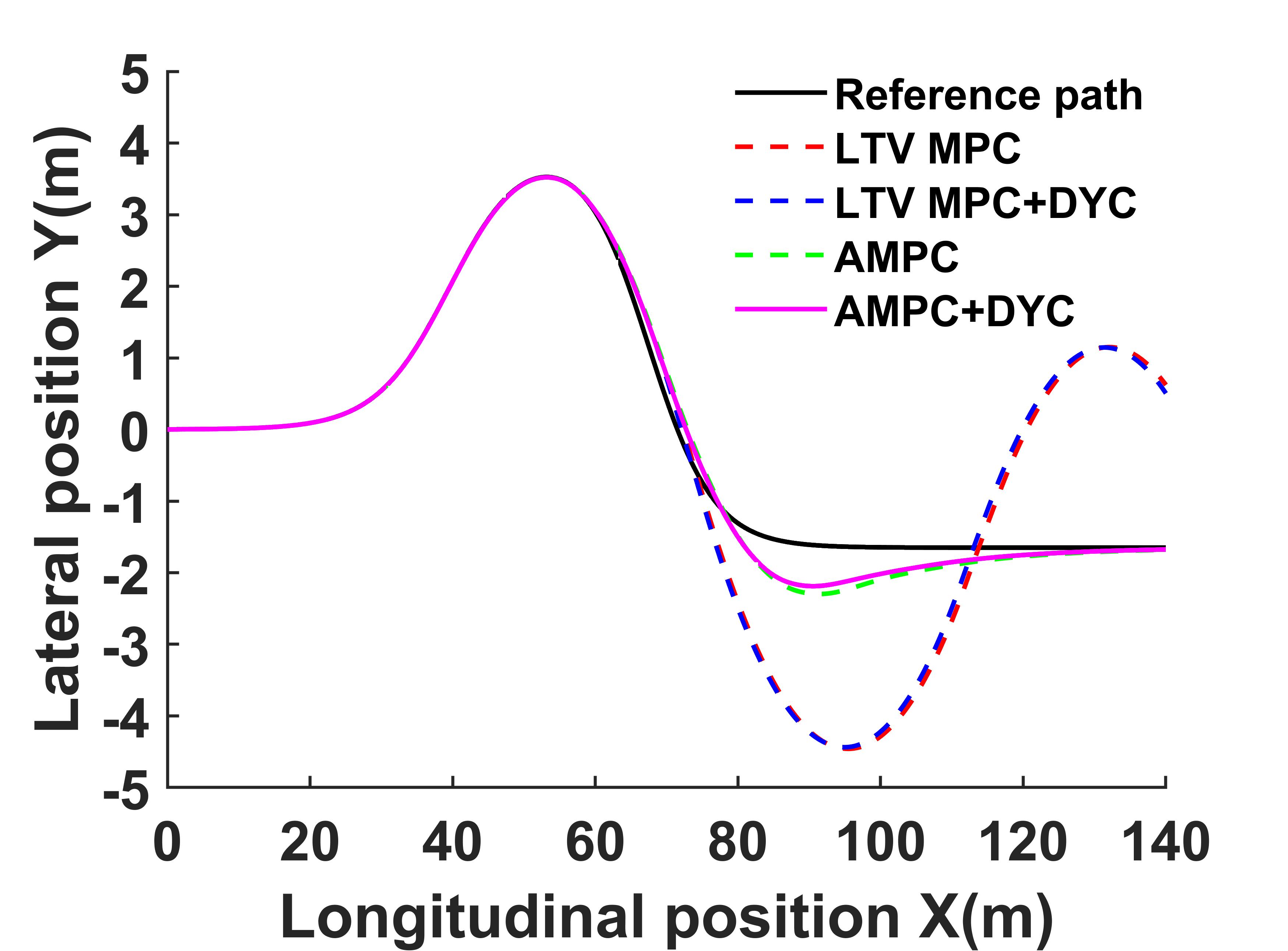}
   }
   \subfigure[Lateral error for 5-65 km/h]{
   \includegraphics[width=0.32\linewidth]{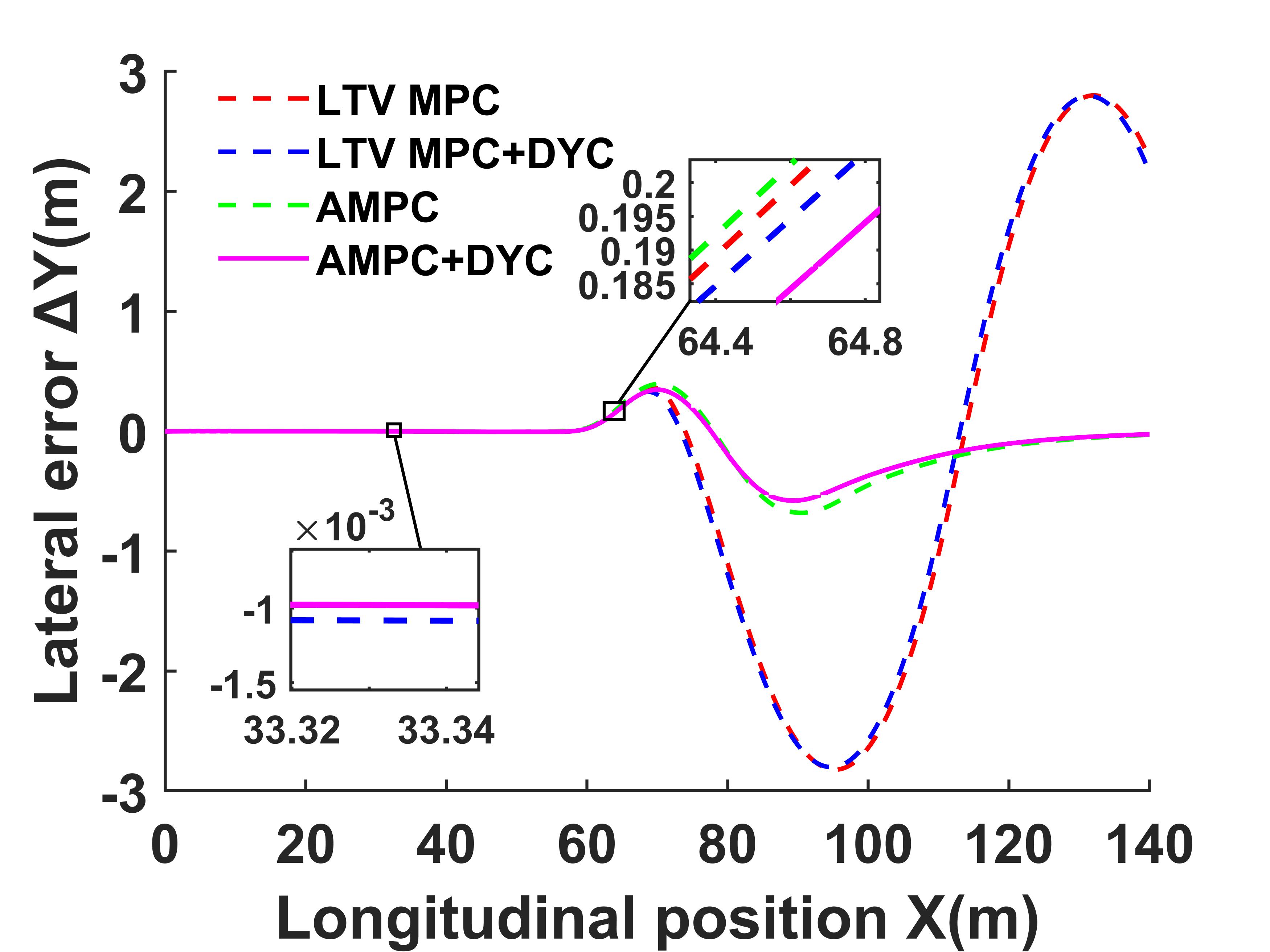}
   }
   \subfigure[Yaw rate for 5-65 km/h]{
   \includegraphics[width=0.32\linewidth]{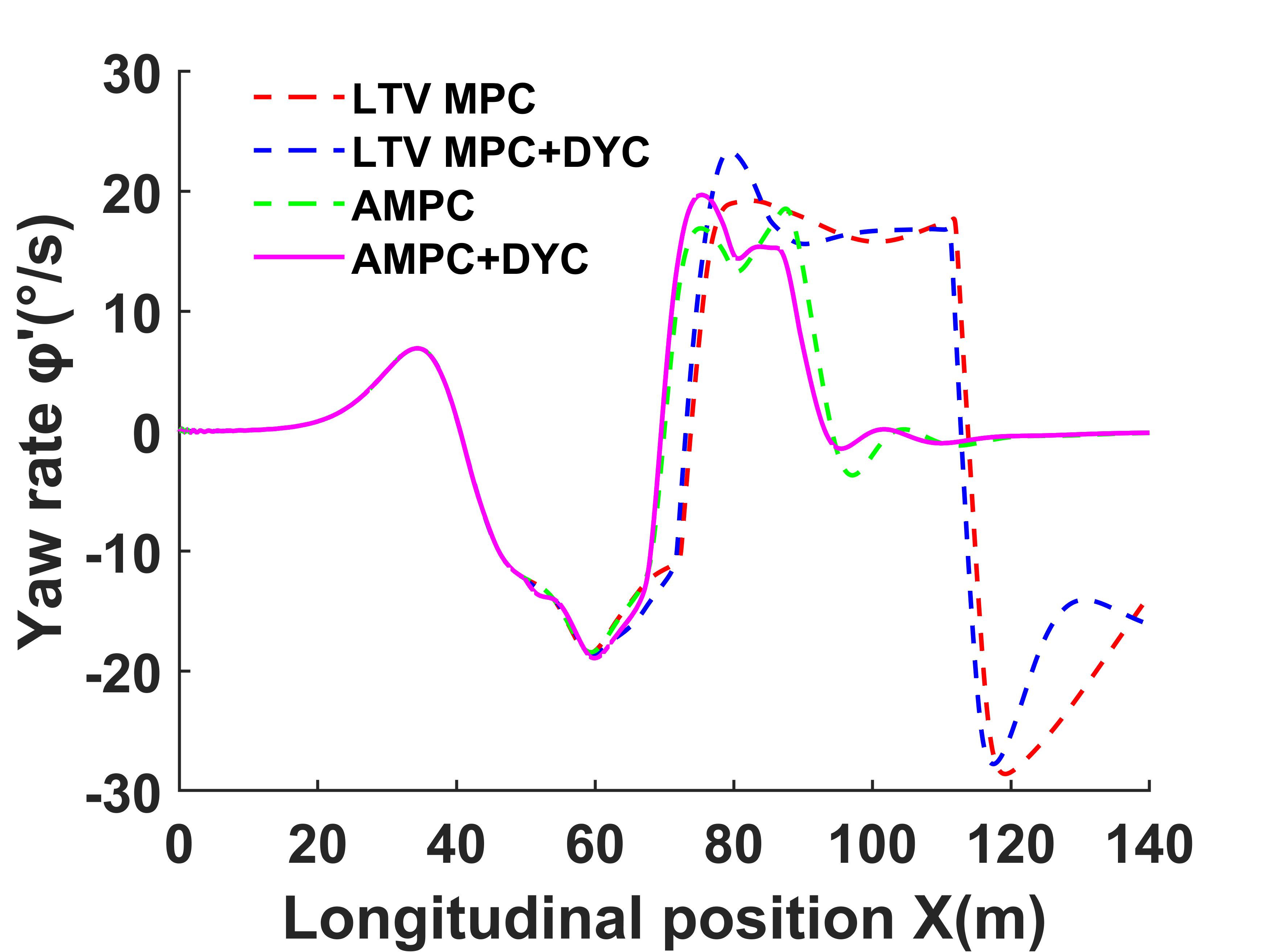}
   }
   \subfigure[Sideslip angle for 5-65 km/h]{
   \includegraphics[width=0.32\linewidth]{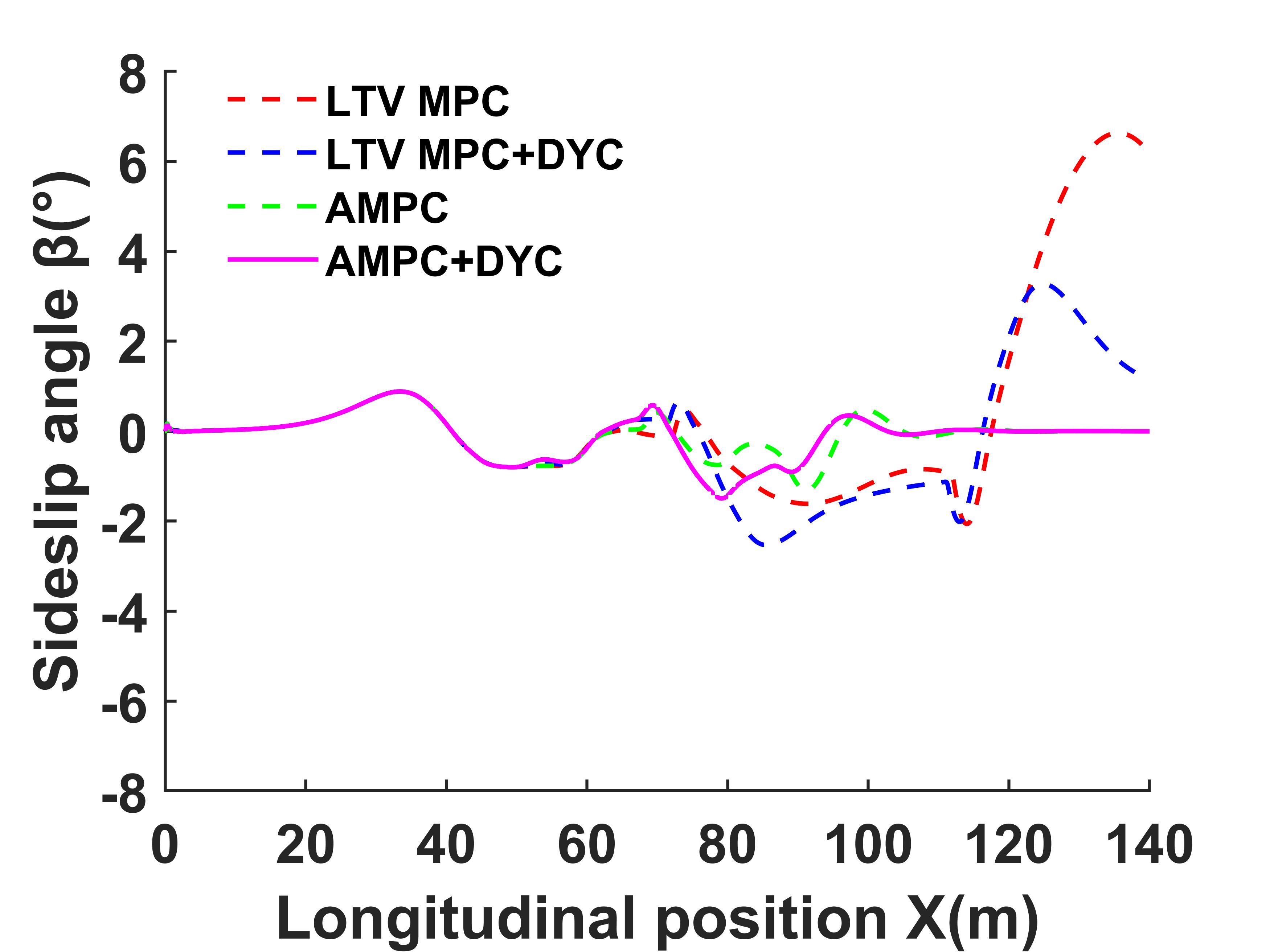}
   }
   \subfigure[Yaw rate error for 5-65 km/h]{
   \includegraphics[width=0.32\linewidth]{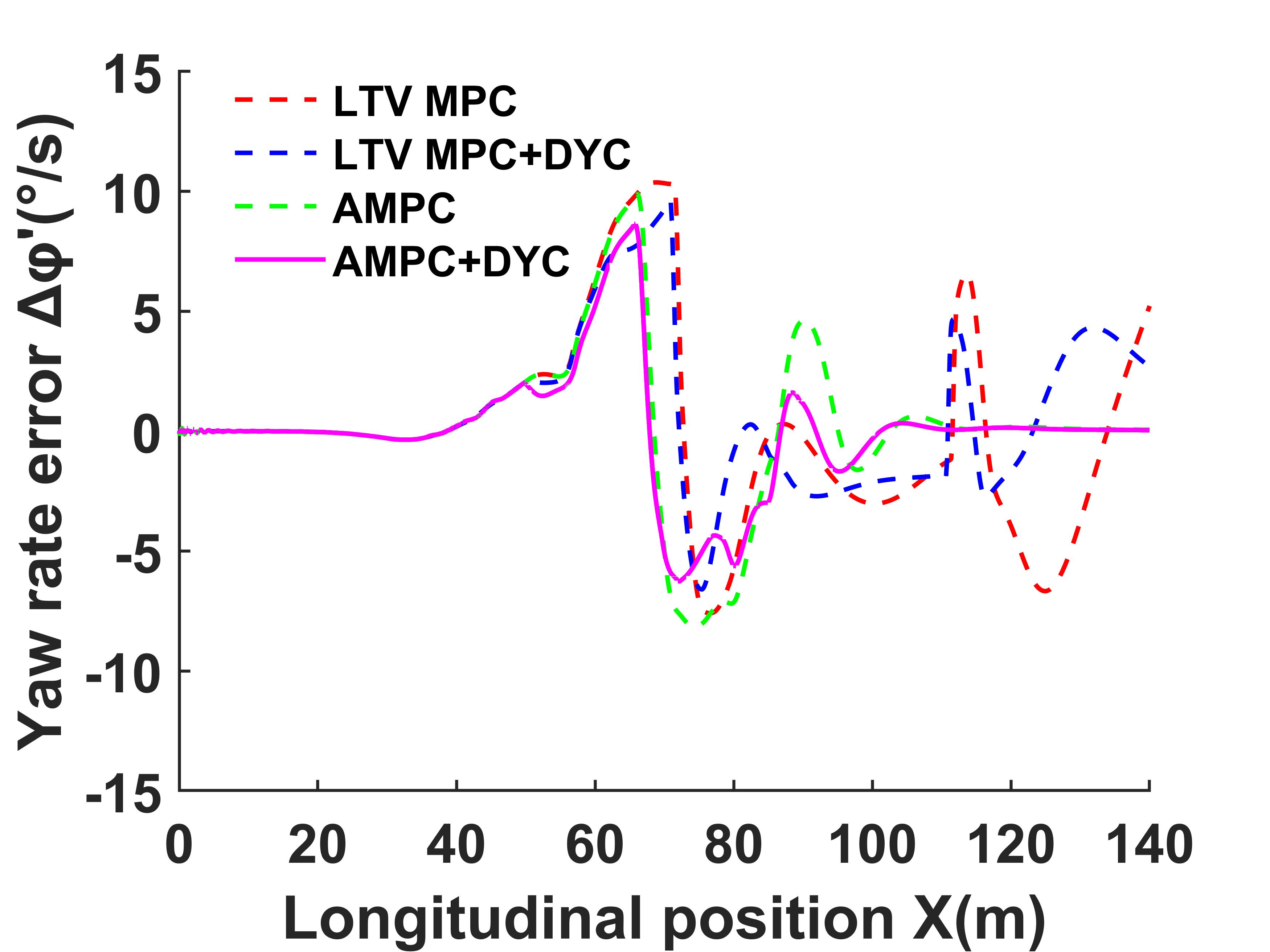}
   }
   \caption{Simulation results of path tracking for 5-65 km/h}
   \label{fig8}
\end{figure*}

\begin{figure*}[h]
   \centering
\subfigure[Intervention signal of DYC for 5-65 km/h]{
   \includegraphics[width=0.32\linewidth]{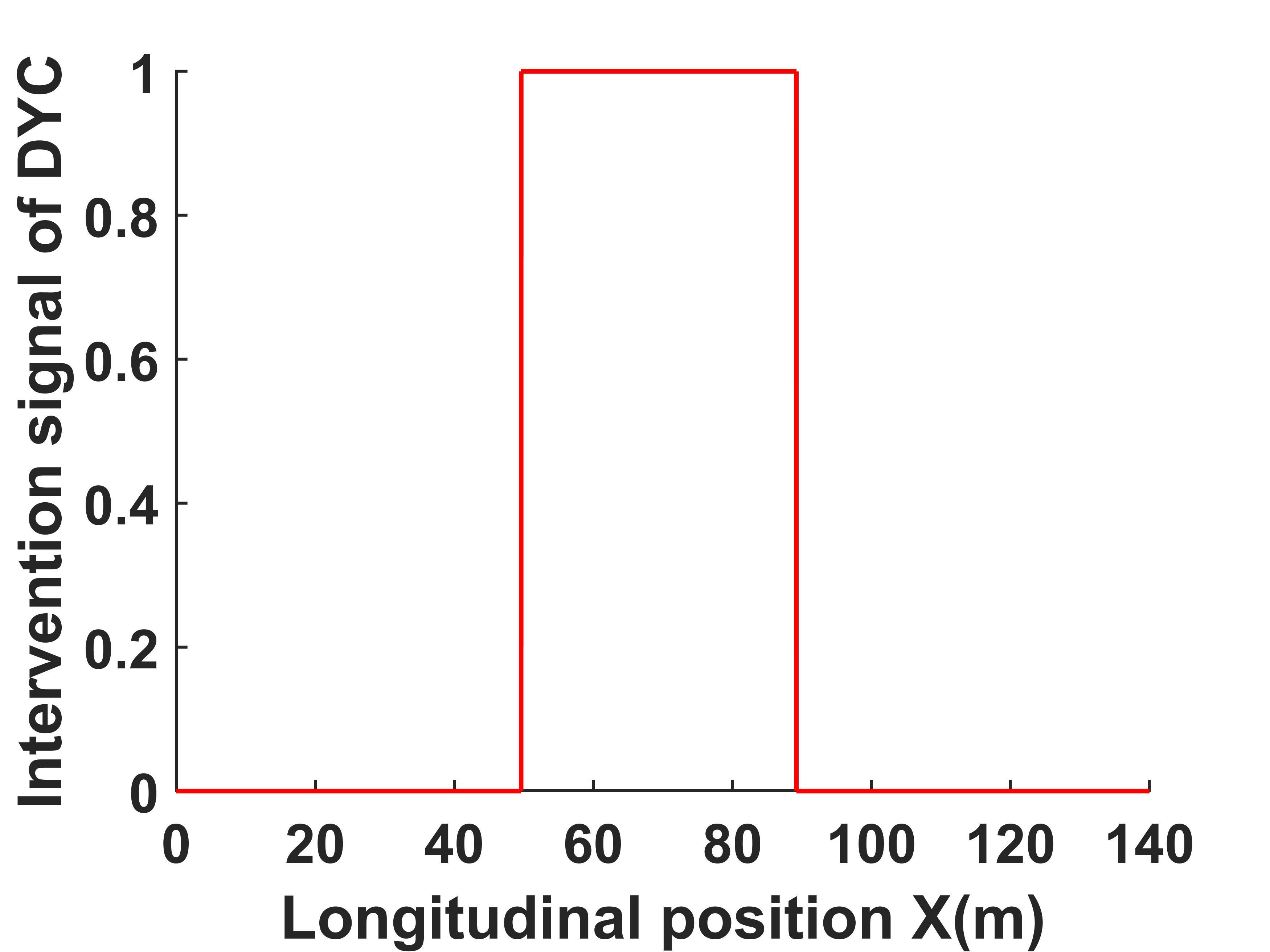}
   }
   \subfigure[Control parameters of AMPC for 5-65 km/h]{
   \includegraphics[width=0.32\linewidth]{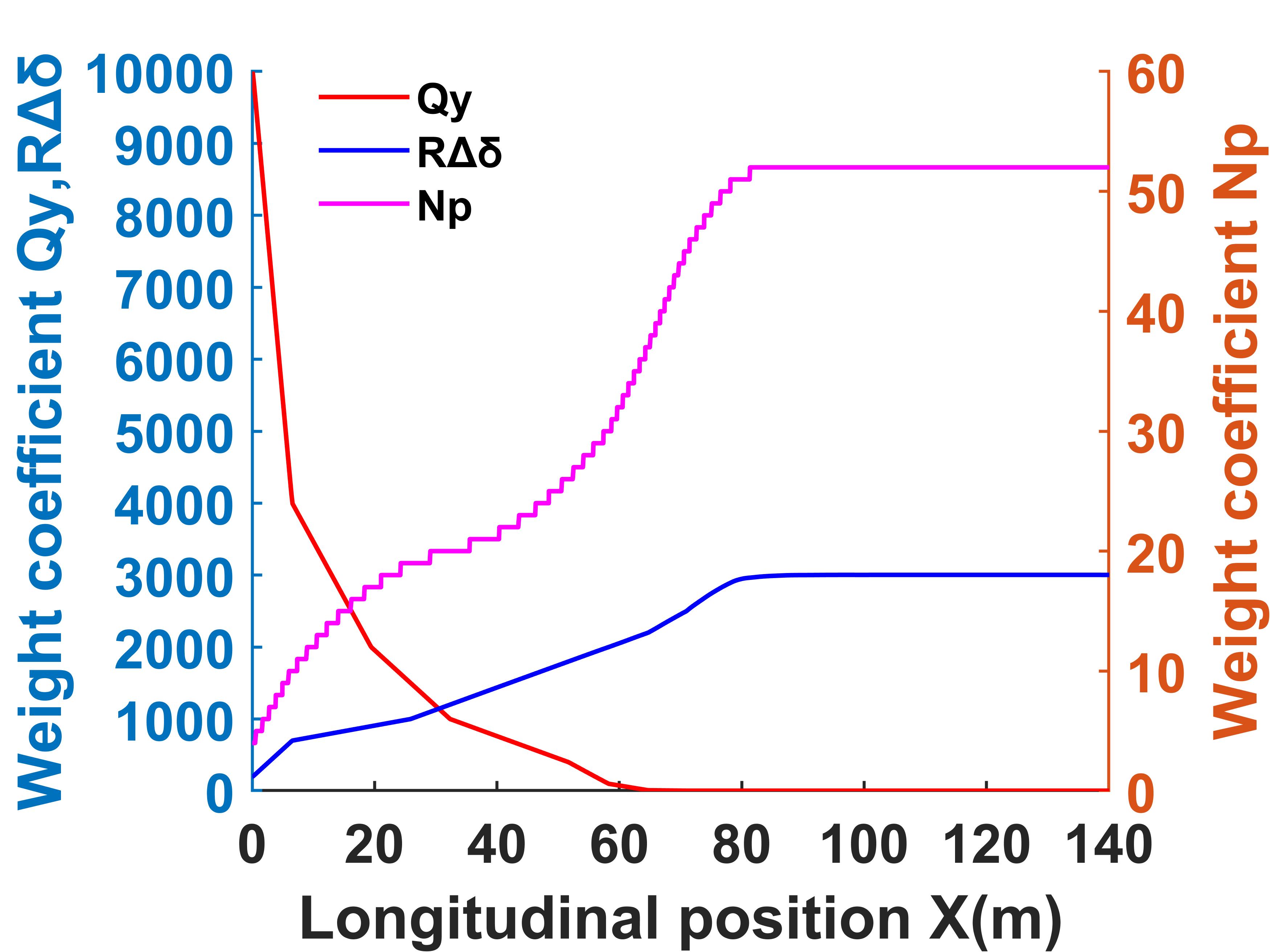}
   }
   \caption{Intervention signal of DYC and control parameters of AMPC for 5-65 km/h}
   \label{fig9}
\end{figure*}

\begin{figure*}[h]
   \centering
\subfigure[Front wheel angle for 5-65 km/h]{
   \includegraphics[width=0.32\linewidth]{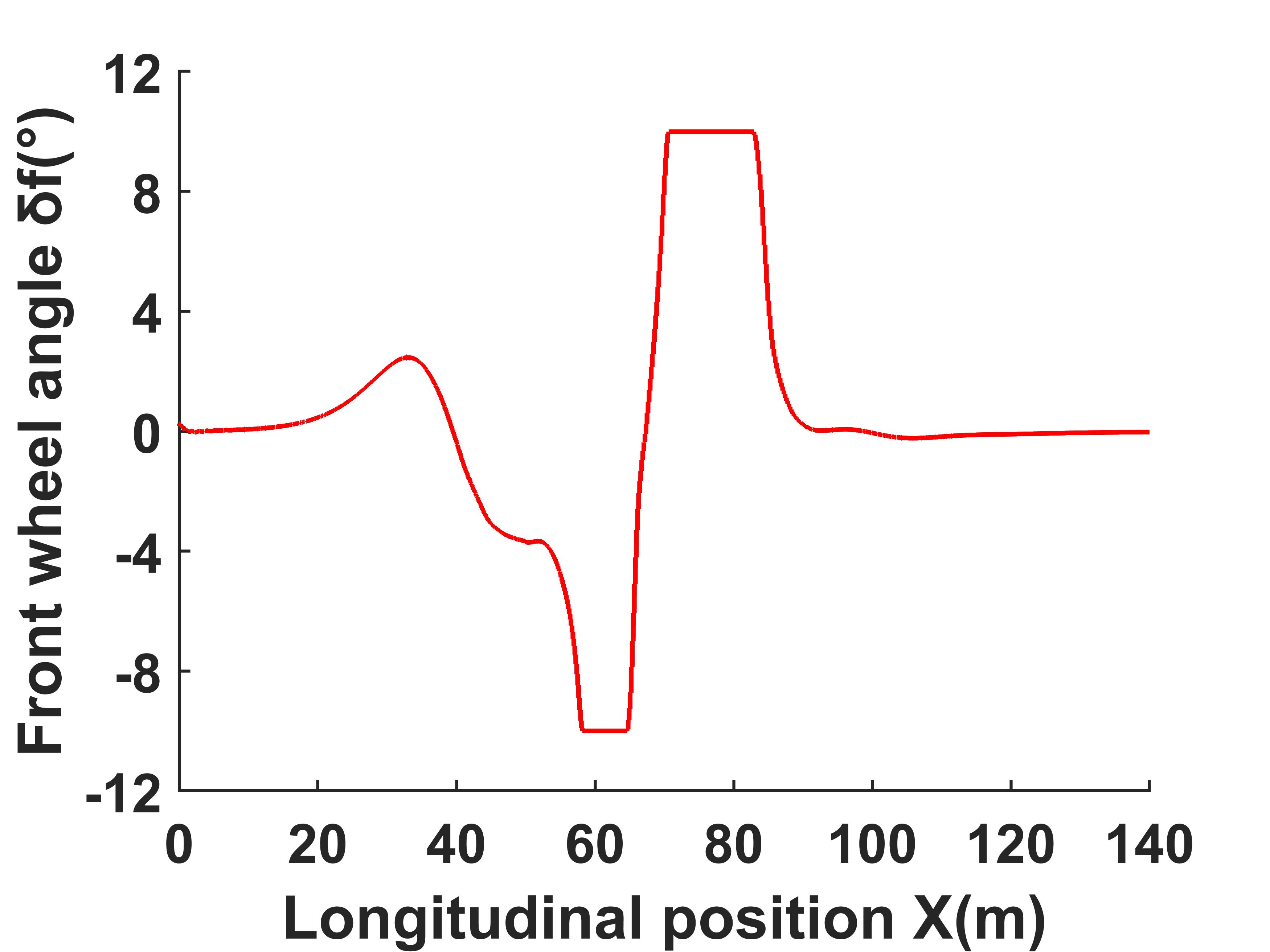}
   }
   \subfigure[Four-wheel torque for 5-65 km/h]{
   \includegraphics[width=0.32\linewidth]{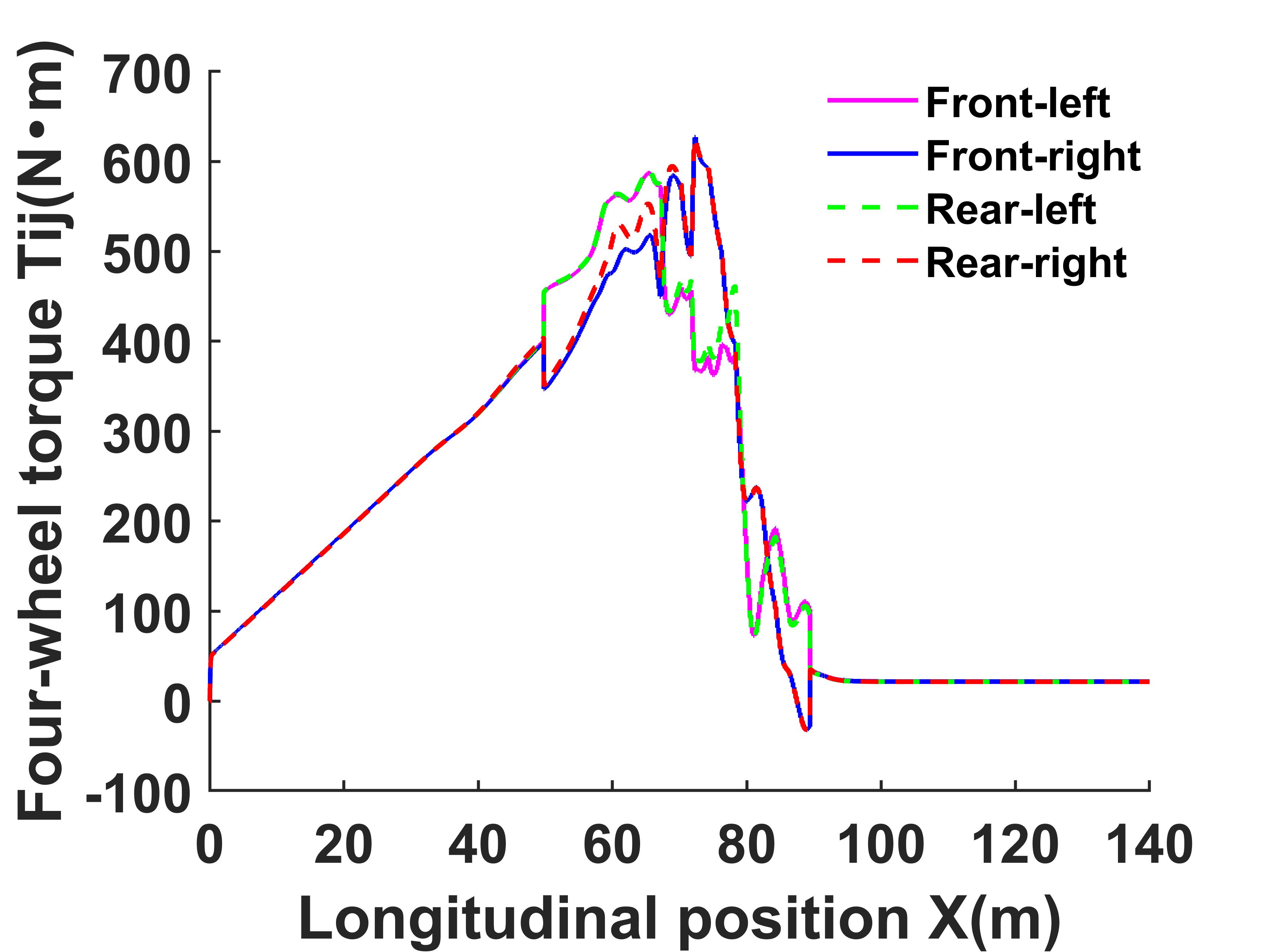}
   }
   \caption{Front wheel angle and four-wheel torque under coordinated control strategy for 5-65 km/h}
   \label{fig10}
\end{figure*}

The longitudinal position $X$-vehicle speed $v_x$ curve is shown in Fig. \ref{fig7}. The vehicle's path-tracking accuracy and stability are compared and analyzed under four control methods: LTV MPC, AMPC, LTV MPC+DYC and AMPC+DYC. As LTV MPC and AMPC require different control parameters, the control parameters for AMPC are taken as the adaptive prediction horizon and weight coefficients corresponding to the vehicle speed of 35 km/h. Under the action of LTV MPC, the prediction horizon $N_p$, weight coefficient $Q_y$ and weight coefficient $R_{\Delta \delta}$ are set to 21, 800 and 1400, respectively. The simulation results are shown in Fig. \ref{fig8}-\ref{fig10} and Tab. \ref{tab4}.

Fig. \ref{fig8} shows the simulation results of path tracking when the vehicle is driven with variable speed in the range of 5-65 km/h. The tracking path and lateral error of the vehicle are shown in Fig. \ref{fig8} (a) and Fig. \ref{fig8} (b), and the yaw rate and sideslip angle of the vehicle are shown in Fig. \ref{fig8} (c) and Fig. \ref{fig8} (d). Fig. \ref{fig9} shows the intervention signal of DYC and the control parameters of AMPC under the coordinated control strategy when the vehicle is driven with variable speed in the range of 5-65 km/h. Fig. \ref{fig10} shows the front wheel angle and four-wheel torque under the coordinated control strategy when the vehicle is driven with variable speed in the range of 5-65 km/h. Tab. \ref{tab4} shows the comparison of root mean square (RMS) of the vehicle's lateral error and yaw rate error under the action of LTV MPC, LTV MPC+DYC, AMPC and AMPC+DYC when the vehicle is driven with variable speed in the range of 5-65 km/h. 
\begin{table*}[h]
\centering
\caption{Comparison of vehicle's performance parameters RMS for 5-65 km/h}
\label{table}
\begin{tabular}{|c|c|c|}
\hline
Algorithm& Lateral error $\Delta \mathrm{Y}$& Yaw rate error $\dot{\varphi}$ \\
\hline
LTV MPC& 1.0135& 2.9742\\
LTV MPC+DYC& 1.0133& 2.3804\\
AMPC& 0.1843& 2.4563\\
AMPC+DYC& 0.1577& 1.9263\\
$\triangle \mathrm{RMS}_{\text {LTV MPC+DYC-LTV MPC }}(\%)$& 0.02$\%$& 19.97$\%$\\
$\triangle \mathrm{RMS}_{\text {AMPC+DYC-LTV MPC }}(\%)$& 81.82$\%$& 17.41$\%$\\
$\triangle \mathrm{RMS}_{\text {AMPC-LTV MPC+DYC }}(\%)$& 81.81$\%$& -3.19$\%$\\
$\triangle \mathrm{RMS}_{\text {AMPC+DYC-AMPC }}(\%)$& 14.43$\%$& 21.58$\%$\\
\hline
\end{tabular}
\label{tab4}
\end{table*}

As shown in Fig. \ref{fig8} and Tab. \ref{tab4}, when compared with LTV MPC, the vehicle's yaw rate and sideslip angle under the action of LTV MPC+DYC are reduced. The RMS of the lateral error is reduced by 0.02$\%$, and the RMS of the yaw rate error is reduced by 19.97$\%$, indicating that the vehicle shows improved stability while maintaining good path-tracking accuracy. From Fig. \ref{fig7}, it can be observed that the vehicle first accelerates and then maintains a constant speed during the path tracking process. The prediction horizon $N_p$, weight coefficient $Q_y$ and weight coefficient $R_{\Delta \delta}$ of AMPC are shown in Fig. \ref{fig9} (b). From Fig. \ref{fig9} (b), it can be observed that the prediction horizon $N_p$ and weight coefficient $R_{\Delta \delta}$ initially increase with the increase in vehicle speed and eventually remain constant, while the weight coefficient $Q_y$ initially decreases with the increase of the vehicle speed and eventually remains constant. From Fig. \ref{fig8}, Fig. \ref{fig9} (b) and Tab. \ref{tab4}, it can be seen that AMPC focuses on improving the vehicle's path-tracking accuracy at low speed while enhancing its stability at high speed. Thus, compared to LTV MPC, AMPC improves the vehicle's path-tracking accuracy at low speed, and enhances the vehicle's stability at high speed with a slight temporary loss of path-tracking accuracy. However, throughout the entire path-tracking process, the RMS of the lateral error and yaw rate are reduced by 81.82$\%$ and 17.41$\%$, respectively, and the yaw rate and sideslip angle of the vehicle are significantly reduced. This indicates that the vehicle shows better path-tracking accuracy and stability. From Fig. \ref{fig8} and Tab. \ref{tab4}, it can be observed that compared to LTV MPC+DYC, the yaw rate and sideslip angle of the vehicle are significantly reduced under the action of AMPC. Additionally, the RMS of lateral error is reduced by 81.81$\%$, while the RMS of yaw rate error is increased by 3.19$\%$. Overall, the results suggest that the vehicle shows better path-tracking accuracy and stability. The intervention signal of DYC when AMPC and DYC act in coordination is shown in Fig. \ref{fig9} (a). When the vehicle is judged to have a tendency of instability, DYC is activated for vehicle stability control. The four-wheel torque decided by DYC is shown in Fig. \ref{fig10} (b), while the vehicle's front wheel angle under the coordination of AMPC and DYC is shown in Fig. \ref{fig10} (a). From Fig. \ref{fig8} and Tab. \ref{tab4}, it can be observed that when AMPC and DYC work in coordination, compared to AMPC, the vehicle's yaw rate changes more smoothly after 85 meters, and the vehicle's sideslip angle does not differ significantly. Additionally, the RMS of lateral error and yaw rate error are reduced by 14.43$\%$ and 21.58$\%$, respectively, indicating that the vehicle shows improved path-tracking accuracy and stability. 

Based on the above analysis, it can be concluded that when the vehicle is driven at variable speeds within the range of 5-65 km/h, the vehicle shows better stability while maintaining good path-tracking accuracy under the action of DYC. However, when DYC and LTV MPC act in coordination, the steering controller still loses the path-tracking ability due to the vehicle's instability, making it difficult to guarantee both stability and path-tracking accuracy of the vehicle. Therefore, it is necessary to perform path-tracking control based on AMPC. The path-tracking controller based on AMPC shows better adaptive ability and path-tracking ability, which can coordinate the path-tracking accuracy and stability of the vehicle at both low and high speeds, and improve the path-tracking accuracy and stability of the vehicle throughout path-tracking process. Furthermore, compared to AMPC, when AMPC and DYC act in coordination, the vehicle shows better stability while maintaining good path-tracking accuracy.

\begin{figure*}[h]
   \centering
\subfigure[Tracking path at 72 km/h]{
   \includegraphics[width=0.32\linewidth]{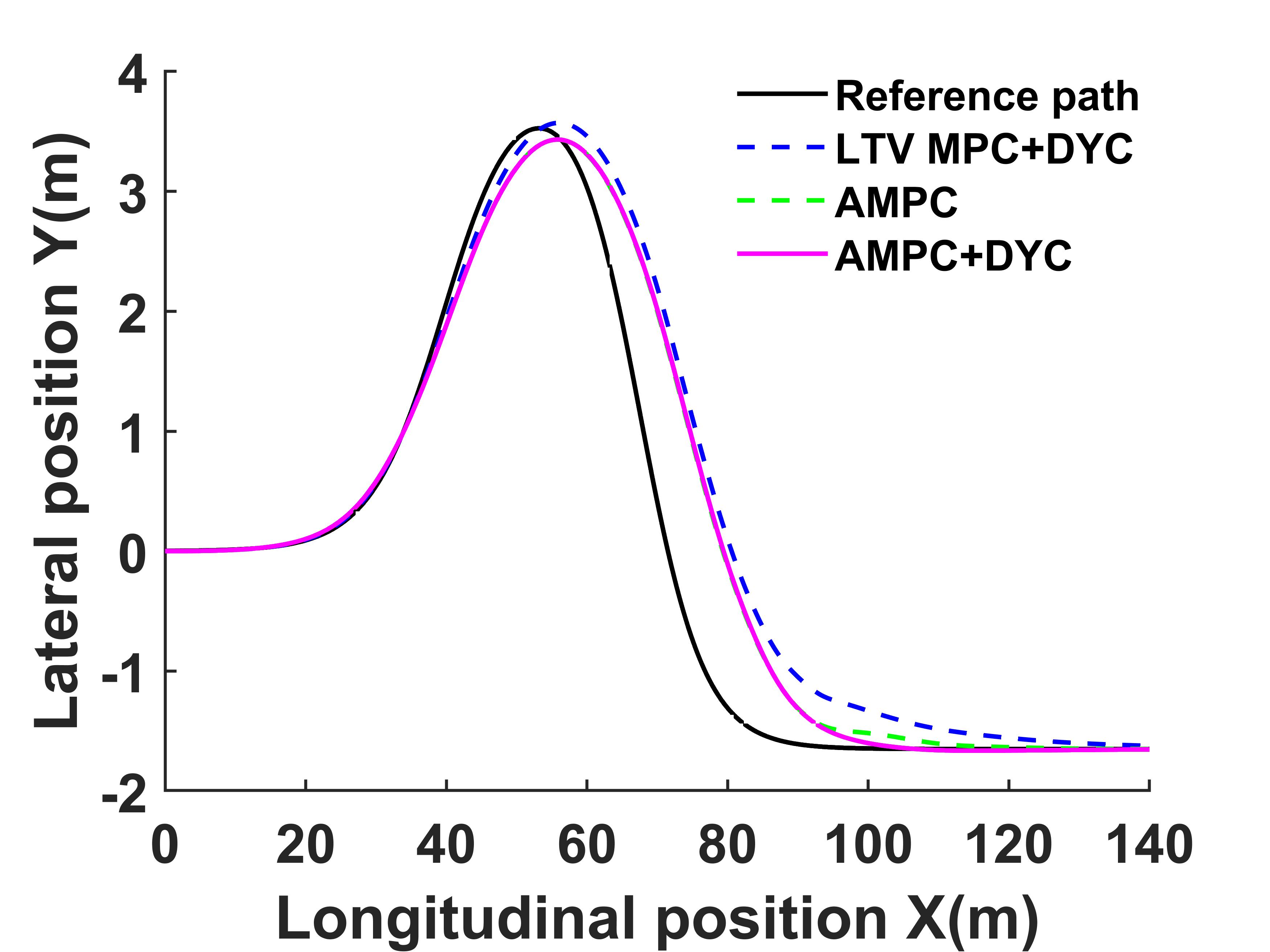}
   }
   \subfigure[Lateral error at 72 km/h]{
   \includegraphics[width=0.32\linewidth]{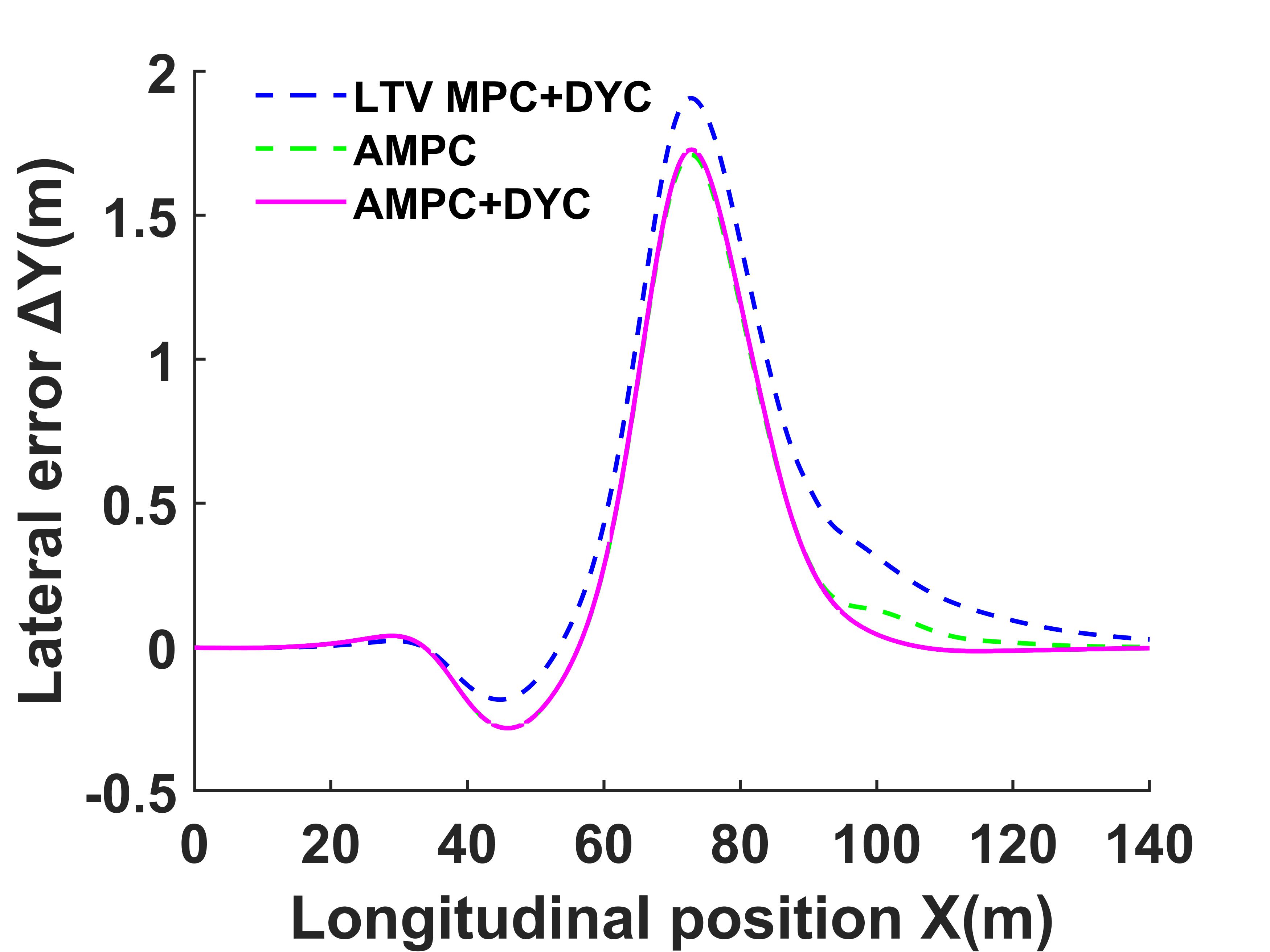}
   }
   \subfigure[Yaw rate at 72 km/h]{
   \includegraphics[width=0.32\linewidth]{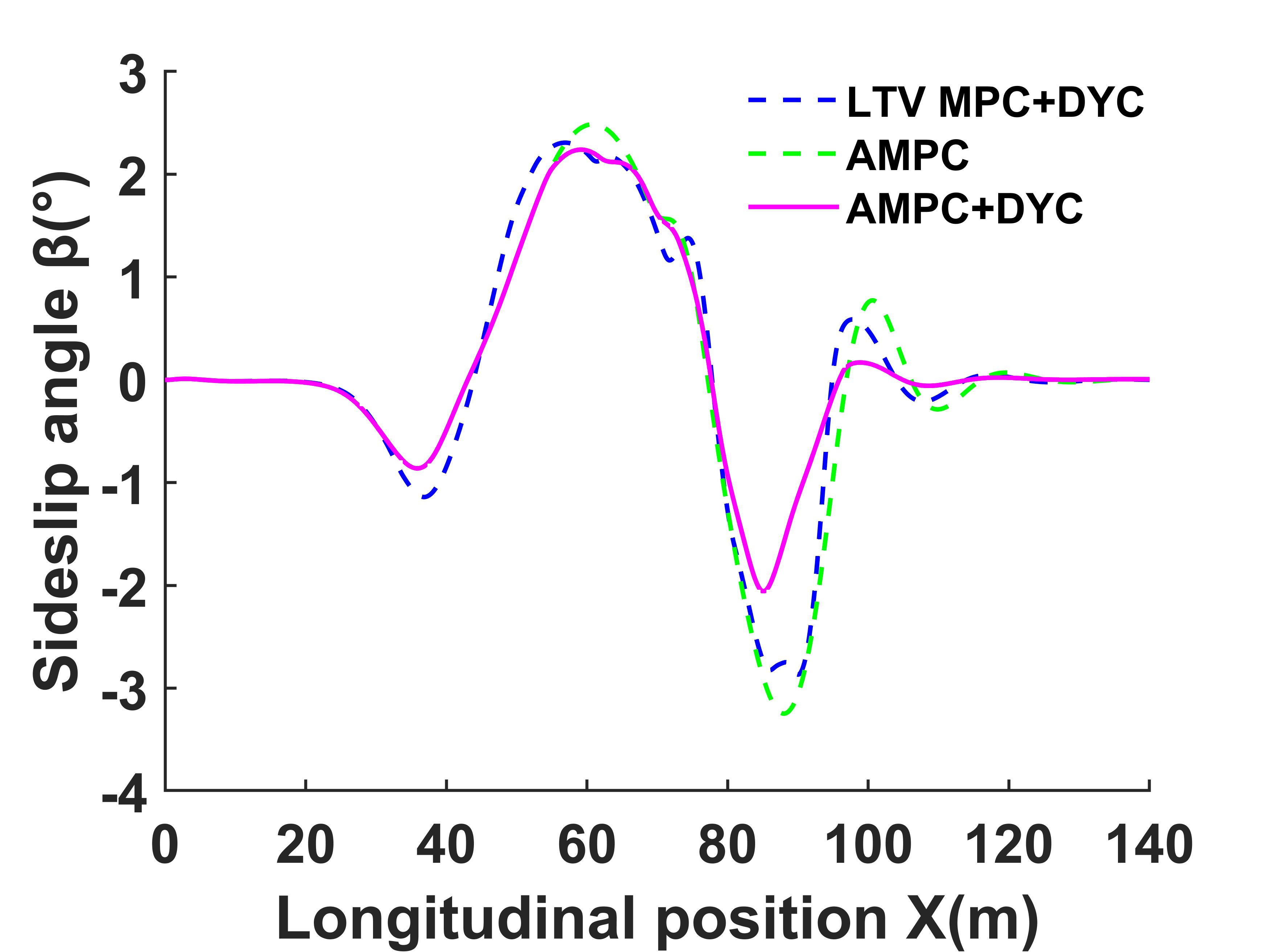}
   }
   \subfigure[Sideslip angle at 72 km/h]{
   \includegraphics[width=0.32\linewidth]{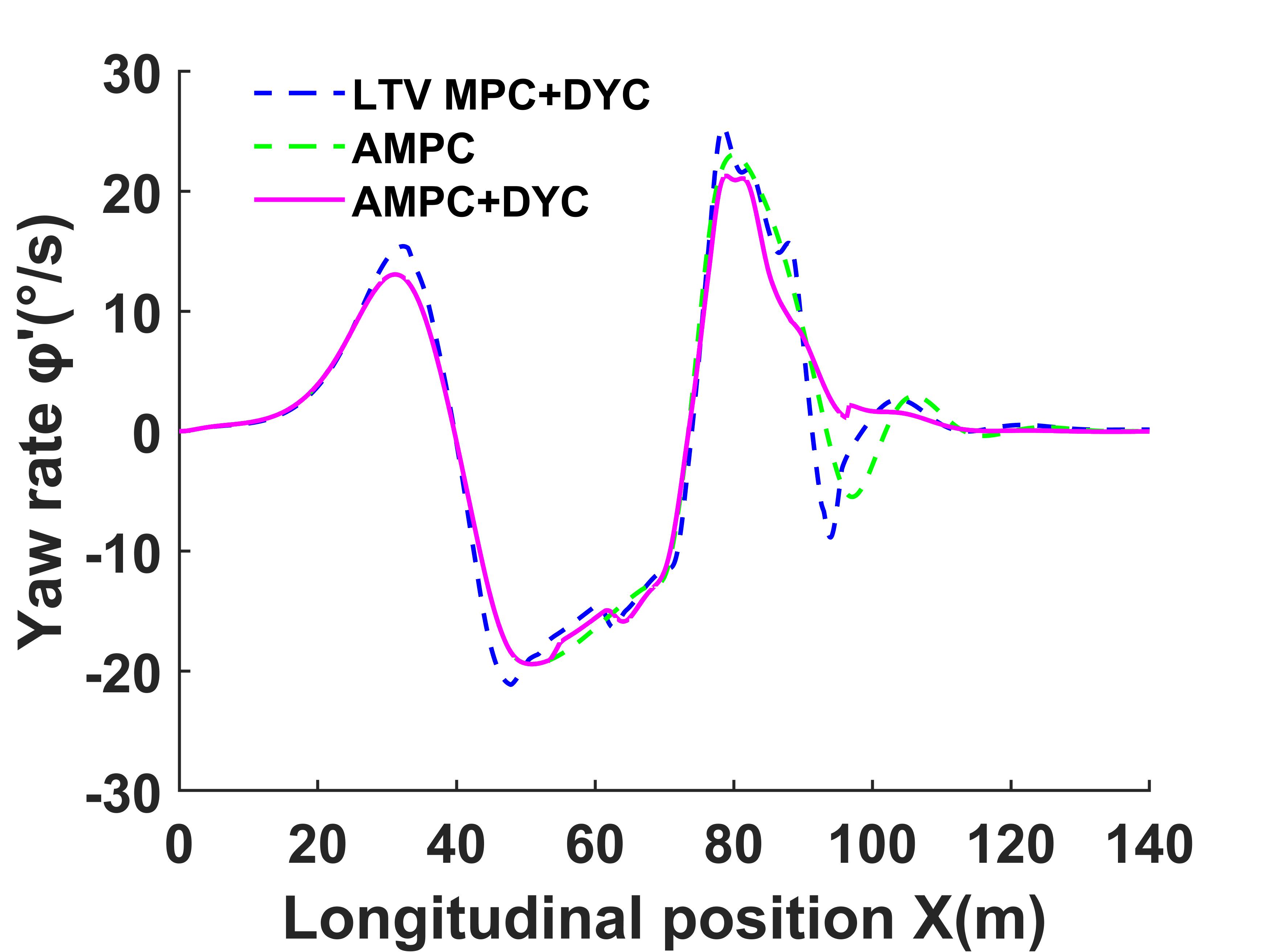}
   }
   \subfigure[Yaw rate error at 72 km/h]{
   \includegraphics[width=0.32\linewidth]{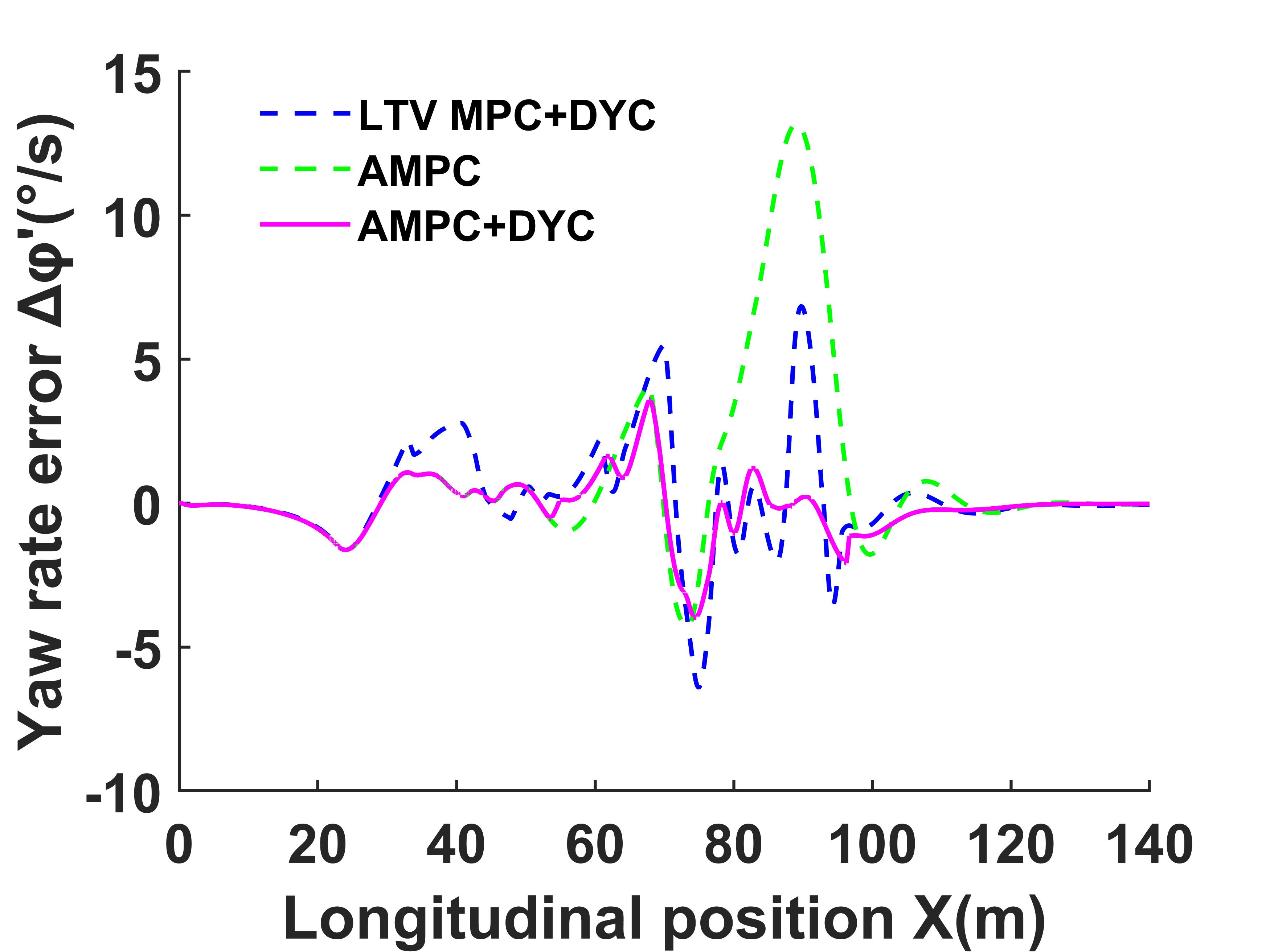}
   }
   \caption{Simulation results of path tracking at 72 km/h}
   \label{fig11}
\end{figure*}
\begin{figure*}[h]
   \centering
   \subfigure[Intervention signal of DYC at 72 km/h]{
   \includegraphics[width=0.32\linewidth]{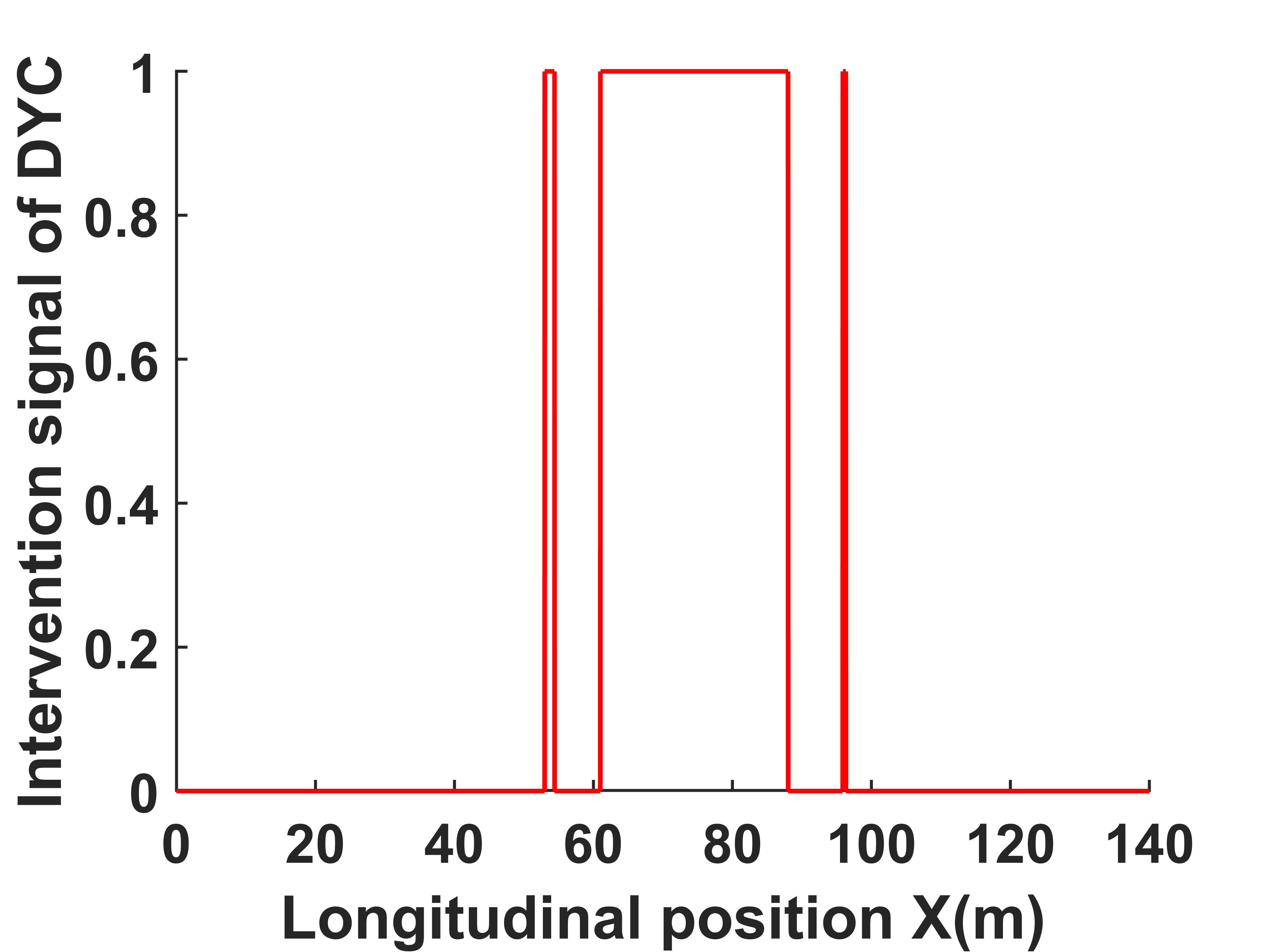}
   }
   \subfigure[Front wheel angle at 72 km/h]{
   \includegraphics[width=0.32\linewidth]{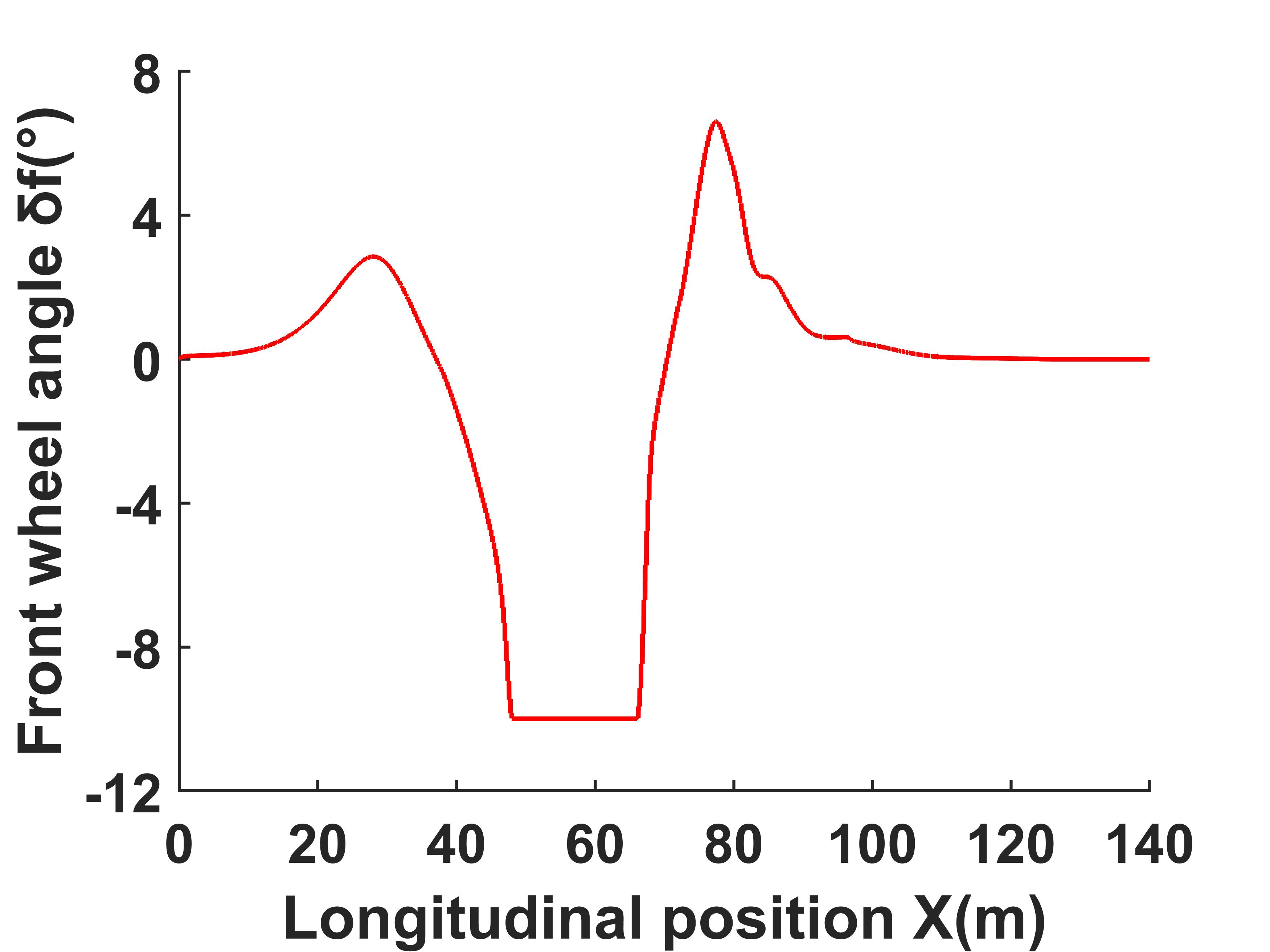}
   }
   \subfigure[Four-wheel torque at 72 km/h]{
   \includegraphics[width=0.32\linewidth]{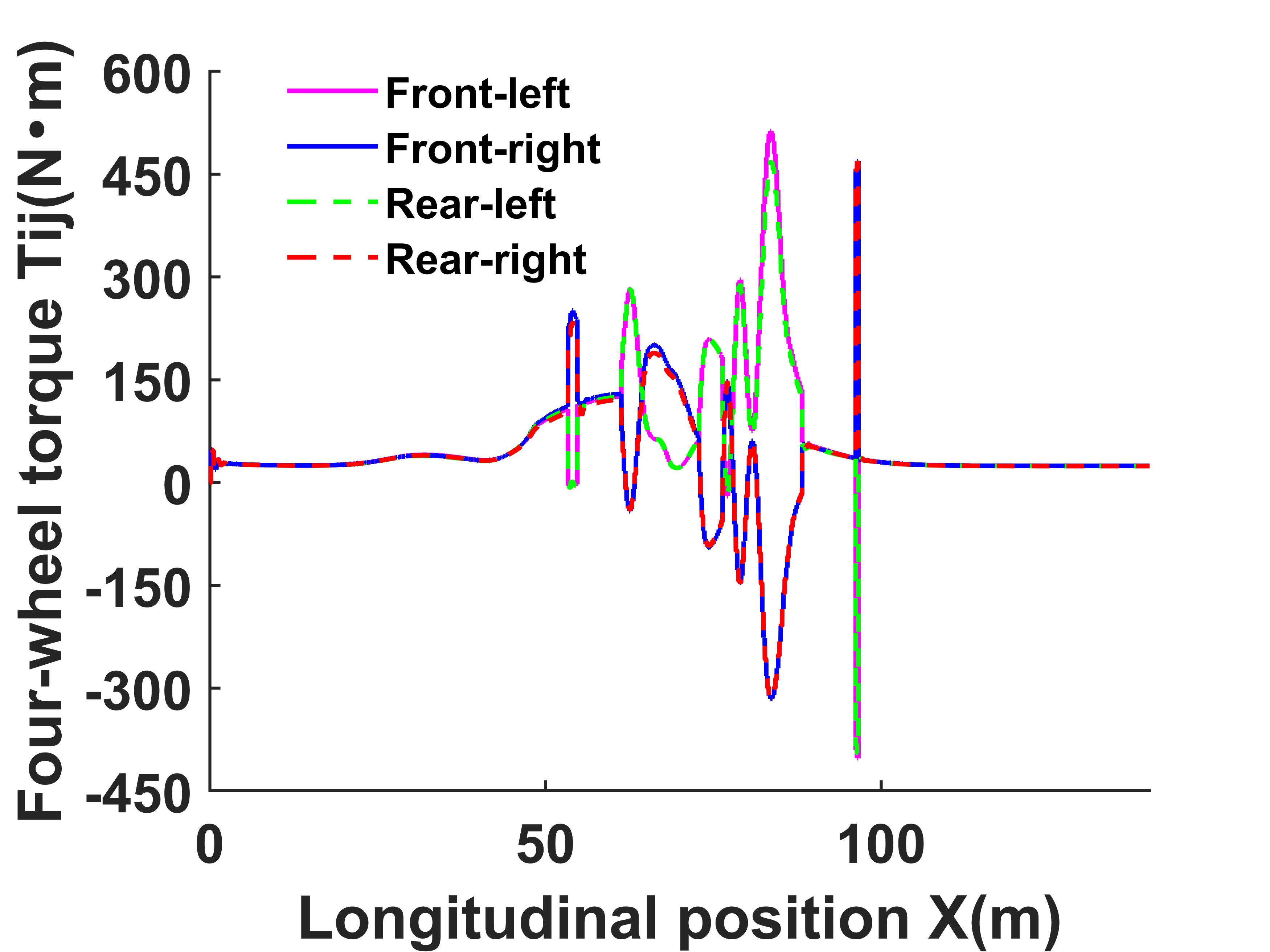}
   }
   \caption{Intervention signal of DYC, front wheel angle and four-wheel torque under coordinated control strategy at 72 km/h}
   \label{fig12}
\end{figure*}
\begin{table*}[h]
\centering
\caption{Comparison of vehicle's performance parameters RMS at 72 km/h}
\label{table}
\begin{tabular}{|c|c|c|}
\hline
Algorithm& Lateral error $\Delta \mathrm{Y}$& Yaw rate error $\dot{\varphi}$ \\
\hline
LTV MPC+DYC& 0.6489& 1.8686\\
AMPC& 0.5518& 3.3919\\
AMPC+DYC& 0.5564& 1.0409\\
$\triangle \mathrm{RMS}_{\text {LTV MPC+DYC-AMPC }}(\%)$& -14.96$\%$& 81.52$\%$\\
$\triangle \mathrm{RMS}_{\text {AMPC+DYC-AMPC }}(\%)$& -0.83$\%$& 69.31$\%$\\
$\triangle \mathrm{RMS}_{\text {AMPC+DYC-LTV MPC+DYC }}(\%)$& 14.25$\%$& 44.30$\%$\\
\hline
\end{tabular}
\label{tab5}
\end{table*}

(2) The vehicle speed is set to 72 km/h.

The path-tracking accuracy and stability of the vehicle are compared and analyzed under the three control methods: LTV MPC+DYC, AMPC and AMPC+DYC. In order to compare the vehicle's performance when LTV MPC and AMPC act in coordination with DYC, respectively, it is necessary to select different control parameters for LTV MPC than for AMPC. Since the vehicle speed's range of adaptive parameter adjustment is 0-120 km/h, the control parameters of LTV MPC are chosen as the adaptive prediction horizon and weight coefficients corresponding to the vehicle speed of 60km/h. Under the action of LTV MPC, the prediction horizon $N_p$, weight coefficient $Q_y$ and weight coefficient $R_{\Delta \delta}$ are set to 45, 4 and 2500, respectively. The simulation results are shown in Fig. \ref{fig11}-\ref{fig12} and Tab. \ref{tab5}. 

Fig. \ref{fig11} shows the simulation results of path tracking at the vehicle speed of 72 km/h. The tracking path and lateral error of the vehicle are shown in Fig. \ref{fig11} (a) and Fig. \ref{fig11} (b) and the yaw rate, sideslip angle and yaw rate error of the vehicle are shown in Fig. \ref{fig11} (c), Fig. \ref{fig11} (d) and Fig. \ref{fig11} (e). Fig. \ref{fig12} shows the intervention signal of DYC, the front wheel angle and four-wheel torque under coordinated control strategy at the vehicle speed of 72 km/h. Tab. \ref{tab5} shows the comparison of root mean square (RMS) of the vehicle's lateral error and yaw rate error under the action of LTV MPC+DYC, AMPC and AMPC+DYC at the vehicle speed of 72 km/h. 

Since AMPC aims to enhance the vehicle's stability at high speed, its prediction horizon $N_p$ and weight coefficient $Q_y$ should be increased, while its weight coefficient $R_{\Delta \delta}$ should be decreased compared to LTV MPC. Based on the adaptive adjustment of the prediction horizon and weight coefficients, the prediction horizon $N_p$, weight coefficient $Q_y$ and weight coefficient $R_{\Delta \delta}$ of AMPC are set to 63, 3.8 and 3700, respectively. As shown in Fig. \ref{fig11} and Tab. \ref{tab5}, compared with AMPC, the yaw rate and sideslip angle of the vehicle under the coordinated action of LTV MPC and DYC are not much different, and the RMS of the lateral error increases by 14.96$\%$, and the RMS of the yaw rate error decreases by 81.52$\%$. These results indicate that the vehicle obviously shows better stability with a small loss of path-tracking accuracy. The intervention signal of DYC when AMPC and DYC act in coordination is shown in Fig. \ref{fig12} (a). When the vehicle is judged to have a tendency of instability, DYC is activated for vehicle stability control. The four-wheel torque decided by DYC is shown in Fig. \ref{fig12} (c) and the vehicle's front wheel angle under the coordination of AMPC and DYC is shown in Fig. \ref{fig12} (b). From Fig. \ref{fig11} and Tab. \ref{tab5}, it can be seen that when AMPC and DYC act in coordination, the yaw rate and sideslip angle of the vehicle, as well as their fluctuation range, are significantly reduced compared to AMPC, and the vehicle's yaw rate changes more smoothly after 76 meters. The RMS of lateral error increases by 0.83$\%$, while the RMS of the yaw rate error decreases by 69.31$\%$. This indicates that the vehicle obviously shows improved stability with only a slight decrease in path-tracking accuracy. Compared with LTV MPC+DYC, the yaw rate, sideslip angle of the vehicle and their fluctuation range are also significantly reduced when AMPC and DYC act in coordination. Additionally, the RMS of the lateral error and yaw rate error is reduced by 14.25$\%$ and 44.30$\%$, respectively. This indicates that the vehicle shows improved path-tracking accuracy and stability.

Based on the above analysis, the following conclusions can be drawn that when the vehicle is driving at the high speed of 72 km/h, compared to AMPC, the vehicle shows significantly better stability with only a slight loss of path-tracking accuracy when LTV MPC and DYC act in coordination. Therefore, it is necessary to perform coordinated control of path tracking and direct yaw moment. Moreover, when AMPC and DYC act in coordination, the vehicle shows better stability while maintaining good path-tracking accuracy compared to AMPC and LTV MPC+DYC.

\section{Conclusion}

This paper proposes a novel coordinated control strategy integrating AMPC and DYC to enhance the performance of DDEV's path tracking and motion stability. The conclusions are as follows:

(1) Compared to LTV MPC, the proposed algorithm based on AMPC demonstrates superior adaptability to different vehicle speeds. Especially, it can autonomously adjust the prediction horizon $N_p$, weight coefficient $Q_y$, and weight coefficient $R_{\Delta \delta}$ of MPC considering vehicle speeds. This effectively coordinates the path-tracking accuracy and stability of the vehicle, resulting in improved yaw stability and path-tracking accuracy under different vehicle speeds.

(2) Compared to standalone AMPC or DYC, the proposed coordinated control strategy can significantly enhance the yaw stability of a vehicle, while also maintaining excellent path-tracking accuracy under extreme conditions. In situations of varying speed and low adhesion, the combined use of AMPC and DYC leads to improved yaw stability and path-tracking accuracy compared to using AMPC alone. Additionally, at high speeds with low adhesion, while the path-tracking accuracy may suffer a minor loss, the yaw stability of the vehicle is significantly improved in comparison to using AMPC alone.

In the future, the impact of factors other than vehicle speed on path-tracking accuracy and stability will be explored. Additionally, a real vehicle test platform will be constructed to verify the effectiveness of the proposed algorithm.

\begin{acks}
This work was supported by Key R\&D Project of Hubei Province, China (2022BAA074) and Foshan Xianhu Laboratory of the Advanced Energy Science and Technology Guangdong Laboratory (XHD2020-003).\\
\end{acks}

\bibliographystyle{IEEEtran}
\bibliography{Refs}

\end{document}